\documentclass[11pt]{article}

\usepackage[final]{acl}

\usepackage{times}
\usepackage{latexsym}

\usepackage[T1]{fontenc}
\usepackage[utf8]{inputenc}

\usepackage{microtype}

\usepackage{inconsolata}

\usepackage{graphicx}

\usepackage{booktabs}  
\usepackage{multirow}   
\usepackage{xcolor}    

\usepackage{newtxtext,newtxmath}
\usepackage[utf8]{inputenc} % allow utf-8 input
\usepackage{hyperref}       % hyperlinks
\usepackage{url}            % simple URL typesetting
\usepackage{booktabs}       % professional-quality tables
\usepackage{amsfonts}       % blackboard math symbols
\usepackage{nicefrac}       % compact symbols for 1/2, etc.
\usepackage{microtype}      % microtypography
\usepackage{colortbl} 
\usepackage{makecell} 
\usepackage[skip=5pt]{caption} 
\usepackage{amsmath}
\usepackage{listings}

\usepackage{float}
\usepackage{subfigure}

\usepackage[most]{tcolorbox}

\newtcolorbox{promptbox}[1]{
  colback=gray!5!white,
  colframe=black!75!black,
  fonttitle=\bfseries,
  title={#1},
  arc=2pt,
  boxrule=1pt,
  left=4pt, right=4pt, top=4pt, bottom=4pt
}

\usepackage{geometry}
\usepackage{enumitem}

\newcommand{\aicalogo}{%
 \raisebox{-1.0ex}{\includegraphics[height=1.8em]{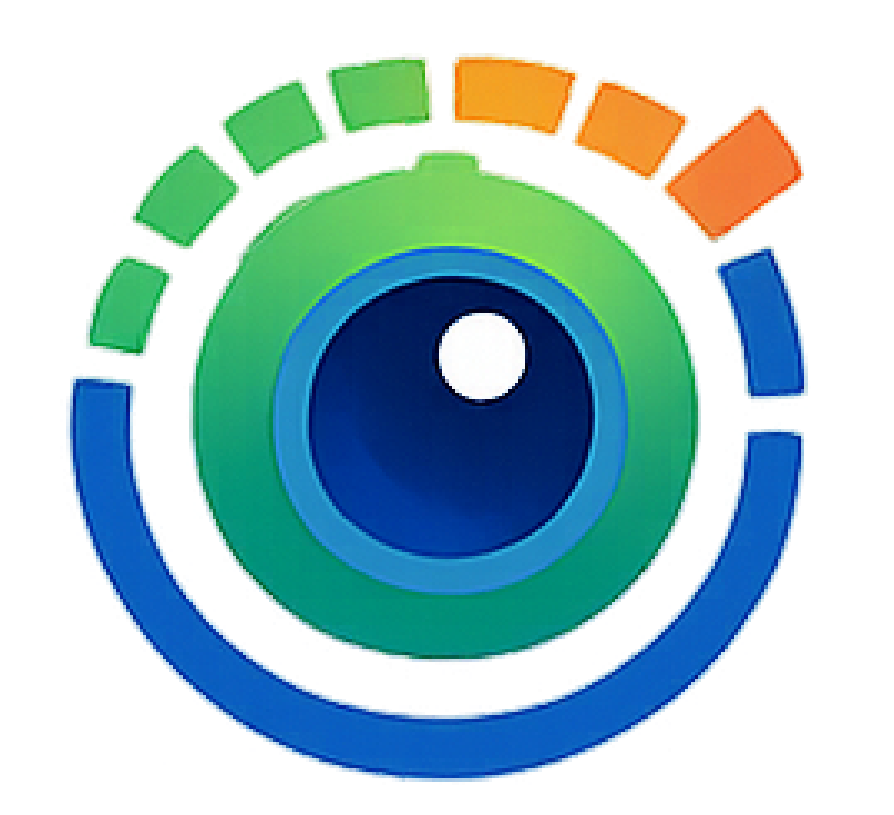}}%
}

\title{\aicalogo AICA-Bench: Holistically Examining the Capabilities of VLMs in Affective Image Content Analysis}

\author{
Dong She$^{*}$ \and Xianrong Yao$^{*}$ \and Liqun Chen \and Jinghe Yu \and Yang Gao \and Zhanpeng Jin$^{\dagger}$ \\
School of Future Technology,South China University of Technology,Guangzhou \\
\texttt{\{ftdshe,ftxryao,ftchenliqvn,ftyujinghe\}@mail.scut.edu.cn} \\
\texttt{\{gaoyang2025,zjin\}@scut.edu.cn} \\
\small{$^{*}$ Equal contribution \hspace{1em} $^{\dagger}$ Corresponding author}
}

\begin{document}
\maketitle
\begin{abstract}
Vision-Language Models (VLMs) have demonstrated strong capabilities in perception, yet holistic Affective Image Content Analysis (AICA)—which integrates perception, reasoning, and generation into a unified framework—remains underexplored. To address this, we introduce AICA-Bench, a comprehensive benchmark comprising three core tasks: Emotion Understanding (EU), Reasoning (ER), and Generation (EGCG). We evaluate 23 VLMs, revealing critical gaps: models struggle with intensity calibration and suffer from descriptive shallowness in open-ended tasks. To bridge these gaps, we propose Grounded Affective Tree (GAT) Prompting, a training-free framework that integrates visual scaffolding with hierarchical reasoning. Experiments show that GAT effectively corrects intensity errors and significantly enhances descriptive depth, establishing a robust baseline for future affective multimodal research.
\end{abstract}

\section{Introduction}
\label{sec:intro}

Recent years have witnessed rapid advances in Vision-Language Models (VLMs)~\cite{liu2023llava, liu2023improvedllava, liu2024llavanext, Qwen2-VL, Qwen2.5-VL, InternVL2.5, InternVL3, yao2024minicpm}, which integrate visual and textual modalities to perform a wide range of tasks, from image captioning and visual question answering to grounded reasoning. 
To assess these capabilities, a variety of benchmarks~\cite{yue2024mmmu, MMBench, fu2024mme, wang2023gvt} have been developed, focusing primarily on factual correctness, semantic grounding, visual reasoning, or multi-discipline understanding.
Yet emotional intelligence remains an underexplored but essential aspect of evaluating VLMs and multimodal large language models (MLLMs). 

\begin{figure}[htb]
    \centering
    \includegraphics[width=\linewidth]{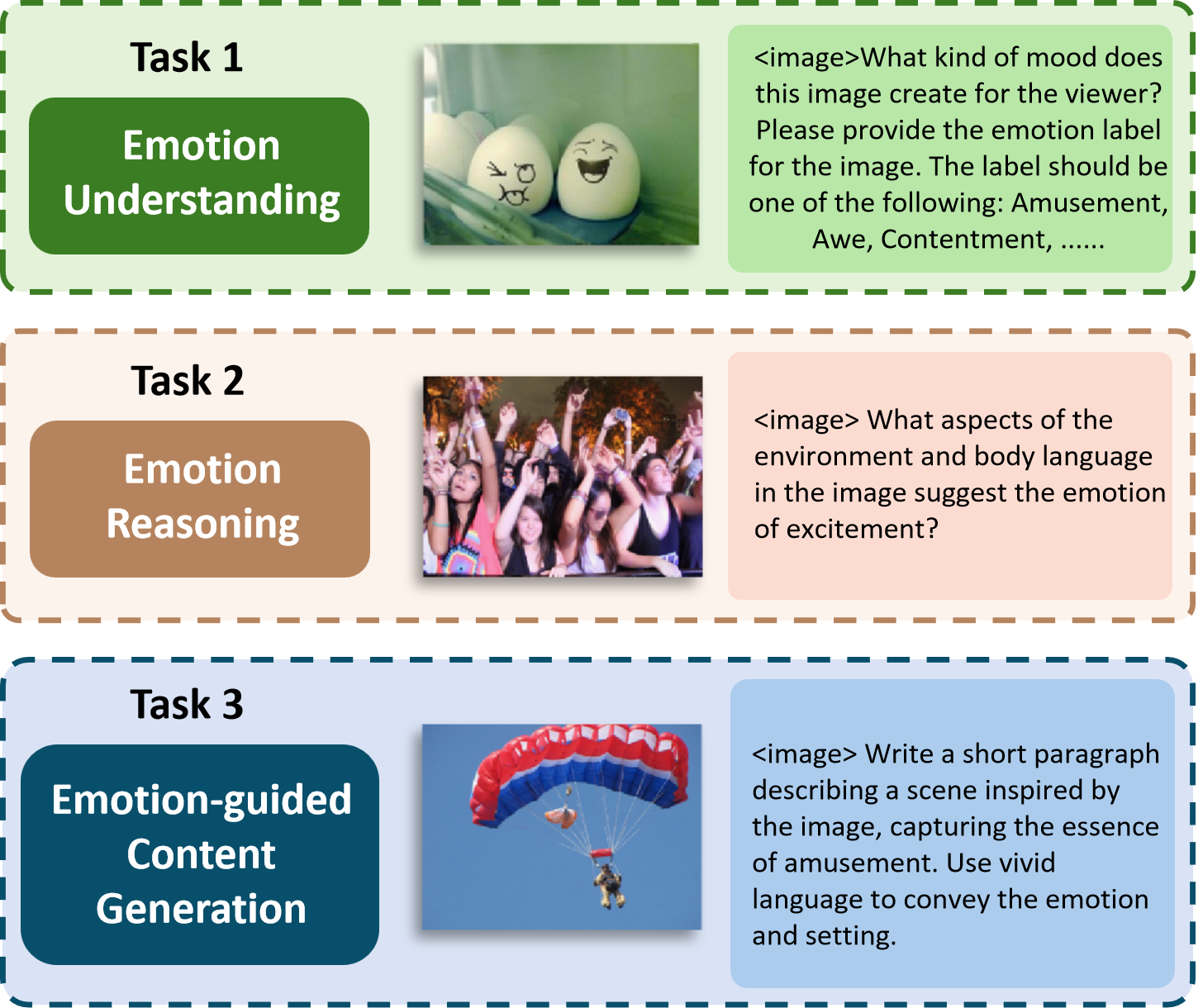}
    \caption{Illustration of the three affective tasks in the AICA-Bench benchmark.}
    \label{fig:aica-vlm-task-demo}
\end{figure}

In response, several recent benchmarks have started to explore the emotional capabilities of VLMs and MLLMs~\cite{yang2024emollmmultimodalemotionalunderstanding, bhattacharyya2025evaluating, pan2025affectgpt, chen2025emobench, song2024mosabench}, as summarized in Table~\ref{tab:emotion-benchmark-comparison}.
Emo-Bench-M~\cite{yang2024emollmmultimodalemotionalunderstanding}
is a benchmark designed to evaluate the emotional intelligence (EI) capabilities of MLLMs.
EVE~\cite{bhattacharyya2025evaluating} introduces an image-based benchmark for emotion recognition, evaluating seven open-source VLMs across classification, grounding, and explanation tasks.
AffectGPT~\cite{pan2025affectgpt} proposes MER-UniBench, a benchmark designed for fine-grained multimodal emotion recognition, with a focus on video and dialogue understanding in MLLMs. 
EEmo-Bench~\cite{chen2025emobench} focuses on image-evoked emotion assessment.
MOSABench~\cite{song2024mosabench} introduces a benchmark for multi-object sentiment analysis in MLLMs.

\begin{table*}[htb]
\centering
\small
% \footnotesize
\renewcommand{\arraystretch}{1.0}
\setlength{\tabcolsep}{3.8pt}
\begin{tabular}{lccccccc}
\hline
Benchmark & Model & Tasks & \#Datasets & \#AICA Datasets & \#Instr. & \#Models & Prompt \\
\hline
EVE~\cite{bhattacharyya2025evaluating} & VLM & EU & 5 & 5 & 8,009 & 7 & B+CoT \\
AffectGPT~\cite{pan2025affectgpt} & MLLM & EU & 9 & 3 & - & 17 & B \\
EEmo-Bench~\cite{chen2025emobench} & MLLM & EU & 1 & 1 & 6,773 & 19 & B \\
EmoBench-M~\cite{hu2025emobenchmbenchmarkingemotionalintelligence} & MLLM & EU & 13 & - & 6,226 & 20 & B \\
MOSABench~\cite{song2024mosabench} & MLLM & EU & 1 & 1 & 1,000 & 8 & B \\
\textbf{AICA-Bench (Ours)} & \textbf{VLM} & \textbf{EU, ER, EGCG} & \textbf{9} & \textbf{9} & \textbf{18,124} & \textbf{23} & \textbf{B+CoT} \\
\hline
\end{tabular}
\caption{
Comparison of recent emotion benchmarks for VLMs and MLLMs. 
\textbf{EU}: Emotion Understanding, \textbf{ER}: Emotion Reasoning, \textbf{EGCG}: Emotion-guided Content Generation. \#Instr. = total evaluation instructions; 
Prompt = \textbf{B}: Basic prompting, \textbf{CoT}: Chain-of-Thought prompting.
}
\label{tab:emotion-benchmark-comparison}
\end{table*}

Those benchmarks have made meaningful progress in evaluating the emotional capabilities of VLMs and MLLMs. However, they have yet to deeply investigate AICA, focusing on how emotions are interpreted, explained, and generated from visual content. As shown in Table~\ref{tab:emotion-benchmark-comparison}, most benchmarks include only a few image-based emotion datasets and focus primarily on basic \textit{emotion understanding} tasks, typically framed as multiple-choice emotion classification. Yet prior work in affective computing and emotion psychology~\cite{picard1997affective, cambria2016affective} highlights that emotional intelligence involves not only recognizing affective cues, but also reasoning about emotional causes and producing contextually appropriate affective expressions.

The lack of comprehensive benchmarks for holistic \underline{A}ffective \underline{I}mage \underline{C}ontent \underline{A}nalysis with VLMs —spanning understanding, reasoning, and generation—is a critical bottleneck in advancing affective intelligence. 
To bridge this gap, we propose \textbf{AICA-Bench} (Sec~\ref{sec:aica}), a holistic benchmark evaluating VLMs across three complementary dimensions: \textbf{Emotion Understanding (EU)}, \textbf{Emotion Reasoning (ER)}, and \textbf{Emotion-Guided Content Generation (EGCG)}. 
AICA-Bench integrates 9 diverse datasets and 18,124 standardized instructions to assess not only cue recognition but also causal explanation and empathetic expression. Using this benchmark, we conduct a comprehensive evaluation of 23 open- and closed-source VLMs under zero-shot settings.

Beyond quantitative analysis (Sec~\ref{sec:exp}), our detailed diagnostic error analysis uncovers two recurrent failure patterns(Sec~\ref{sec:analysis_method}): (1) \textit{Emotion Intensity Hallucination}, where models frequently confuse high-arousal emotions (e.g., \textit{Amusement}) with low-arousal ones (e.g., \textit{Contentment}) due to weak visual grounding; and (2) \textit{Descriptive Shallowness}, where generated responses suffer from generic, template-like content. 
In response, we introduce \textbf{Grounded Affective Tree (GAT)} Prompting (Sec~\ref{sec:analysis_method}). This training-free framework leverages visual scaffolding to steer models toward precise calibration and richer descriptive depth. Across the three tasks, GAT produces consistent improvements, with the EU Task increasing by 6.15 percentage points and the ER and EGCG Tasks increasing by 3.54 and 3.96 percentage points, respectively.

\section{Related Work}

Vision-Language Models (VLMs) have rapidly advanced from early pretraining frameworks such as Flamingo~\cite{Flamingo2022} and BLIP~\cite{li2022blip} to modern instruction-following systems capable of handling diverse multimodal tasks. Recent open-source representatives include the LLaVA family~\cite{liu2023llava, liu2023improvedllava, liu2024llavanext}, the Qwen2-VL and Qwen2.5-VL series~\cite{Qwen2-VL, Qwen2.5-VL}, and InternVL2.5/3~\cite{InternVL2.5, InternVL3}, alongside compact models such as MiniCPM-V~\cite{yao2024minicpm}. Commercial systems such as GPT-4o and Gemini 2.5 further demonstrate the practical utility of this paradigm.

Most existing benchmarks primarily evaluate factual perception, VQA, captioning, and instruction following~\cite{schwenk2022aokvqa, yin2023survey, tu2023how, yue2024mmmu, thapliyal2022crossmodal, liu2023query}, leaving the affective dimension—how models perceive, interpret, and generate emotional content—largely understudied. Recent work has begun probing VLMs’ affective abilities~\cite{yang2024emollmmultimodalemotionalunderstanding, bhattacharyya2025evaluating, pan2025affectgpt, chen2025emobench, song2024mosabench}, but these studies typically rely on limited datasets and focus narrowly on emotion classification, without addressing affective reasoning or emotionally grounded generation.

Motivated by holistic evaluation efforts such as HELM~\cite{liang2023helm} and VHELM~\cite{liang2023holistic}, we introduce \textbf{AICA-Bench}, a holistic benchmark that systematically evaluates VLMs across the full spectrum of affective understanding, reasoning, and generation.
%  It systematically assesses models across a diverse set of affective tasks, including emotion understanding, emotion reasoning, and emotion-conditioned content generation.

\begin{figure*}[ht]
    \centering
    \includegraphics[width=\linewidth]{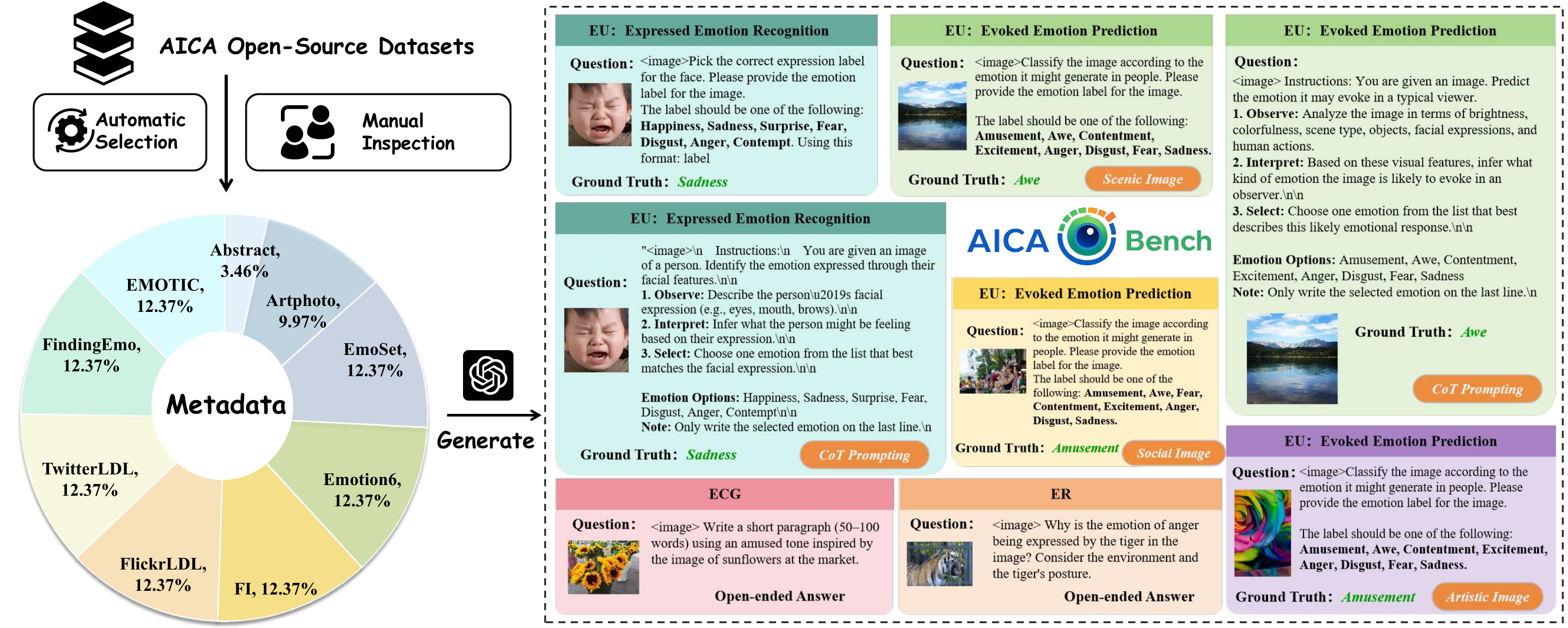}
    \caption{
        The instruction curation pipeline of the AICA-Bench benchmark. 
        It consists of two stages: (1) image sample selection and filtering from multiple affective datasets, and (2) automatic instruction generation across three tasks—emotion understanding (EU), emotion reasoning (ER), and emotion-guided content generation (EGCG).
        }
    \label{fig:aica-vlm-dataset}
\end{figure*}

\section{The AICA-Bench Benchmark}
\label{sec:aica}
\subsection{Evaluation Tasks Design}
As discussed in the Introduction section, the existing affective benchmarks for VLMs focus primarily on \textit{emotional understanding} tasks. However, emotional intelligence extends beyond perception, encompassing the ability to reason about the underlying causes of emotions and generate emotionally aligned content. 
To fill this gap, we propose the AICA-Bench benchmark, which evaluates VLMs on three tasks—emotional understanding(\textbf{EU}), emotion reasoning(\textbf{ER}), and emotion-guided content generation(\textbf{EGCG})—reflecting different capability dimensions of VLMs in "understanding–reasoning–generation" for affective intelligence.
(see Fig.~\ref{fig:aica-vlm-task-demo} for task demonstrations).

\begin{enumerate}
    \item \textbf{EU}: assesses a model’s ability to identify emotions explicitly expressed in an image (\textit{Expressed Emotion Recognition}) 
    and infer the likely emotional response elicited in viewers (\textit{Evoked Emotion Prediction});
    \item \textbf{ER}: measures a model’s ability to explain why an image evokes a particular emotion, 
    requiring causal reasoning grounded in visual context;
    \item \textbf{EGCG}: evaluates a model’s ability to produce emotionally congruent descriptions 
    conditioned on an image and a target emotion.
\end{enumerate}

\subsection{Evaluation Dataset Construction}
\label{sec:aica-vlm-becnhmark-method}
The dataset construction process in AICA-Bench comprises two stages to ensure both representativeness and quality of the evaluation prompts. 
Stage~1 curates high-quality affective image samples from existing public datasets, and Stage~2 leverages GPT-4o to automatically generate structured evaluation instructions for emotion understanding, reasoning, and generation tasks. 
The overall pipeline is illustrated in Figure~\ref{fig:aica-vlm-dataset}.

% In this section, we introduce AICA-Bench, a holistic benchmark for evaluating the affective image content analysis capabilities of VLMs.

% \subsection{Instruction Curation}

\paragraph{Image Collection and Filtering} 
We curate a total of 8,086 affective images as the visual foundation of our benchmark.  
These samples are drawn from a range of publicly available affective computing datasets and undergo a two-step filtering process.  
Candidate images are first selected using automated scripts, after which 20 trained annotators manually review them to remove any content that is not safe for work or emotionally ambiguous.  
Rejected samples are replaced to maintain both emotional clarity and dataset diversity.  
Specifically, we include 1,000 samples each from EmoSet~\cite{emoset2023}, Emotion6~\cite{emotionsix2016}, FI~\cite{fi2020}, FlickrLDL~\cite{flickrldl2016}, TwitterLDL~\cite{flickrldl2016}, FindingEmo~\cite{findingo2019}, and EMOTIC~\cite{kosti-emotic}, along with 806 samples from Artphoto~\cite{artphoto} and 280 from Abstract~\cite{artphoto}.

\paragraph{Evaluation Instruction Generation}
In the second stage, we automatically generate structured evaluation instructions for each pair of curated image-label using GPT-4o. This automated approach enables the consistent and scalable creation of task-specific prompts while ensuring alignment with the evaluation dimensions of the benchmark.

The instructions cover three task categories. \textbf{EU} includes two subtasks: \textit{Expressed Emotion Recognition}, identifying the emotion explicitly displayed by individuals in the image, and \textit{Evoked Emotion Prediction}, inferring the likely emotional response of viewers. Both subtasks use two prompt styles: Basic prompts, which directly query the emotion from a predefined category list, and CoT prompts, which guide the model to reason step by step over visual cues such as color and brightness, contextual elements like scene type and objects, and semantic indicators such as facial expressions before predicting the label. \textbf{ER} provides the image and its emotion label and asks the model to explain why the image evokes that emotion, testing causal reasoning grounded in visual context. \textbf{EGCG} similarly takes an image and target emotion but prompts the model to generate an emotionally congruent scene description, assessing its ability to produce affect-aligned content.

\subsection{Evaluation Strategy}
\label{sec:eval_strategy}
Our evaluation strategy is designed to accommodate the heterogeneous task formats in AICA-Bench, including multiple-choice emotion recognition (EU) and open-ended generation tasks (EGCG and ER).

\textbf{EU: }
For the multiple-choice emotion understanding task, we adopt the weighted F1 score as the primary evaluation metric. 
The weighted F1 better reflects the performance in the unbalanced emotion categories, taking into account both the precision of each class and the recall while weighting each class proportionally to its frequency. 
These two sub-tasks are defined following the taxonomy in~\cite{kosti-emotic}, which distinguishes between \textit{Expressed Emotion Recognition} and \textit{Evoked Emotion Prediction}.

\paragraph{EGCG and ER: }
Both emotion-guided content generation and emotion reasoning involve open-ended outputs where traditional automatic metrics (e.g., BLEU~\cite{papineni-etal-2002-bleu}) are insufficient. These conventional metrics emphasize lexical overlap or global semantic similarity but fail to capture critical affective dimensions.

To address these limitations, we propose an \textbf{AICA-Bench Scoring Model} based on QwenVL2.5-7B, fine-tuned with instruction-style supervision grounded in human-annotated evaluations. Instead of relying solely on synthetic LLM scores, we collect human ratings across multiple affective criteria and use them to guide the learning process. 
Our scoring system evaluates open-ended responses along three criteria: (1)~\textbf{Emotion alignment}—the degree to which the response reflects the target emotion; (2)~\textbf{Descriptiveness}—the richness and specificity of the generated content; and (3)~\textbf{Causal soundness}, which we introduce specifically for the \textit{emotion reasoning} task to assess whether the response provides a plausible explanation linking visual cues to the predicted emotion. These dimensions are inspired by prior work~\cite{buechel2018learning, celikyilmaz2020evaluation} and extended to suit the multimodal and reasoning-oriented nature of our benchmark.

\begin{table}[htb]
\centering
\small
\renewcommand{\arraystretch}{1.2}
\setlength{\tabcolsep}{3pt} % 缩小列间距
\resizebox{\columnwidth}{!}{%
\begin{tabular}{lccc|ccc}
\hline
Model & \multicolumn{3}{c|}{\textbf{ER}} & \multicolumn{3}{c}{\textbf{EGCG}} \\
\hline
 & MSE & MAE & Pearson & MSE & MAE & Pearson \\
\hline
Qwen2.5VL-7B & 1.220 & 0.775 & 0.472 & 0.845 & 0.645 & 0.528 \\
Gemini2.5-Pro & 1.940 & 0.940 & 0.441 & 0.855 & 0.620 & 0.735 \\
ChatGPT-4o    & 1.360 & 0.790 & 0.502 & 0.580 & 0.470 & 0.775 \\
\textbf{Ours} & \textbf{0.400} & \textbf{0.295} & \textbf{0.880} & \textbf{0.260} & \textbf{0.240} & \textbf{0.900} \\
\hline
\end{tabular}%
}
\caption{Performance comparison on ER and EGCG tasks against human annotations.}
\label{tab:evaluation_results}
\end{table}

We construct a dataset of 10,000 open-ended question–answer pairs via GPT‑4o (5,000 for \textit{emotion reasoning} and 5,000 for \textit{emotion-guided content generation}), which is subsequently partitioned into training, validation, and test sets with a ratio of 8:1:1.
Each response is independently annotated by five different annotators randomly selected from a pool of ten, using our proposed affective criteria on a 1 to 5 scale (interannotator agreement measured by Krippendorff’s $\alpha$ is 0.78, indicating substantial agreement suitable for model training). We then fine-tune Qwen2.5VL‑7B with the human-labeled data in a supervised manner, producing the AICA‑Bench scoring model, which demonstrates superior alignment with human judgment compared to baseline and closed-source LLMs (Table~\ref{tab:evaluation_results}).
To facilitate comparison on the leaderboard, we normalize the raw 1--5 ratings ($s$) to a percentage score via $S_{\%} = \frac{s}{5} \times 100$ and report the macro-average across criteria.

\section{Experiments}
\label{sec:exp}

\begin{table*}[t]
\centering
\small
\setlength{\tabcolsep}{6pt}
\renewcommand{\arraystretch}{1.1}
\begin{tabular}{lcccccc}
\hline
\textbf{Model} & EU Basic & EU CoT & EU Avg. & ER Avg. & EG Avg. & \textbf{Overall Avg. (\%)} \\
\hline
\multicolumn{7}{c}{\textit{Closed-source Models}} \\
\hline
Gemini-2.5-pro            & 66.97 & 67.57 & 67.27 & 79.08 & 74.13 & \textbf{73.49} \\
Qwen-VL-max               & 64.07 & 65.98 & 65.02 & 77.75 & 75.93 & \textbf{72.90} \\
ChatGPT-4o                 & 64.44 & 65.42 & 64.93 & 77.81 & 75.73 & \textbf{72.82} \\
Gemini-2.5-flash          & 68.05 & 69.32 & 68.68 & 76.55 & 68.19 & \textbf{71.14} \\
ChatGPT-4o-mini            & 60.15 & 63.68 & 61.91 & 76.45 & 74.09 & \textbf{70.81} \\
Qwen-VL-plus              & 60.04 & 67.81 & 63.92 & 72.39 & 66.86 & \textbf{67.73} \\
Gemini-2.0-flash          & 67.16 & 68.98 & 68.07 & 71.05 & 63.93 & \textbf{67.68} \\
\hline
\multicolumn{7}{c}{\textit{Open-source Models (Size $>$ 6B)}} \\
\hline
Qwen2.5VL-7B~\cite{Qwen2.5-VL}                 & 56.43 & 57.25 & 56.84 & 74.50 & 66.00 & \textbf{65.78} \\
Ovis2-16B~\cite{lu2024ovis}                    & 54.38 & 54.70 & 54.54 & 68.24 & 71.56 & \textbf{64.78} \\
Ovis2-8B~\cite{lu2024ovis}                     & 53.63 & 52.73 & 53.18 & 68.89 & 70.81 & \textbf{64.29} \\
InternVL3-14B~\cite{zhu2025internvl3exploringadvancedtraining} & 52.91 & 52.04 & 52.47 & 68.27 & 66.50 & \textbf{62.41} \\
InternVL3-8B~\cite{zhu2025internvl3exploringadvancedtraining}  & 52.18 & 52.98 & 52.58 & 67.21 & 67.27 & \textbf{62.35} \\
InternVL2.5-8B~\cite{chen2024internvl}         & 51.89 & 51.03 & 51.46 & 66.48 & 68.86 & \textbf{62.27} \\
MiniCPM-o-2.6~\cite{yao2024minicpm}            & 52.73 & 48.65 & 50.69 & 70.16 & 64.98 & \textbf{61.94} \\
Qwen2VL-7B~\cite{Qwen2-VL}                     & 53.52 & 55.19 & 54.36 & 65.23 & 64.76 & \textbf{61.45} \\
LLaVA-1.6-13B~\cite{liu2024llavanext}          & 36.78 & 46.82 & 41.80 & 73.57 & 64.51 & \textbf{59.96} \\
LLaVA-1.6-7B~\cite{liu2024llavanext}           & 36.58 & 50.22 & 43.40 & 73.81 & 59.58 & \textbf{58.93} \\
MiniCPM-V-2.6~\cite{yao2024minicpmvgpt4vlevelmllm} & 43.70 & 47.25 & 45.48 & 65.77 & 63.00 & \textbf{58.08} \\
LLaVA-onevision~\cite{li2024llava}             & 54.02 & 53.25 & 53.64 & 63.78 & 54.18 & \textbf{57.20} \\
\hline
\end{tabular}
\caption{Main results on the AICA-Bench benchmark. \textbf{EU}: Emotion Understanding, \textbf{ER}: Emotion Reasoning, \textbf{EGCG}: Emotion-guided Content Generation.The complete leaderboard is provided in Appendix Table~\ref{tab:more_main_results_simplified}.}
\label{tab:main_results}
\end{table*}

\begin{figure*}[thb]
    \centering
    \includegraphics[width=\linewidth]{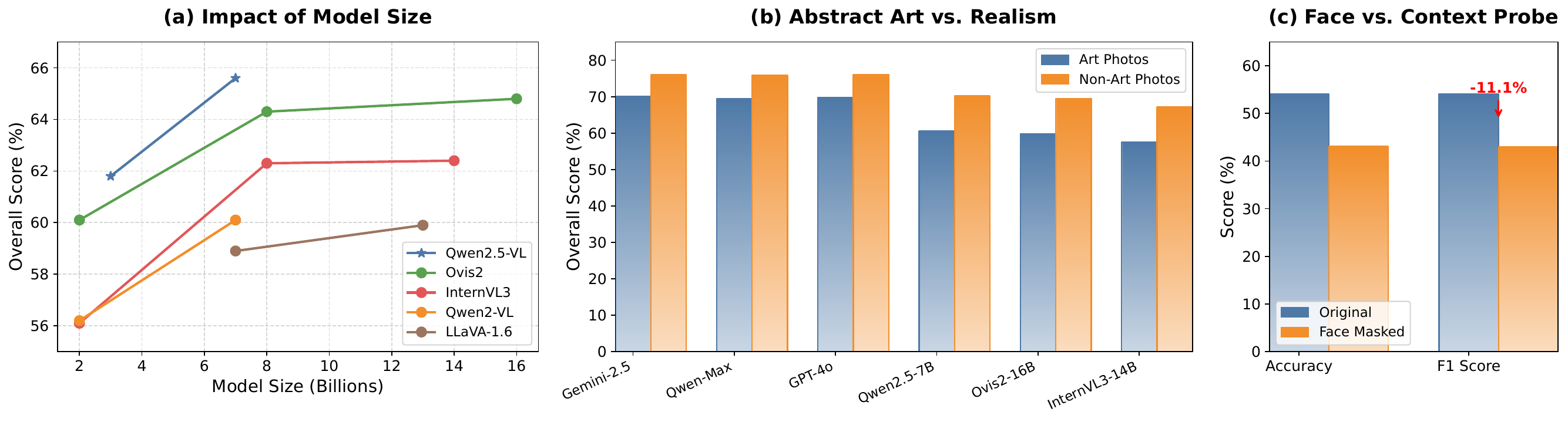}
    \caption{\textbf{(a)} Performance gains from model scaling diminish beyond 7B parameters. \textbf{(b)} Models consistently struggle with abstract art compared to realistic photos. \textbf{(c)} Masking facial cues causes an 11.1\% performance drop, revealing the models' heavy reliance on visual shortcuts (faces) over holistic context.}
    \label{fig:combined-analysis}
\end{figure*}

\subsection{Experimental Setup}
We benchmark a total of 23 VLMs, comprising 7 closed-source commercial models via APIs and 16 open-source models ranging from 2B to 16B parameters. 

\paragraph{Closed-source models.} We evaluate the following commercially hosted VLMs via API access: Gemini-2.5-Pro, Gemini-2.5-Flash, Gemini-2.0-Flash, GPT-4o, GPT-4o-Mini, Qwen-VL-Max, Qwen-VL-Plus.

\paragraph{Open-source models.} We include a diverse set of open-source models covering different architectures and sizes: Qwen series (Qwen2.5VL-7B-Instruct, Qwen2.5-VL-3B-Instruct~\cite{Qwen2.5-VL}, Qwen2-VL-7B-Instruct, Qwen2-VL-2B-Instruct~\cite{Qwen2-VL}), InternVL series (InternVL3-14B-Instruct, InternVL3-8B-Instruct, InternVL3-2B-Instruct~\cite{zhu2025internvl3exploringadvancedtraining}, InternVL2.5-8B~\cite{chen2024internvl}), Ovis series (Ovis2-16B, Ovis2-8B, Ovis2-2B~\cite{lu2024ovis}), LLaVA series (LLaVA-1.6-13B, LLaVA-1.6-7B, LLaVA-OneVision~\cite{li2024llava}), and MiniCPM series (MiniCPM-V-2.6~\cite{yao2024minicpmvgpt4vlevelmllm}, MiniCPM-O-2.6~\cite{yao2024minicpm}).

\paragraph{Computing environment.} All open-source models are evaluated on a cloud computing platform using NVIDIA A100 GPUs. Closed-source models are accessed through their respective APIs under standard inference settings.

\subsection{Main Results and Analysis}

We present the comprehensive evaluation results in Table~\ref{tab:main_results} and visualize key analytical dimensions in Figure~\ref{fig:combined-analysis}. Our analysis highlights four critical observations regarding the current state of multimodal affective intelligence, with additional breakdowns and extended quantitative results provided in Appendix~\ref{appendix:more_analysis}.

\paragraph{The Understanding-Reasoning-Generation Gap.}
Although closed-source models still hold the lead, the gap between them and top-tier open-source models varies significantly across tasks. In reasoning-heavy tasks (ER), this gap has narrowed to just $\sim$4.5\%, but in visual perception (EU), it remains substantial at over 10\%. This reflects a consistent \textbf{"Top-Heavy" pattern} across all evaluated models: scores for Reasoning and Generation are consistently 15-30\% higher than Understanding scores. A clear example is LLaVA-1.6-13B, which achieves high reasoning performance (73.57\%) comparable to state-of-the-art models, yet performs poorly in basic perception (41.80\% in EU). This indicates that current models rely heavily on their language priors to infer emotions rather than truly perceiving them from visual cues. Furthermore, simply increasing model size offers limited benefits (Figure~\ref{fig:combined-analysis}a); for instance, scaling parameters from 8B to 16B in the Ovis2 series yields negligible gains. This collectively suggests that the primary bottleneck is not the model size or textual capabilities, but the quality of fine-grained visual encoding.

\paragraph{Sensitivity to Visual Abstraction.}
We observe a consistent performance degradation when models process Abstract Art Photos compared to Realistic Non-Art Photos, as shown in Figure~\ref{fig:combined-analysis}(b). Real-world photos typically contain clear objects and facial expressions that VLMs are well-trained to recognize. In contrast, art photos convey emotion through abstract features such as color theory and composition. The universal performance drop indicates that current VLMs lack effective affective transfer capability, struggling to generalize from concrete visual objects to abstract representations.

\paragraph{Over-reliance on High-Level Facial Cues.}
Inspired by psychological studies \cite{emoset2023} which posit facial expressions as critical high-level emotion attributes, we investigate the extent to which current VLMs rely on this specific visual cue for EU. While facial expressions undoubtedly influence human emotional perception, a robust VLM should ideally integrate diverse visual information across levels. To rigorously test this, we conducted a controlled ablation study on a diagnostic subset of 500 images sampled from the context-rich EMOTIC and EmoSet datasets. For each image, facial regions were detected and masked with opaque bounding boxes using MediaPipe. Crucially, to ensure experimental validity, we manually verified the masked subset to explicitly retain only samples where the emotional intent remained visually inferable through remaining scene-level or action-level cues. The results, as illustrated in Figure~\ref{fig:combined-analysis}(c), reveal a critical dependency: when high-level facial cues are occluded, the model's performance suffers a sharp decline, with the F1 score dropping by 11.1\%. This significant degradation confirms that current VLMs heavily utilize facial expressions as a primary heuristic, struggling to effectively synthesize alternative contextual information when this visual shortcut is removed.
\section{Diagnostic Analysis and Affective Steering}
\label{sec:analysis_method}

\begin{figure*}[thb]
    \centering
    \includegraphics[width=\linewidth]{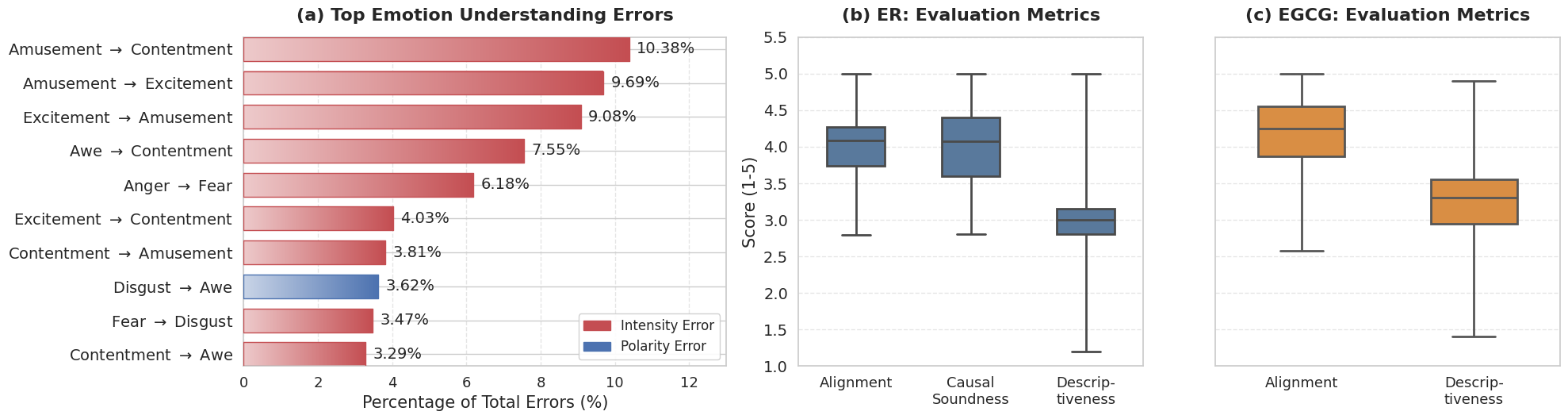}
    \caption{\textbf{(a) EU:} The dominance of intensity errors (Blue) over valence errors (Red) reveals an arousal bottleneck. \textbf{(b)-(c) ER \& EGCG:} Across both tasks, models achieve high emotion alignment but consistently lack descriptive depth.}
    \label{fig:combined_error}
\end{figure*}

\begin{figure*}[tb]
    \centering
    \captionsetup{skip=2pt}
    \includegraphics[width=\linewidth]{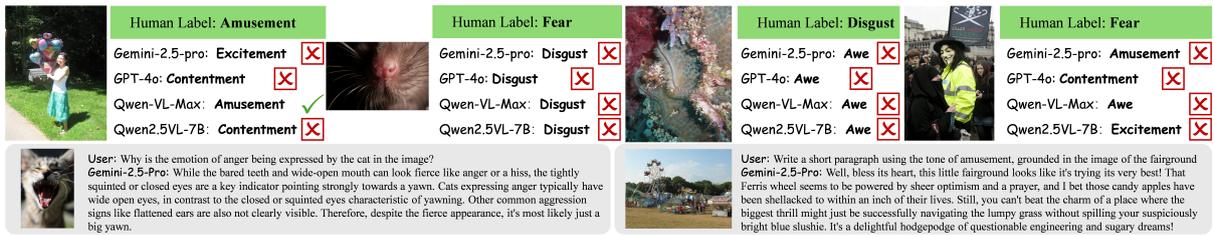}
    \caption{Representative failure cases of EU, ER, and EGCG tasks.}
    \label{fig:error-analysis-showcase}
\end{figure*}

While Section~\ref{sec:exp} establishes the performance benchmarks, it does not reveal the underlying cognitive mechanisms behind model failures. In this section, we first conduct a diagnostic error analysis to pinpoint the limitations of current VLMs. Based on the identified bottlenecks—specifically the lack of visual grounding and the confusion in arousal estimation—we introduce a theoretically grounded framework, the Grounded Affective Tree (GAT) Prompting. This approach integrates visual scaffolding with hierarchical reasoning to steer models towards robust affective alignment and descriptive depth across all three tasks.

\subsection{Diagnostic Error Analysis}
\label{subsec:error_analysis}

\paragraph{EU Misclassification Patterns: Intensity vs. Polarity.}
A comprehensive evaluation of total errors reveals that misclassifications are disproportionately driven by \textbf{Intensity Errors} ($72.25\%$), significantly overshadowing \textbf{Polarity Errors} ($27.75\%$). This fundamental split indicates that while models generally succeed in distinguishing positive from negative affective states, they struggle severely with gauging arousal magnitude from visual cues. This trend is visually reinforced in Figure~\ref{fig:combined_error}a, where the top ten misclassification patterns are dominated by red gradient bars representing within-polarity confusion, such as mislabeling \textit{Amusement} as lower-arousal \textit{Contentment} ($10.38\%$) or higher-arousal \textit{Excitement} ($9.69\%$). Notably, the presence of a blue gradient bar signifies critical cross-valence failures, exemplified by the severe confusion of negative \textit{Disgust} with positive \textit{Awe} ($3.62\%$). This demonstrates that despite the lower overall frequency of polarity errors in the top tier, models remain susceptible to catastrophic valence flips, particularly when faced with high-arousal visual features.

\paragraph{Descriptive Shallowness in Reasoning \& Generation.}
For open-ended tasks, Figure~\ref{fig:combined_error}(b)-(c) reveals a consistent dichotomy across both ER and EGCG: while models consistently achieve high scores in \textit{Emotion Alignment} (Median $\approx$ 4.1), indicating effective topic adherence, their \textit{Descriptiveness} lags significantly (Median $\approx$ 3.0). 
This performance gap highlights a "Safe Response Trap." Figure~\ref{fig:error-analysis-showcase} qualitatively confirms this pattern, illustrating how models often resort to ``template-filling''—producing generic, safe descriptions rather than grounding their reasoning in specific, subtle visual evidence like lighting, texture, or social interactions.

% =================================================================================
% Section 5.2: Steering via Grounded Affective Tree (GAT) Prompting
% =================================================================================

\subsection{Steering via Grounded Affective Tree Prompting}
\label{subsec:gat_method}

To address the \textit{Intensity} and \textit{Polarity Errors} alongside the \textit{Descriptive Shallowness} identified in Section~\ref{subsec:error_analysis}, we introduce \textbf{Grounded Affective Tree (GAT) Prompting}.

\paragraph{Why Training-free-based Steering?} First, \textit{Efficiency}: simple instructions have been proven to steer latent capabilities effectively without altering weights \citep{wu2025axbench}. Second, \textit{Scalability}: prompting facilitates test-time scaling, where structured strategies theoretically enhance reasoning performance \citep{liu-etal-2025-rethinking}. Third, \textit{Universality}: unlike fine-tuning, prompting serves as a training-free, model-agnostic solution applicable to both open-source and black-box models.
Training-free-based

% \paragraph{Why Training-free-based Steering?} We adopt this paradigm for three reasons: (1) \textit{Efficiency}: it steers latent capabilities without the cost of parameter updates \citep{wu2025axbench}; (2) \textit{Scalability}: it facilitates test-time scaling, where structured reasoning enhances performance by allocating inference compute \citep{liu-etal-2025-rethinking}; and (3) \textit{Universality}: it offers a model-agnostic solution compatible with both open-source and proprietary black-box systems.

\subsubsection{GAT Prompting}

Drawing inspiration from recent work on scaffolding coordinates for vision-language coordination \cite{lei-etal-2025-scaffolding} and the Tree-of-Thoughts (ToT) reasoning framework\cite{yao2023tree}, our proposed GAT Prompting orchestrates a structured cognitive process.

\begin{figure}[htb]
    \centering
    \includegraphics[width=\linewidth]{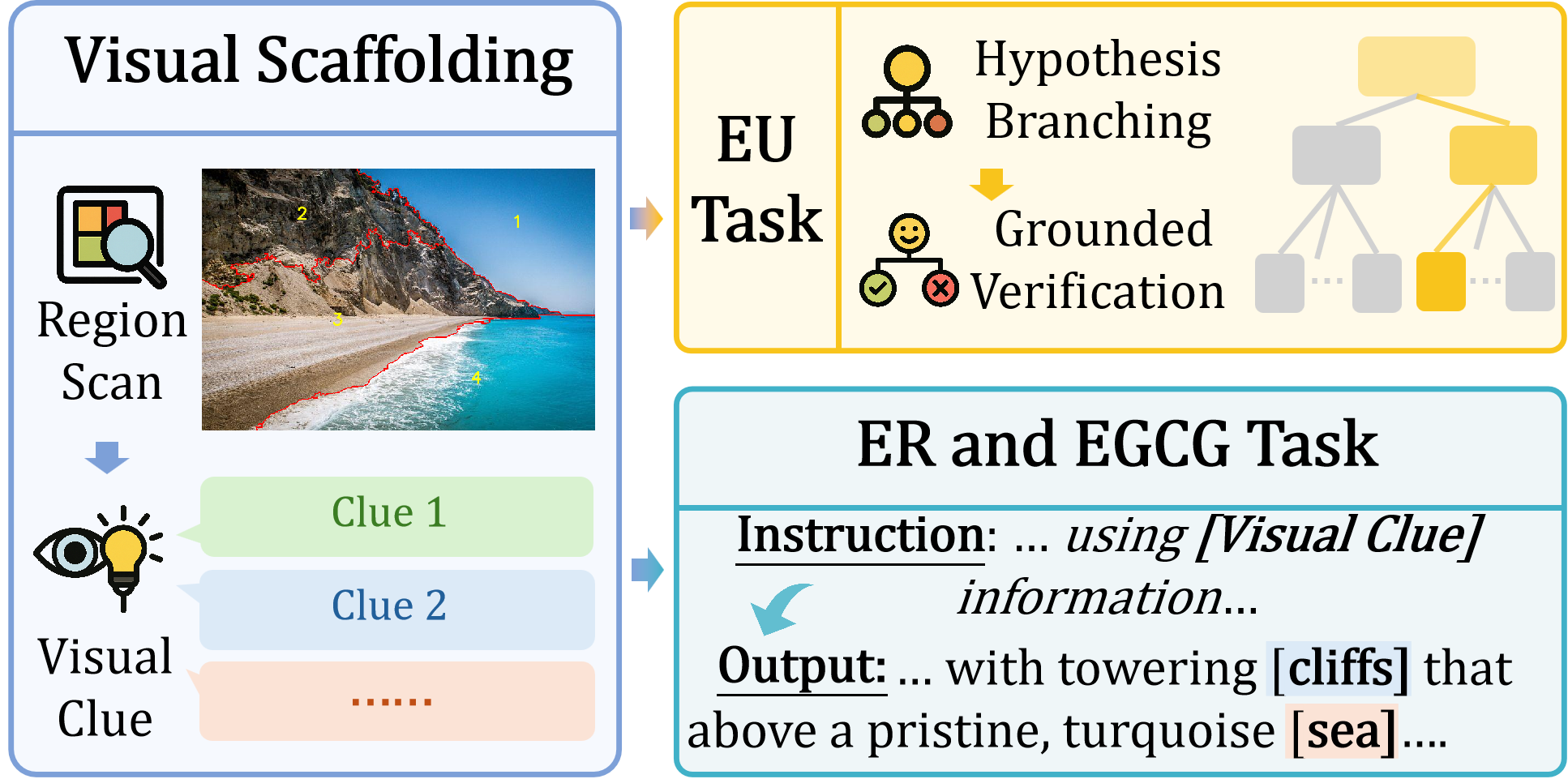}
    \caption{The GAT Prompting Framework.}
    \label{fig:gat}
\end{figure}

\paragraph{Visual Scaffolding.}
The visual scaffolding is generated using an efficient graph-based image segmentation method\cite{felzenszwalb_segmentation_github}, which creates large, contiguous regions that serve as explicit visual anchors in the prompt to guide the VLM's attention. (Please refer to Appendix~\ref{app:visual_scaffolding} for illustrative examples). Based on these segmented anchors, the prompt instructs the VLM to execute a systematic "region scan." For each designated region, the model is required to identify and list objective visual elements, thereby extracting distinct visual clues grounded in specific areas of the image.

\paragraph{AffectToT Reasoning (for EU).}
To address classification ambiguity in the EU task, we introduce AffectToT, a structured cognitive appraisal framework adapted from the ToT paradigm with a fixed search depth ($d=3$) and breadth ($k=3$). Leveraging the objective region descriptions from the initial scan (Step 1) as the foundation, the workflow proceeds to a \emph{Hypothesis Branching} phase (Step 2), where the model generates $k=3$ distinct competing emotion-intensity pairs, explicitly citing specific region IDs as grounding evidence. Subsequently, a \emph{Grounded Verification} stage (Step 3) functions as a critic to evaluate the logical consistency between each proposed intensity and the visual facts. By identifying contradictions—such as a relaxed posture invalidating a high-arousal hypothesis—the system actively prunes invalid branches. This single-pass propose-and-verify mechanism effectively eliminates intensity hallucinations within a constrained token budget, ensuring the final decision is strictly grounded in verified visual evidence.

\paragraph{Grounded Generation (for ER \& EGCG).} In parallel, for open-ended generation tasks—ER and EGCG—the strategy pivots to directly leveraging the objective foundation established by visual scaffolding to enhance descriptive fidelity.

\subsubsection{Effectiveness of GAT Prompting}
\label{subsec:gat_effectiveness}

\paragraph{Impact on Task-Level Performance.}
Table~\ref{tab:gat_improvement_on_tasks} demonstrates that GAT yields consistent performance improvements across EU, ER, and EGCG for all evaluated models. Specifically, the framework significantly boosts accuracy in EU while bridging the performance gap in higher-level ER and EGCG tasks, enabling smaller open-source models to approach the capabilities of proprietary baselines. These results confirm GAT as a robust enhancer that effectively grounds abstract emotional concepts in concrete visual evidence to benefit both perception and generation.

\begin{table}[thbp]
\centering
\scriptsize 
\setlength{\tabcolsep}{5.0pt} 
\renewcommand{\arraystretch}{0.85}
\caption{Comprehensive performance comparison across 15 models. We compare Baseline and GAT for EU, ER, and EGCG, where \textbf{\textcolor[rgb]{0.0, 0.6, 0.0}{green}} values indicate improvements of GAT over the Baseline.}
\label{tab:gat_improvement_on_tasks}
\begin{tabular}{l|cc|cc|cc}
\toprule
\multirow{2}{*}{\textbf{Model}} & \multicolumn{2}{c|}{\textbf{EU}} & \multicolumn{2}{c|}{\textbf{ER}} & \multicolumn{2}{c}{\textbf{EGCG}} \\
\cmidrule(lr){2-3} \cmidrule(lr){4-5} \cmidrule(lr){6-7}
 & Basic & \textbf{GAT} & Basic & \textbf{GAT} & Basic & \textbf{GAT} \\
\midrule
Gemini-2.5-Pro   & 66.97 & \textbf{\textcolor[rgb]{0.0, 0.6, 0.0}{71.15}} & 79.08 & \textbf{\textcolor[rgb]{0.0, 0.6, 0.0}{82.45}} & 74.13 & \textbf{\textcolor[rgb]{0.0, 0.6, 0.0}{78.25}} \\
Gemini-2.5-Flash & 68.05 & \textbf{\textcolor[rgb]{0.0, 0.6, 0.0}{70.41}} & 76.55 & \textbf{\textcolor[rgb]{0.0, 0.6, 0.0}{79.02}} & 68.19 & \textbf{\textcolor[rgb]{0.0, 0.6, 0.0}{72.07}} \\
ChatGPT-4o       & 64.44 & \textbf{\textcolor[rgb]{0.0, 0.6, 0.0}{69.82}} & 77.81 & \textbf{\textcolor[rgb]{0.0, 0.6, 0.0}{83.10}} & 75.73 & \textbf{\textcolor[rgb]{0.0, 0.6, 0.0}{79.40}} \\
GPT-4o-mini      & 60.15 & \textbf{\textcolor[rgb]{0.0, 0.6, 0.0}{66.02}} & 76.45 & \textbf{\textcolor[rgb]{0.0, 0.6, 0.0}{77.92}} & 74.09 & \textbf{\textcolor[rgb]{0.0, 0.6, 0.0}{77.54}} \\
Qwen-VL-max      & 64.07 & \textbf{\textcolor[rgb]{0.0, 0.6, 0.0}{69.34}} & 77.75 & \textbf{\textcolor[rgb]{0.0, 0.6, 0.0}{79.18}} & 75.93 & \textbf{\textcolor[rgb]{0.0, 0.6, 0.0}{79.66}} \\
Qwen-VL-plus     & 60.04 & \textbf{\textcolor[rgb]{0.0, 0.6, 0.0}{65.41}} & 72.39 & \textbf{\textcolor[rgb]{0.0, 0.6, 0.0}{76.31}} & 66.86 & \textbf{\textcolor[rgb]{0.0, 0.6, 0.0}{70.12}} \\
\midrule
Qwen2.5-VL-7B    & 56.43 & \textbf{\textcolor[rgb]{0.0, 0.6, 0.0}{61.94}} & 74.50 & \textbf{\textcolor[rgb]{0.0, 0.6, 0.0}{78.92}} & 66.00 & \textbf{\textcolor[rgb]{0.0, 0.6, 0.0}{70.55}} \\
Qwen2-VL-7B      & 53.52 & \textbf{\textcolor[rgb]{0.0, 0.6, 0.0}{59.08}} & 65.23 & \textbf{\textcolor[rgb]{0.0, 0.6, 0.0}{69.15}} & 64.76 & \textbf{\textcolor[rgb]{0.0, 0.6, 0.0}{68.30}} \\
Ovis2-16B        & 54.38 & \textbf{\textcolor[rgb]{0.0, 0.6, 0.0}{58.06}} & 68.24 & \textbf{\textcolor[rgb]{0.0, 0.6, 0.0}{71.05}} & 71.56 & \textbf{\textcolor[rgb]{0.0, 0.6, 0.0}{74.12}} \\
Ovis2-8B         & 53.63 & \textbf{\textcolor[rgb]{0.0, 0.6, 0.0}{57.41}} & 68.89 & \textbf{\textcolor[rgb]{0.0, 0.6, 0.0}{71.50}} & 70.81 & \textbf{\textcolor[rgb]{0.0, 0.6, 0.0}{73.20}} \\
InternVL3-14B    & 52.91 & \textbf{\textcolor[rgb]{0.0, 0.6, 0.0}{57.02}} & 68.27 & \textbf{\textcolor[rgb]{0.0, 0.6, 0.0}{71.46}} & 66.50 & \textbf{\textcolor[rgb]{0.0, 0.6, 0.0}{70.21}} \\
InternVL3-8B     & 52.18 & \textbf{\textcolor[rgb]{0.0, 0.6, 0.0}{56.33}} & 67.21 & \textbf{\textcolor[rgb]{0.0, 0.6, 0.0}{70.30}} & 67.27 & \textbf{\textcolor[rgb]{0.0, 0.6, 0.0}{70.85}} \\
MiniCPM-o-2.6    & 52.73 & \textbf{\textcolor[rgb]{0.0, 0.6, 0.0}{57.63}} & 70.16 & \textbf{\textcolor[rgb]{0.0, 0.6, 0.0}{73.24}} & 64.98 & \textbf{\textcolor[rgb]{0.0, 0.6, 0.0}{68.56}} \\
LLaVA-1.6-13B    & 36.78 & \textbf{\textcolor[rgb]{0.0, 0.6, 0.0}{47.60}} & 73.57 & \textbf{\textcolor[rgb]{0.0, 0.6, 0.0}{75.71}} & 64.51 & \textbf{\textcolor[rgb]{0.0, 0.6, 0.0}{67.29}} \\
LLaVA-1.6-7B     & 36.58 & \textbf{\textcolor[rgb]{0.0, 0.6, 0.0}{51.05}} & 73.81 & \textbf{\textcolor[rgb]{0.0, 0.6, 0.0}{75.88}} & 59.58 & \textbf{\textcolor[rgb]{0.0, 0.6, 0.0}{63.45}} \\
\bottomrule
\end{tabular}
\end{table}

\begin{figure}[thb]
    \centering
    \includegraphics[width=\linewidth]{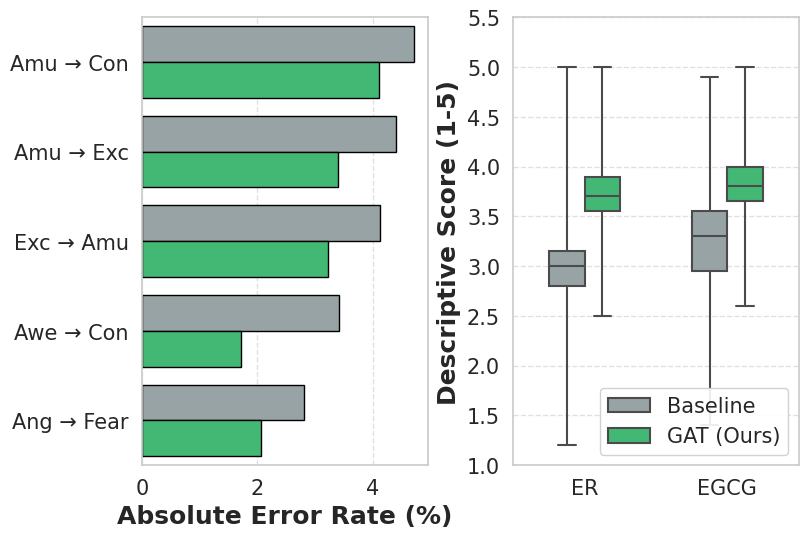}
   \caption{GAT corrects major intensity-confusion errors in EU (left) and improves descriptive depth for ER and EGCG (right). Abbreviations: Amu = Amusement, Con = Contentment, Exc = Excitement, Awe = Awe, Ang = Anger, Fear = Fear.}
    \label{fig:gat_improvement}
\end{figure}

\vspace{-10pt}

\paragraph{Improvements in Error Calibration and Descriptive Quality.}
Figure~\ref{fig:gat_improvement} illustrates the specific improvements in error handling and content quality. The left panel shows that GAT acts as an effective \textit{intensity calibrator}, significantly suppressing confusion between high- and low-arousal emotions by correcting visual hallucinations. The right panel highlights a structural improvement in \textit{descriptive depth} for ER and EGCG. Unlike baselines that tend to produce shallow responses, GAT shifts the score distribution upwards with condensed variance. This confirms that our method effectively eliminates generic outputs, ensuring that the generated content is descriptively rich and contextually grounded.

\begin{table}[thbp]
\centering
\scriptsize
\setlength{\tabcolsep}{4.0pt}
\renewcommand{\arraystretch}{0.88}
\caption{Component-wise ablation of GAT on four representative models. VS denotes visual scaffolding. AT denotes the AffectToT reasoning module for EU. Clue denotes the grounded clue construction module for ER and EGCG. \textbf{\textcolor[rgb]{0.0, 0.6, 0.0}{Green}} values indicate the best result within each task block.}
\label{tab:gat_ablation}
\begin{tabular}{lcccc}
\toprule
\textbf{Method} & \shortstack[c]{\textbf{Gemini-}\\\textbf{2.5-Pro}} & \shortstack[c]{\textbf{ChatGPT-}\\\textbf{4o}} & \shortstack[c]{\textbf{Qwen2.5-}\\\textbf{VL-7B}} & \shortstack[c]{\textbf{Qwen2-}\\\textbf{VL-7B}} \\
\midrule
\multicolumn{5}{c}{\textbf{EU}} \\
\midrule
Baseline        & 66.97 & 64.44 & 56.43 & 53.52 \\
w/ only VS      & 68.40 & 66.55 & 58.50 & 55.60 \\
w/ only AT      & 69.95 & 68.15 & 60.10 & 57.40 \\
w/ VS+AT (GAT)  & \textbf{\textcolor[rgb]{0.0, 0.6, 0.0}{71.15}} & \textbf{\textcolor[rgb]{0.0, 0.6, 0.0}{69.82}} & \textbf{\textcolor[rgb]{0.0, 0.6, 0.0}{61.94}} & \textbf{\textcolor[rgb]{0.0, 0.6, 0.0}{59.08}} \\
\midrule
\multicolumn{5}{c}{\textbf{ER}} \\
\midrule
Baseline          & 79.08 & 77.81 & 74.50 & 65.23 \\
w/ only VS        & 80.60 & 80.20 & 76.40 & 66.85 \\
w/ VS+Clue (GAT)  & \textbf{\textcolor[rgb]{0.0, 0.6, 0.0}{82.45}} & \textbf{\textcolor[rgb]{0.0, 0.6, 0.0}{83.10}} & \textbf{\textcolor[rgb]{0.0, 0.6, 0.0}{78.92}} & \textbf{\textcolor[rgb]{0.0, 0.6, 0.0}{69.15}} \\
\midrule
\multicolumn{5}{c}{\textbf{EGCG}} \\
\midrule
Baseline          & 74.13 & 75.73 & 66.00 & 64.76 \\
w/ only VS        & 76.10 & 77.50 & 68.10 & 66.00 \\
w/ VS+Clue (GAT)  & \textbf{\textcolor[rgb]{0.0, 0.6, 0.0}{78.25}} & \textbf{\textcolor[rgb]{0.0, 0.6, 0.0}{79.40}} & \textbf{\textcolor[rgb]{0.0, 0.6, 0.0}{70.55}} & \textbf{\textcolor[rgb]{0.0, 0.6, 0.0}{68.30}} \\
\bottomrule
\end{tabular}
\end{table}

\paragraph{Component-wise Ablation of GAT.}
Since GAT is compositional, we further ablate its major components on four representative models: Gemini-2.5-Pro, ChatGPT-4o, Qwen2.5-VL-7B, and Qwen2-VL-7B. For EU, we compare \textit{Baseline}, \textit{VS-only}, \textit{AT-only}, and \textit{VS+AT}. For ER and EGCG, we compare \textit{Baseline}, \textit{VS-only}, and \textit{VS+Clue}, where the clue construction step is defined on top of VS. As shown in Table~\ref{tab:gat_ablation}, both VS and task-specific affective steering improve over the baseline, while the full pipeline consistently achieves the best results across all models and tasks. For EU, AT contributes more strongly than VS, indicating that structured affective reasoning is the main driver for resolving emotion ambiguity. For ER and EGCG, VS already yields stable gains, and adding Clue further improves performance, showing that visual grounding and task-specific steering are complementary.

\section{Conclusion}
\label{sec:conclusion}

We introduced AICA-Bench, a holistic benchmark evaluating affective intelligence in vision-language models through perception, reasoning, and generation tasks. Our study reveals weaknesses in VLMs—particularly intensity miscalibration and shallow descriptive grounding—and demonstrates that GAT Prompting, a lightweight test-time strategy, substantially mitigates these issues across models. While establishing a foundation for multimodal affective understanding, broader challenges remain. We hope this advances research toward VLMs that understand, reason, and generate emotional content more accurately and naturally.

\newpage
\section*{Limitations}
\label{sec:limitations}

While AICA-Bench provides a holistic and much-needed perspective on evaluating multimodal affective intelligence, it represents an initial step rather than a complete solution. However, several challenges remain intrinsic to this domain and are not fully addressed in our work. Emotional interpretation is inherently subjective and culturally situated, making it difficult to construct a benchmark that captures the full spectrum of human affect across demographics or social contexts. Our tasks operate on static images and English prompts, leaving broader settings—such as cross-cultural affective reasoning, temporal emotion dynamics, or multilingual understanding—outside our current scope. In addition, although our assessor offers consistent scoring across open-ended tasks, any automated judgment of emotional nuance inevitably carries modeling biases and may not fully approximate human interpretation. Finally, GAT Prompting focuses on test-time steering; its interaction with fine-tuning, memory-based reasoning, or personalized affect modeling remains unexplored. These limitations open important opportunities for future work toward developing more culturally aware, temporally grounded, and human-aligned affective VLMs.

\bibliography{custom}

\appendix

\newpage
% \section{Appendix}

% \section{Emotion Models}

% In this benchmark, we use three widely adopted categorical emotion models, each corresponding to the labeling scheme used in different datasets:

% \begin{itemize}
%   \item \textbf{Ekman model}: A compact taxonomy consisting of six basic emotions: \textit{anger, disgust, fear, joy, sadness, surprise}.

%   \item \textbf{Mikels model}: Defines eight emotions with a mix of positive and negative affective states: \textit{amusement, awe, contentment, excitement, anger, disgust, fear, sadness}.

%   \item \textbf{Plutchik model}: A fine-grained classification system capturing 26 emotional categories, including both basic and nuanced states such as \textit{affection, annoyance, anticipation, aversion, embarrassment, excitement, fatigue, happiness, peace, sadness, suffering, surprise, yearning}, and others.
% \end{itemize}

\section{More Details of AICA-Bench}

\subsection{Datasets}
To construct a comprehensive and diverse affective image benchmark, we select 9 publicly available datasets spanning various sources, labeling models, and visual styles. These datasets include both social and artistic domains, and cover a range of emotion annotation schemes, such as Ekman's basic emotions, the Mikels model, and Plutchik’s wheel. Table~\ref{sup:tab:datasets-overview} summarizes the key statistics and features of each dataset used in AICA-Bench.

\begin{table*}[h]
\centering
\small
\caption{Datasets Used in the AICA Benchmark}
\resizebox{\textwidth}{!}{
\begin{tabular}{@{} lccccc c @{}}
\toprule
\textbf{Dataset} & \textbf{Year} & \textbf{Original Image Count} & \textbf{Images Used in AICA} & \textbf{Proportion in AICA} & \textbf{Type} & \textbf{Emotion Label Model} \\
\midrule
EMOTIC & 2019 & 23,571 & 1,000 & 12.37\% & social & Ekman \\
EmoSet-118K & 2023 & 118,102 & 1,000 & 12.37\% & social & Mikels \\
FindingEmo & 2024 & 25,000 & 1,000 & 12.37\% & social & Plutchik \\
FI & 2016 & 23,308 & 1,000 & 12.37\% & social & Mikels \\
Emotion6 & 2015 & 1,980 & 1,000 & 12.37\% & social & Ekman \\
FlickrLDL & 2017 & 10,700 & 1,000 & 12.37\% & social & Mikels \\
TwitterLDL & 2017 & 10,045 & 1,000 & 12.37\% & social & Mikels \\
Abstract & 2010 & 280 & 280 & 3.46\% & abstract & Mikels \\
ArtPhoto & 2010 & 806 & 806 & 9.97\% & artistic & Mikels \\
\bottomrule
\end{tabular}
}
\label{sup:tab:datasets-overview}
\end{table*}

\begin{itemize}
  \item \textbf{EMOTIC\cite{emotic2016}}: Real-world images featuring individuals in diverse everyday contexts, annotated with Ekman’s basic emotion categories. The dataset is designed to support \textit{Expressed Emotion Recognition} based on visual cues within natural scenes.
  \item \textbf{EmoSet-118K\cite{emoset2023}}: A subset of a large-scale dataset with human-annotated labels, designed primarily for \textit{Evoked Emotion Prediction}. It provides rich multi-level emotion attributes, including low-level (e.g., color, brightness), mid-level (e.g., scene type, object category), and high-level (e.g., facial expression, human action) features, following the Mikels emotion model.
  \item \textbf{FindingEmo\cite{findingo2019}}:  A dataset of 25,000 images tailored for \textit{Expressed Emotion Recognition}, where each image is annotated with Plutchik emotion labels.
  \item \textbf{FI\cite{fi2020}}: A large-scale dataset from Flickr and Instagram, annotated using the Mikels model. Contains diverse everyday scenes and subtle affective cues in social media imagery.
  \item \textbf{Emotion6\cite{emotionsix2016}}: Small-scale dataset labeled with six Ekman emotions plus neutral, offering clean and balanced data for \textit{Evoked Emotion Prediction}.
  \item \textbf{FlickrLDL\cite{flickrldl2016}}: Applies label distribution learning (LDL) on images from Flickr using the Mikels model, capturing the uncertainty of emotion perception through soft labels.
  \item \textbf{TwitterLDL\cite{flickrldl2016}}: Similar to FlickrLDL but sourced from Twitter. Annotated via LDL under the Mikels model, reflecting noisy and ambiguous affective signals in social content.
  \item \textbf{Abstract\cite{artphoto}}: Consists of abstract paintings labeled with the Mikels model. Emotions are conveyed through shape, color, and form rather than representational content.
  \item \textbf{ArtPhoto\cite{artphoto}}: Aesthetic photographic images labeled with the Mikels model. Explores affect elicited through compositional and stylistic elements in artistic photos.
\end{itemize}

\subsection{Instructions}
To support diverse affective tasks, we construct four types of natural language instructions within the AICA-Bench benchmark: \textbf{EU Basic}, \textbf{EU CoT}, \textbf{ER}, and \textbf{EGCG}. In total, we generate \textbf{18,124} instructions covering a wide range of image-emotion combinations across all task types.
Each instruction type corresponds to a specific task category. \textbf{EU Basic} and \textbf{EU CoT} are designed for Emotion Understanding. \textbf{ER} (Emotion Reasoning) instructions require the model to explain the cause of a given emotion in an image. \textbf{EGCG} (Emotion-guided Content Generation) prompts the model to generate emotionally aligned descriptions based on both visual input and a target emotion. Together, these instructions enable fine-grained evaluation of a model’s ability to comprehend, reason about, and generate emotionally meaningful content.

\begin{tcolorbox}[colback=gray!5, colframe=gray!50!black, title=Example Prefix of Instructions templates for Expressed Emotion Recognition in EU Basic, fonttitle=\bfseries, boxrule=0.5pt]
\begin{itemize}
  \item "Identify the emotion displayed by the person in the image."
  \item "What emotion is the person in the picture showing?"
  \item "Determine the person’s emotional expression in this photo."
  \item "Classify the emotion of the person visible in the image."
  \item "From the visual cues, what emotion is the person exhibiting?"
  \item "..."
\end{itemize}
\end{tcolorbox}

\begin{tcolorbox}[colback=gray!5, colframe=gray!50!black, title=Example Prefix of Instructions templates for Evoked Emotion Prediction in EU Basic, fonttitle=\bfseries, boxrule=0.5pt]
\begin{itemize}
  \item "Identify the emotion this image is likely to evoke in a human viewer."
  \item "Based on the visual content, what emotion would a typical observer feel?"
  \item "Determine the emotional response this image is meant to trigger."
  \item "Classify the feeling this image might evoke in someone who sees it."
  % \item "What kind of emotion does this image provoke in a human observer?"
  \item "..."
\end{itemize}
\end{tcolorbox}

For the \textbf{EU Basic} task, we construct the instruction set to directly prompt models to identify the most appropriate emotion from a predefined list. This task is divided into two subtypes: \textit{Expressed Emotion Recognition} and \textit{Evoked Emotion Prediction}. For each subtype, we use GPT-4o to generate 50 diverse question templates phrased in natural language, in order to enhance lexical variety and reduce model overfitting to fixed instruction patterns. These templates serve as natural language \textit{prefixes}, generalizable across datasets and capturing variation in phrasing, tone, and focus.

To create a complete instruction for a given image instance, we concatenate a selected prefix with the emotion label space corresponding to the dataset’s annotation schema (e.g., Ekman, Mikels, or Plutchik). Formally, we define a complete \textbf{EU Basic instruction} as:

\begin{equation}
\text{Instruction} = \text{Prefix}_i + \text{OptionList}(D_j)
\end{equation}

where $\text{Prefix}_i$ is the $i$-th natural language template, and $\text{OptionList}(D_j)$ denotes the set of emotion labels associated with dataset $D_j$. This dynamic construction ensures alignment between the instruction and the emotion taxonomy used in each dataset.

For the \textbf{EU CoT} setting, we design chain-of-thought style instructions to guide the model through a structured three-step reasoning process: \textit{Observe}, \textit{Interpret}, and \textit{Select}. This format encourages the model to reflect on visual features before making a decision, aiming to improve both accuracy and interpretability.

We define two CoT templates corresponding to the two subtypes of Emotion Understanding:

\begin{itemize}
  \item \textit{Evoked Emotion Prediction}:
  The model is instructed to assess how an image may affect a typical viewer. It analyzes the scene based on brightness, colorfulness, scene type, and social context, then selects an emotion label from the provided list.

  \item \textit{Expressed Emotion Recognition}:
  The model is asked to reason about the emotional state of a person depicted in the image by examining facial expressions, body posture, and environmental cues.
\end{itemize}

\begin{figure*}[htb]
    \centering
    % \captionsetup{skip=2pt}
    \includegraphics[width=\linewidth]{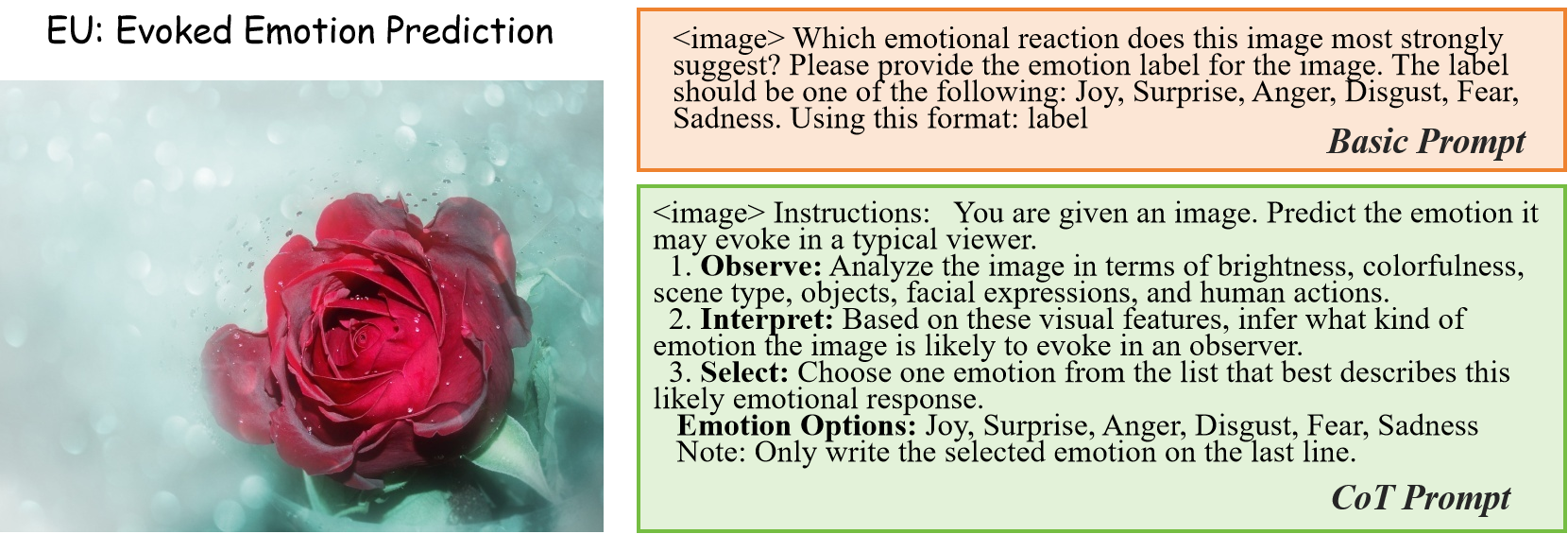}
    \caption{Example of an Evoked Emotion Prediction instruction in the EU Basic and CoT setting.}
    \label{sup:fig:eu-eep-example}
\end{figure*}

\begin{figure*}[htb]
    \centering
    % \captionsetup{skip=2pt}
    \includegraphics[width=\linewidth]{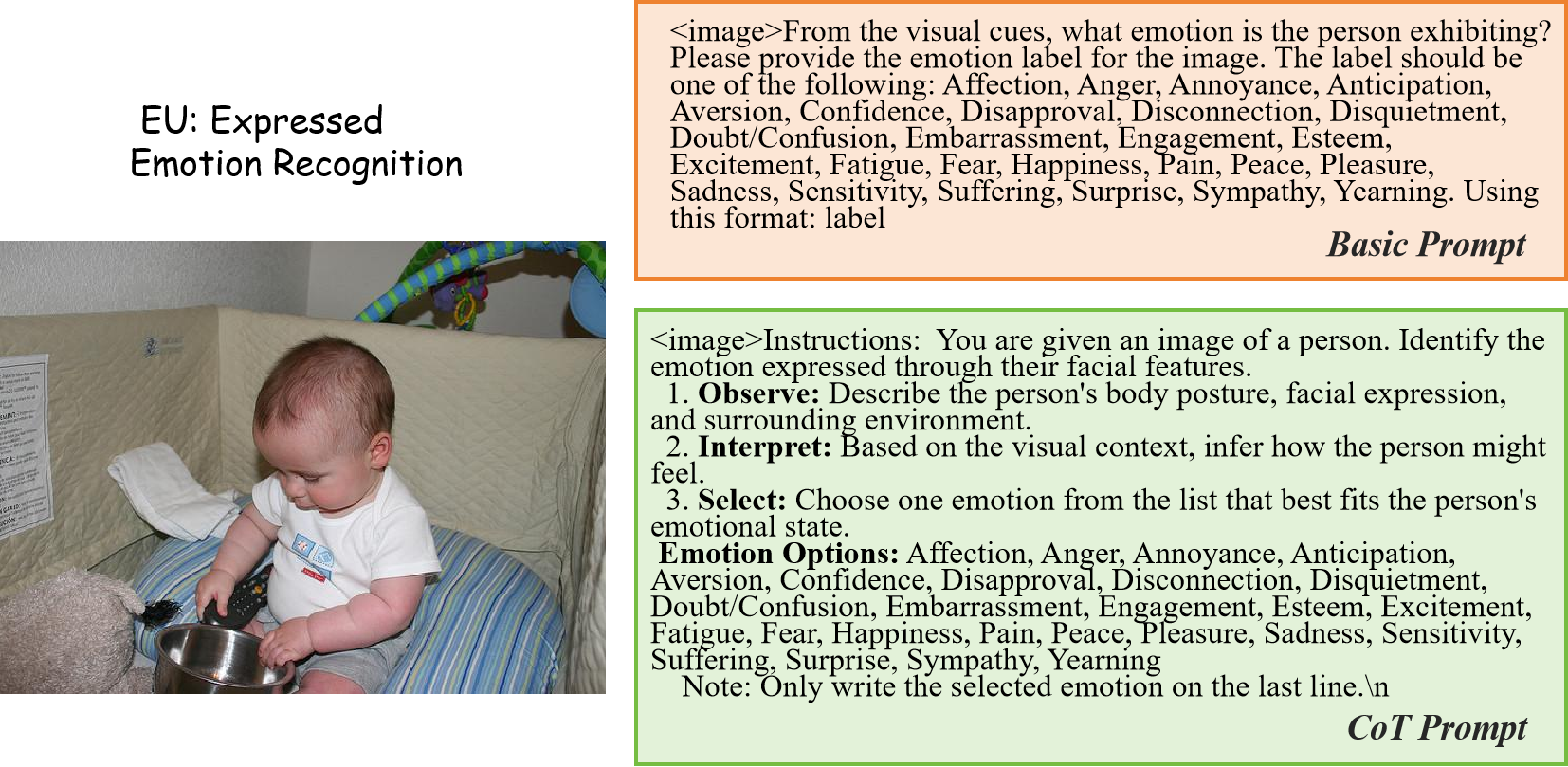}
    \caption{Example of an Expressed Emotion Recognition instruction in the EU Basic and CoT setting.}
    \label{sup:fig:eu-eer-example}
\end{figure*}

To illustrate how EU Basic instructions are instantiated for different subtypes, we present two representative examples. Figure~\ref{sup:fig:eu-eep-example} shows an \textit{Evoked Emotion Prediction} prompt, where the model is asked to infer the likely emotional response of a typical viewer. In contrast, Figure~\ref{sup:fig:eu-eer-example} presents an \textit{Expressed Emotion Recognition} prompt, which focuses on identifying the emotion explicitly displayed by the individual in the image.

\begin{tcolorbox}[colback=gray!5!white, colframe=gray!50!black, title=Prompt Used for ER and EGCG Instruction Generation, boxrule=0.5pt]
\label{fig:er-ecg-prompt-box}
You are building a benchmark to evaluate a vision-language model's emotional reasoning and emotional content generation abilities.

You will receive an image and an emotion label.

Your task is to produce two distinct tasks:
1. Emotion Reasoning: A reasoning question that asks the model to explain \textbf{why} the given emotion is being expressed in the image.
2. Emotion-Guided Content Generation: Generate a writing instruction that asks the model to generate a short paragraph (50--100 words) using the target emotion tone, grounded in the image. Then provide an expressive answer.

Format your response like this:

Reasoning Question: ...

Reasoning Answer: ...

Generation Instruction: ...

Generation Answer: ...
\end{tcolorbox}

To construct instructions for the \textbf{ER} and \textbf{EGCG} tasks, we leverage GPT-4o to generate natural language prompts along with corresponding sample answers. Each generation is conditioned on an image-emotion pair, and the model is asked to produce two distinct instruction types:

\begin{itemize}
    \item \textit{Emotion Reasoning}: A question that requires the model to explain why the given emotion is appropriate for the image.
    \item \textit{Emotion-guided Content Generation}: A creative writing instruction that asks the model to describe the image using the tone or perspective of the given target emotion.
\end{itemize}

\begin{figure*}[htb]
    \centering
    % \captionsetup{skip=2pt}
    \includegraphics[width=\linewidth]{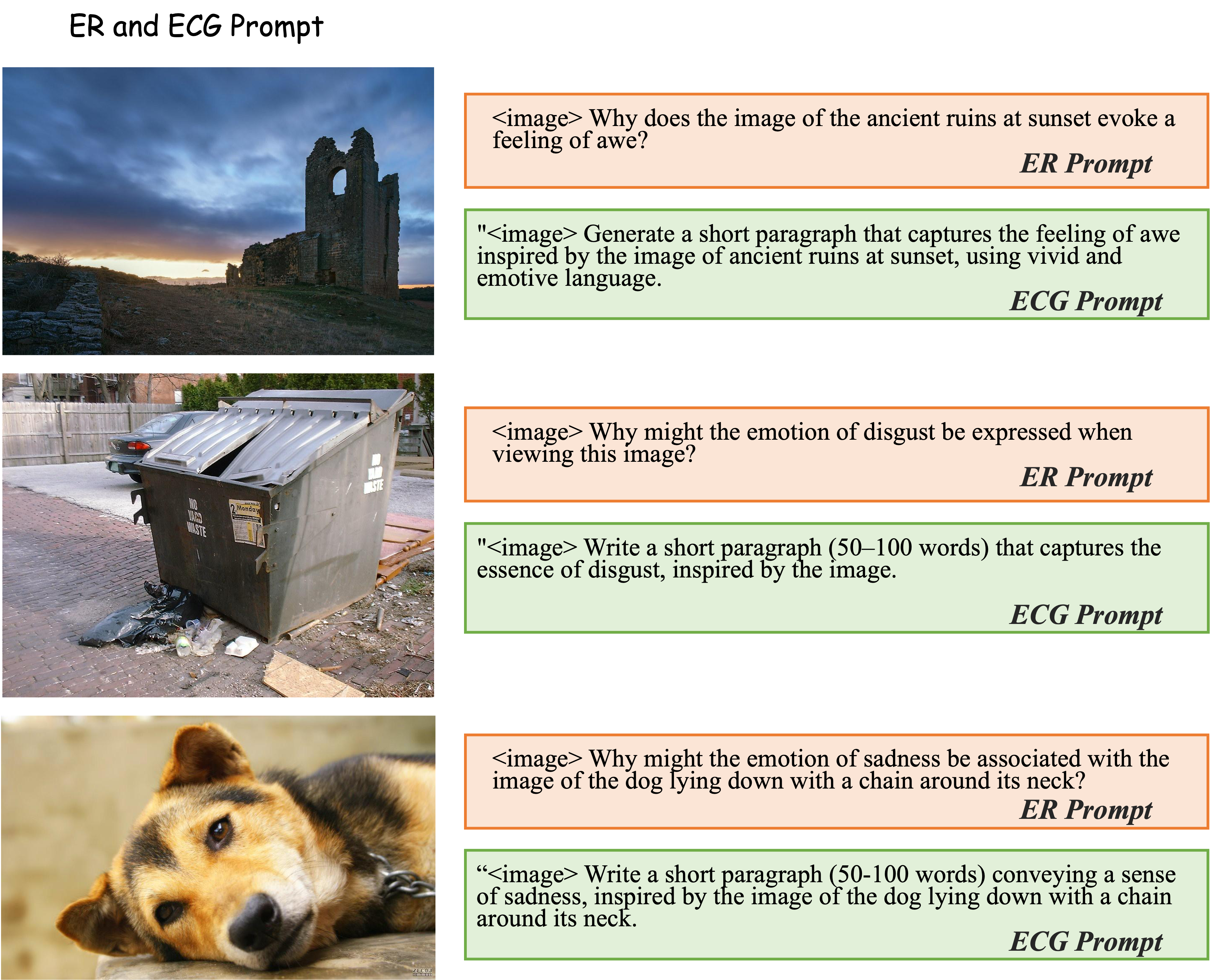}
    \caption{Examples of an ER and EGCG.}
    \label{sup:fig:er-eecg-prompt-case}
\end{figure*}

We use a single, structured system prompt to generate both instruction types in a consistent and controlled manner. A representative example of this prompt is shown in Figure~\ref{fig:er-ecg-prompt-box}.
To illustrate how these instructions are applied in practice, Figure~\ref{sup:fig:er-eecg-prompt-case} presents an example based on a specific image, demonstrating how the model is guided to explain the given emotion and produce a corresponding emotionally aligned description.

\subsection{Annotation  for AICA-Bench}
\label{app:aica_human_annotation}

To construct the visual foundation of AICA-Bench, we relied on 20 trained annotators recruited from our institution and collaborating labs. The annotators were students (including undergraduate and M.S. students), typically 20--27 years old, and were required to be proficient in English to follow the annotation guidelines. They had prior experience in affective computing or related areas. Participation was voluntary, and annotators were informed that they could stop at any time without penalty. Each annotator received a flat honorarium of \$20 for their contribution, which corresponds to a typical short annotation session in our locale and is comparable to the standard compensation level for student research assistants performing similar expert tasks. The 8{,}086 candidate images selected by automated scripts (Section~\ref{sec:aica-vlm-becnhmark-method}) were divided into subsets and assigned to these annotators for manual review.

Before the main filtering phase, all annotators completed a brief training and calibration session. We provided written guidelines describing the goal of AICA-Bench (to build a clear and diverse affective image benchmark), the target emotion categories and valence--arousal space, and several worked examples of acceptable and unacceptable images. During training, annotators practiced on a small set of pilot images and received feedback to align their judgments. In the main annotation stage, we applied the two-step filtering process described in Section~\ref{sec:aica-vlm-becnhmark-method}: images were first selected using automated scripts from a range of publicly available affective computing datasets, and then each candidate was manually reviewed by a trained annotator. Annotators were instructed to reject images that are unsafe for work, contain offensive or highly sensitive content, or are too emotionally ambiguous to support reliable labels; rejected samples were replaced to maintain both emotional clarity and dataset diversity.

The images used in AICA-Bench are drawn from existing publicly released affective image datasets that provide de-identified visual content under research-friendly licenses, and we do not collect any new personally identifying information about the individuals depicted. Our work focuses on secondary use and additional annotations over these public resources. Annotators were informed that their decisions would be used for research purposes and for releasing the benchmark to the community, and we store only anonymized annotator identifiers and binary keep/reject decisions. In line with the content policies of the source datasets, and consistent with our two-step filtering protocol, images that are NSFW, offensive, or otherwise inappropriate for an affective benchmark are removed prior to release, and annotators are encouraged to flag any remaining samples that may be sensitive or ambiguous so that they can be excluded from the final dataset.

\subsection{Metrics}

We use task-specific metrics to evaluate model performance across the AICA-Bench benchmark.

For the \textbf{EU} task, we report \textit{Accuracy}, \textit{Macro F1}, and \textit{Weighted F1}. Among these, Weighted F1 is used as the primary metric, as it accounts for class imbalance across datasets.

For the open-ended generation tasks—\textbf{ER} and \textbf{EGCG}—we employ a fine-tuned scoring model to assess the quality of responses. For ER, we evaluate three dimensions: \textit{Emotion Alignment}, which measures how well the response reflects the target emotion; \textit{Descriptiveness}, which assesses the richness and specificity of the explanation; and \textit{Causal Soundness}, which captures whether the reasoning provides a plausible link between the visual content and the stated emotion. For EGCG, we evaluate \textit{Emotion Alignment} and \textit{Descriptiveness} using the same criteria.

Table~\ref{tab:metrics-summary} summarizes the evaluation metrics used for each task.

\begin{table*}[h]
\centering
\small
\caption{Evaluation metrics used for each task in AICA-Bench}
\label{tab:metrics-summary}
\begin{tabular}{l l}
\toprule
\textbf{Task} & \textbf{Evaluation Metrics} \\
\midrule
Emotion Understanding (EU) &  \textit{Accuracy, Macro F1, Weighted F1} \\
Emotion Reasoning (ER) &  \textit{Emotion Alignment, Descriptiveness, Causal Soundness} \\
Emotion-guided Content Generation (EGCG) &  \textit{Emotion Alignment, Descriptiveness} \\
\bottomrule
\end{tabular}
\end{table*}

\subsection{VLM Models Details}
\label{appendix:model-details}

To ensure a comprehensive evaluation, we include both closed-source and open-source vision-language models (VLMs). Within the open-source group, we select models at three parameter scales—2--3B, 7--8B, and 14--16B—covering a range of compute capacities and deployment settings.

\paragraph{Closed-source models.}
These models are accessed via API and represent strong general-purpose commercial systems. We include:
\begin{itemize}
    \item \textbf{GPT-4o and GPT-4o-Mini}: The latest generation of OpenAI’s VLMs with multimodal reasoning capabilities.
    \item \textbf{Gemini-2.5-Pro / Flash / 2.0-Flash}: Google's vision-language models optimized for both generation and perception tasks.
    \item \textbf{Qwen-VL-Max / Plus}: High-performing commercial VLMs developed by Alibaba, designed for instruction-following in multimodal contexts.
\end{itemize}

\paragraph{Open-source models.}
We include diverse open-source models from different families and architecture designs. All selected models support vision input and are instruction-tuned.

\begin{itemize}
    \item \textbf{Qwen series} (2B, 3B, 7B):
    \begin{itemize}
        \item \textit{Qwen2-VL-2B / Qwen2.5-VL-3B / Qwen2.5-VL-7B}: Lightweight VLMs designed for practical deployment with strong performance in perception-based tasks.
    \end{itemize}

    \item \textbf{InternVL series} (2B, 8B, 14B):
    \begin{itemize}
        \item \textit{InternVL3-2B / 8B / 14B, InternVL2.5-8B}: VLMs optimized for high-resolution visual reasoning, with a multi-scale fusion backbone.
    \end{itemize}

    \item \textbf{Ovis series} (2B, 8B, 16B):
    \begin{itemize}
        \item \textit{Ovis2-2B / 8B / 16B}: Focused on vision-language alignment and long-form generation, designed for instruction-based multimodal understanding.
    \end{itemize}

    \item \textbf{LLaVA series} (7B, 13B):
    \begin{itemize}
        \item \textit{LLaVA-1.6-7B / 13B, LLaVA-OneVision}: Popular open-source VLMs pretrained with image-text alignment and dialogue fine-tuning.
    \end{itemize}

    \item \textbf{MiniCPM series}:
    \begin{itemize}
        \item \textit{MiniCPM-V / MiniCPM-O-2.6}: Compact VLMs designed for mobile and edge deployment, with surprisingly strong performance on vision tasks.
    \end{itemize}
\end{itemize}

\section{Detail of AICA-Bench Scoring Model}

\subsection{Dataset Details}

\begin{figure*}[htb]
    \centering
    \includegraphics[width=\linewidth]{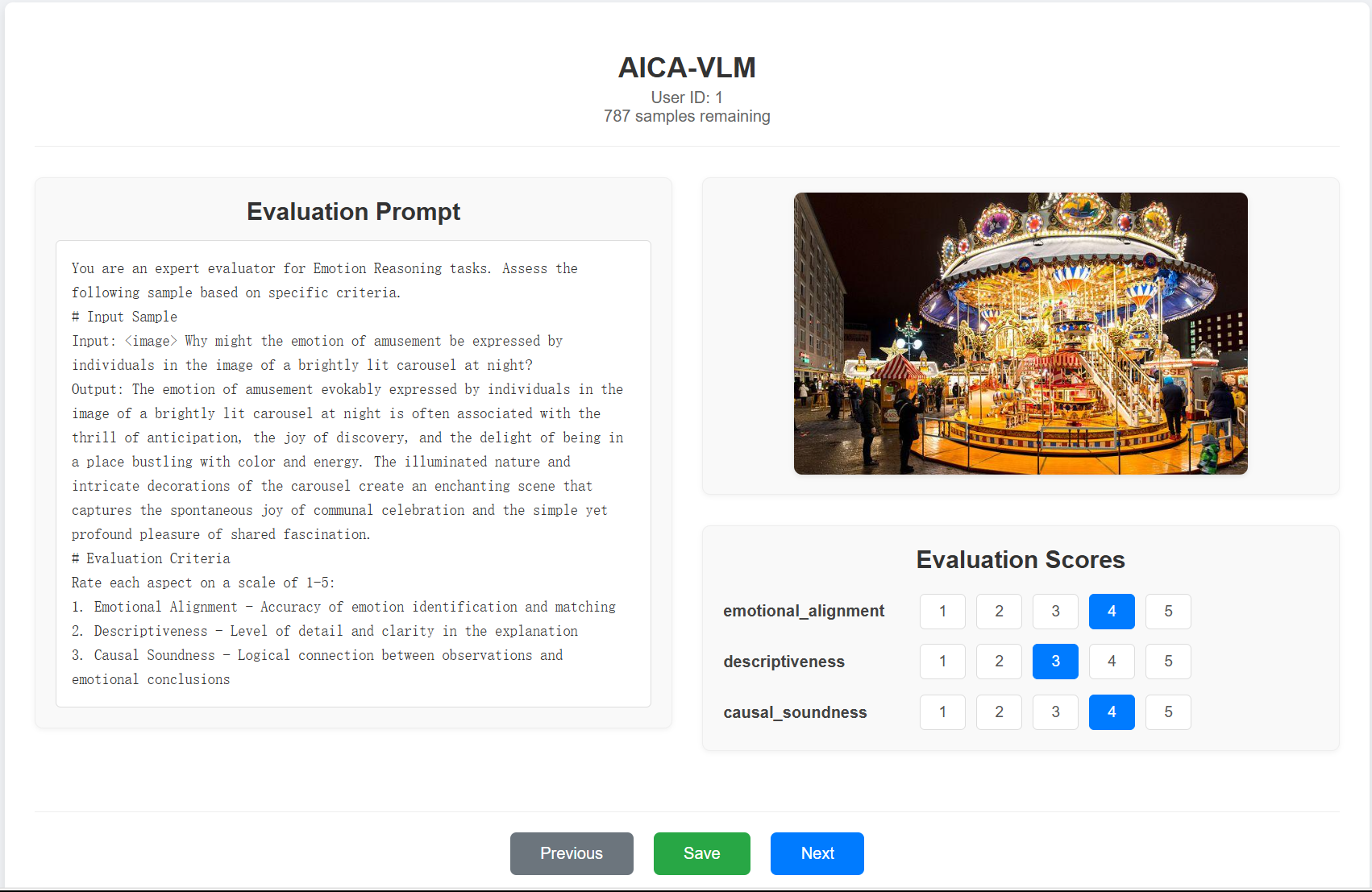}
    \caption{Human~Annotation~Interface}
    \label{sup:fig:human_annotation_interface}
\end{figure*}

\begin{figure*}[htb]
    \centering
    \includegraphics[width=\linewidth]{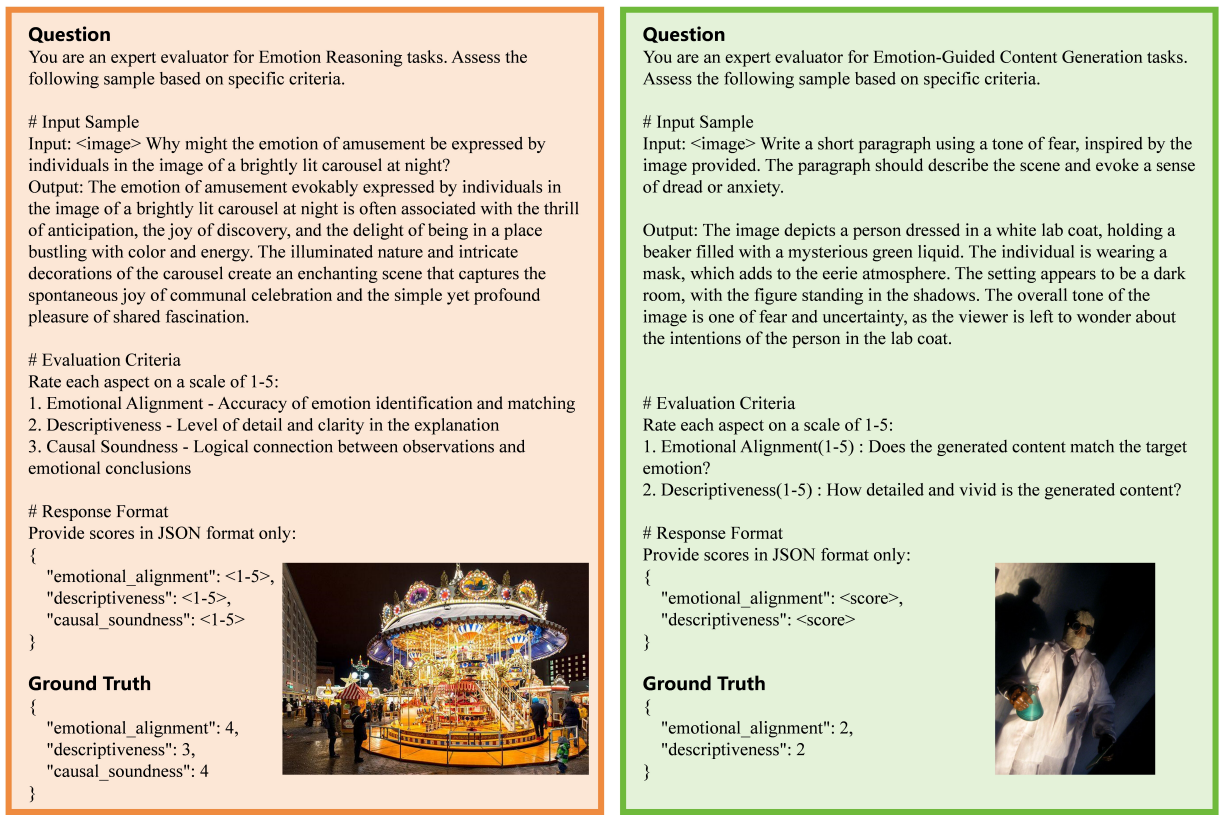}
    \caption{Examples of the Instruction-Tuning Dataset Format}
    \label{sup:fig:score_instruction}
\end{figure*}

To effectively train our scoring model, we meticulously constructed a high-quality human-annotated dataset. This dataset serves as the foundation for fine-tuning our scoring model, enabling it to learn the nuances of human judgment regarding MLLM output.

Figure \ref{sup:fig:human_annotation_interface} illustrates the custom human annotation interface employed for collecting high-quality data for our scoring model. This platform facilitated expert evaluators in assessing model-generated responses (e.g., for emotion reasoning tasks) against predefined criteria, such as 'emotional\_alignment', 'descriptiveness', and 'causal\_soundness', ensuring a rigorous and consistent scoring process.

The creation of this dataset involved five domain experts independently evaluating 10K distinct data samples. Each sample comprised an image-based question, the MLLM's generated answer, and a set of predefined evaluation criteria. The experts provided scores for each criterion, reflecting the quality of the MLLM's response. This human-annotated data was then formatted into a specialized instruction-tuning dataset, specifically designed for training our scoring model.

Figure \ref{sup:fig:score_instruction} provides illustrative examples of the instruction-tuning format used in our dataset. As depicted, each entry includes a "Question" prompt that instructs the expert evaluator on the task (e.g., Emotion Reasoning, Emotion-Guided Content Generation) and the specific criteria for assessment. An "Input Sample" section presents the original image-based prompt and the MLLM's generated "Output" response. Crucially, the "Evaluation Criteria" section lists the specific aspects to be rated on a 1-5 scale, such as "Emotional Alignment," "Descriptiveness," and "Causal Soundness." Finally, the "Ground Truth" section contains the actual scores provided by the human experts in a JSON format, serving as the target labels for our scoring model's training. This structured format ensures that our fine-tuning process directly aligns the model's output with expert human assessments of MLLM response quality.

\subsection{Training Details}

Our scoring model was fine-tuned from the Qwen2.5-VL-7B model using the LLaMA-Factory framework with Low-Rank Adaptation (LoRA). This approach was chosen for its efficiency in adapting large pre-trained models with reduced computational overhead. Training was performed on a single NVIDIA A100 80GB GPU, leveraging a high-quality human-annotated dataset. The detailed training configuration is summarized in Table \ref{tab:training_config}. Key parameters, including an effective batch size of 8, a learning rate of 1.0e-4, a cosine learning rate schedule with a 0.1 warmup ratio, and 3 training epochs, were optimized to ensure stable convergence and prevent overfitting on our specialized dataset.

\begin{table*}[htbp]
\centering
\caption{Scoring model training configuration}
\label{tab:training_config}
\begin{tabular}{ll}
    \toprule
    \textbf{Parameter}                   & \textbf{Value}                                            \\
    \midrule
    Base Model                           & Qwen2.5-VL-7B                                             \\
    Fine-tuning Framework                & LLaMA-Factory                                             \\
    Fine-tuning Method                   & LoRA                                                      \\
    GPU                                  & 1 $\times$ NVIDIA A100                              \\
    % Training Dataset Size                & 1,600 samples                                             \\
    Effective Batch Size                 & 8 (per-device: 2, gradient accumulation: 4)               \\
    Learning Rate                        & 1.0e-4                                                    \\
    Learning Rate Scheduler Type         & Cosine                                                    \\
    Warmup Ratio                         & 0.1                                                       \\
    Number of Training Epochs            & 3                                                         \\
    \bottomrule
\end{tabular}
\end{table*}

\begin{table*}[h]
\centering
\small
\setlength{\tabcolsep}{6pt}
\renewcommand{\arraystretch}{1.0}
\begin{tabular}{lcccccc}
\hline
\textbf{Model} & EU Basic & EU CoT & EU Avg. & ER Avg. & EG Avg. & \textbf{Overall Avg. (\%)} \\
\hline
\multicolumn{7}{c}{\textbf{Closed-source Models}} \\
\hline
Gemini-2.5-pro            & 66.97 & 67.57 & 67.27 & 79.08 & 74.13 & \textbf{73.49} \\
Qwen-VL-max               & 64.07 & 65.98 & 65.02 & 77.75 & 75.93 & \textbf{72.90} \\
ChatGPT-4o                 & 64.44 & 65.42 & 64.93 & 77.81 & 75.73 & \textbf{72.82} \\
Gemini-2.5-flash          & 68.05 & 69.32 & 68.68 & 76.55 & 68.19 & \textbf{71.14} \\
ChatGPT-4o-mini            & 60.15 & 63.68 & 61.91 & 76.45 & 74.09 & \textbf{70.81} \\
Qwen-VL-plus              & 60.04 & 67.81 & 63.92 & 72.39 & 66.86 & \textbf{67.73} \\
Gemini-2.0-flash          & 67.16 & 68.98 & 68.07 & 71.05 & 63.93 & \textbf{67.68} \\
\hline
\multicolumn{7}{c}{\textbf{Open-source Models}} \\
\hline
Qwen2.5VL-7B~\cite{Qwen2.5-VL}                 & 56.43 & 57.25 & 56.84 & 74.50 & 66.00 & \textbf{65.78} \\
Ovis2-16B~\cite{lu2024ovis}                    & 54.38 & 54.70 & 54.54 & 68.24 & 71.56 & \textbf{64.78} \\
Ovis2-8B~\cite{lu2024ovis}                     & 53.63 & 52.73 & 53.18 & 68.89 & 70.81 & \textbf{64.29} \\
InternVL3-14B~\cite{zhu2025internvl3exploringadvancedtraining}  & 52.91 & 52.04 & 52.47 & 68.27 & 66.50 & \textbf{62.41} \\
InternVL3-8B~\cite{zhu2025internvl3exploringadvancedtraining}   & 52.18 & 52.98 & 52.58 & 67.21 & 67.27 & \textbf{62.35} \\
InternVL2.5-8B~\cite{chen2024internvl}         & 51.89 & 51.03 & 51.46 & 66.48 & 68.86 & \textbf{62.27} \\
MiniCPM-o-2.6~\cite{yao2024minicpm}            & 52.73 & 48.65 & 50.69 & 70.16 & 64.98 & \textbf{61.94} \\
Qwen2.5VL-3B~\cite{Qwen2.5-VL}                 & 54.58 & 53.67 & 54.13 & 69.68 & 61.63 & \textbf{61.81} \\
Qwen2VL-7B~\cite{Qwen2-VL}                     & 53.52 & 55.19 & 54.36 & 65.23 & 64.76 & \textbf{61.45} \\
Ovis2-2B~\cite{lu2024ovis}                     & 49.32 & 46.45 & 47.88 & 66.38 & 66.04 & \textbf{60.10} \\
LLaVA-1.6-13B~\cite{liu2024llavanext}          & 36.78 & 46.82 & 41.80 & 73.57 & 64.51 & \textbf{59.96} \\
LLaVA-1.6-7B~\cite{liu2024llavanext}           & 36.58 & 50.22 & 43.40 & 73.81 & 59.58 & \textbf{58.93} \\
Qwen2VL-2B~\cite{Qwen2-VL}                     & 51.85 & 49.38 & 50.62 & 67.33 & 57.97 & \textbf{58.64} \\
MiniCPM-V-2.6~\cite{yao2024minicpmvgpt4vlevelmllm} & 43.70 & 47.25 & 45.48 & 65.77 & 63.00 & \textbf{58.08} \\
LLaVA-onevision~\cite{li2024llava}             & 54.02 & 53.25 & 53.64 & 63.78 & 54.18 & \textbf{57.20} \\
InternVL3-2B~\cite{zhu2025internvl3exploringadvancedtraining}   & 43.87 & 39.53 & 41.70 & 65.78 & 60.93 & \textbf{56.14} \\
\hline
\end{tabular}
\caption{Performance comparison on AICA-Bench tasks. EU: Emotion Understanding, ER: Emotion Reasoning, EGCG: Emotion-guided Content Generation. Scores are shown as percentages.}
\label{tab:more_main_results_simplified}
\end{table*}

\subsection{More Experiment Results}
\label{appendix:more_analysis}

\subsection{Comprehensive Model Comparison}
Table~\ref{tab:more_main_results_simplified} provides the full evaluation results for all closed-source and open-source models assessed in the AICA-Bench benchmark. While the main paper reports the aggregated findings, this appendix table offers a complete breakdown across five metrics: EU Basic, EU CoT, EU Average, ER Average, and EGCG Average, as well as the final overall average score.

\subsection{Performance Radar Charts by Model Series}

\begin{figure*}[htb]
    \centering
    % \captionsetup{skip=2pt}
    \includegraphics[width=\linewidth]{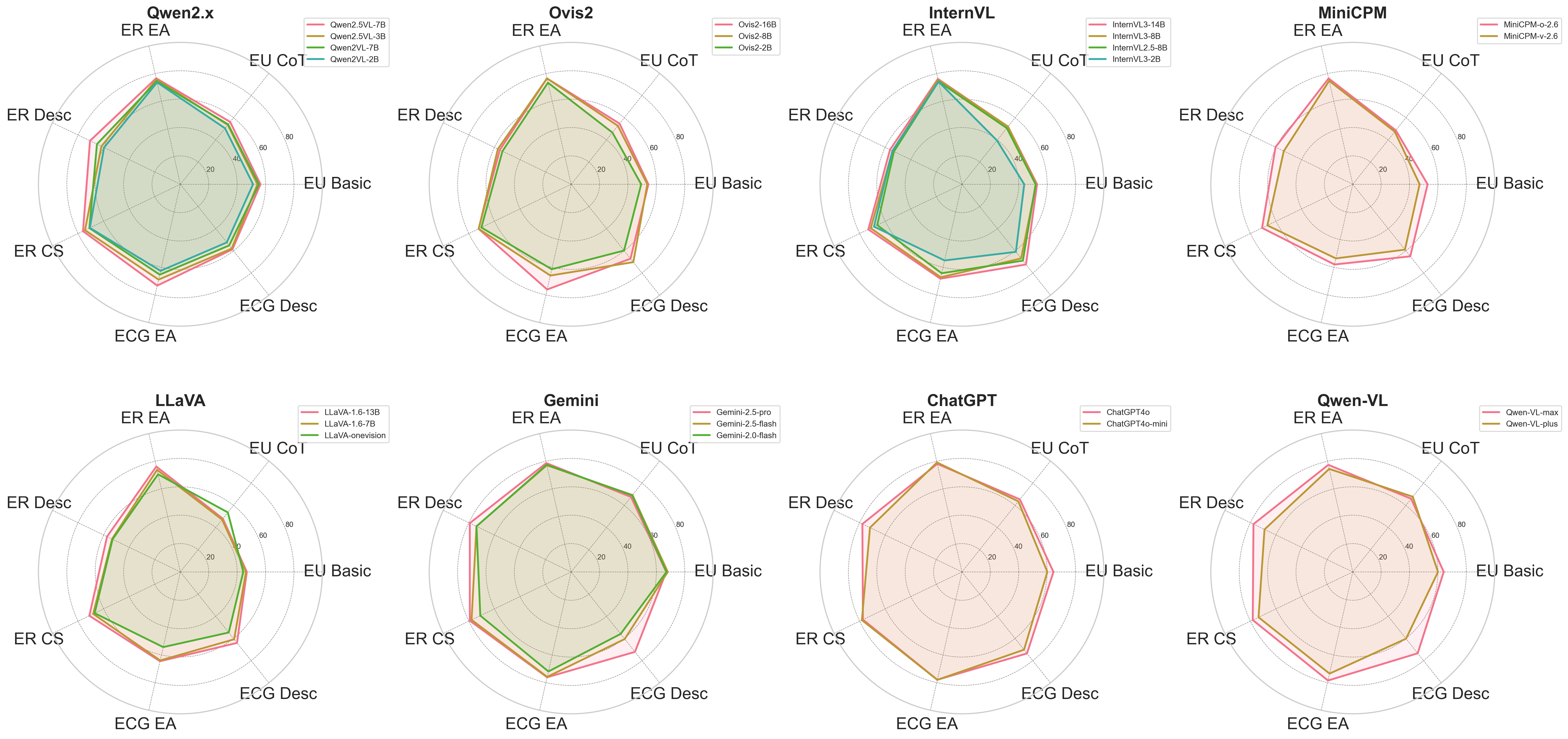}
    \caption{Radar charts comparing performance across tasks for eight major VLM model series (both open-source and closed-source)}
    \label{sup:fig:radar-all-series}
\end{figure*}

To complement the quantitative leaderboard presented in the main paper, Figure~\ref{sup:fig:radar-all-series} visualizes the performance of all evaluated VLMs across seven core metrics using radar charts. Each subplot corresponds to a distinct model series—either open-source or closed-source—and includes all variants within that series. The plotted axes cover: EU Basic, EU CoT, ER Emotional Alignment (ER EA), ER Descriptiveness (ER Desc.), ER Causal Soundness (ER CS), EGCG Emotional Alignment (EGCG EA), and EGCG Descriptiveness (EGCG Desc.).

% This visualization enables a more intuitive comparison of each model’s strength and weakness pattern. For instance, Gemini and ChatGPT models exhibit relatively balanced performance across all dimensions, while open-source models such as Qwen2 and InternVL show noticeable variance across reasoning and generation metrics. The radar plots also make it easier to detect trends within a series, such as consistent improvements or regressions with increasing model size.

Radar chart comparisons reveal several consistent trends across model series. Closed-source models generally demonstrate strong and balanced performance across all affective tasks. In contrast, open-source models often achieve high scores in reasoning tasks (e.g., ER-EA, ER-CS) but underperform in emotion-guided generation, especially in descriptive quality. Additionally, within many model families, performance improvements do not scale uniformly with model size, highlighting the influence of optimization and alignment strategies.

\begin{figure*}[htbp]
    \centering
    \subfigure[Overall Avg. Score vs. Model Size]{
        \includegraphics[width=0.40\linewidth]{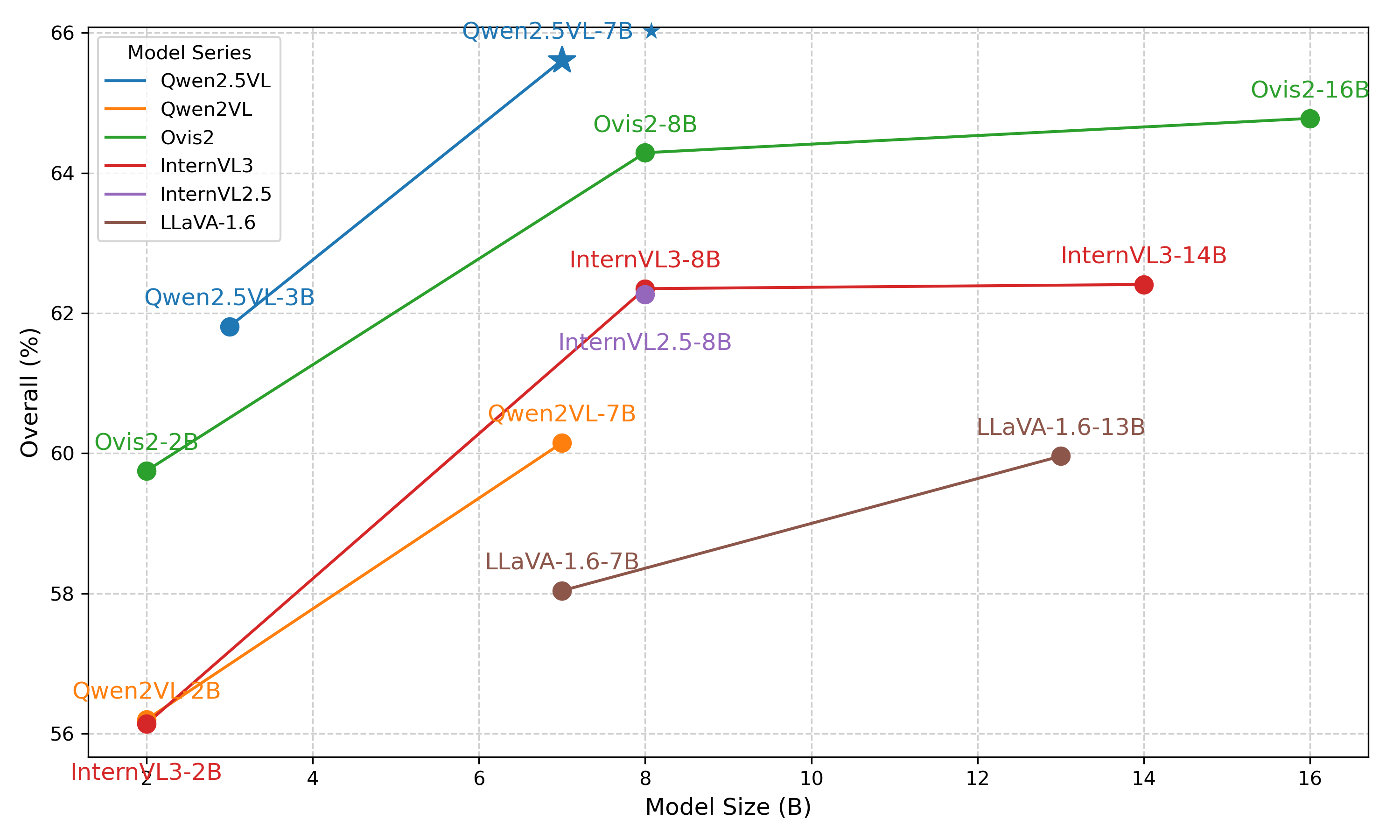}
    }
    \hfill
    \subfigure[EU Avg. vs. Model Size]{
        \includegraphics[width=0.40\linewidth]{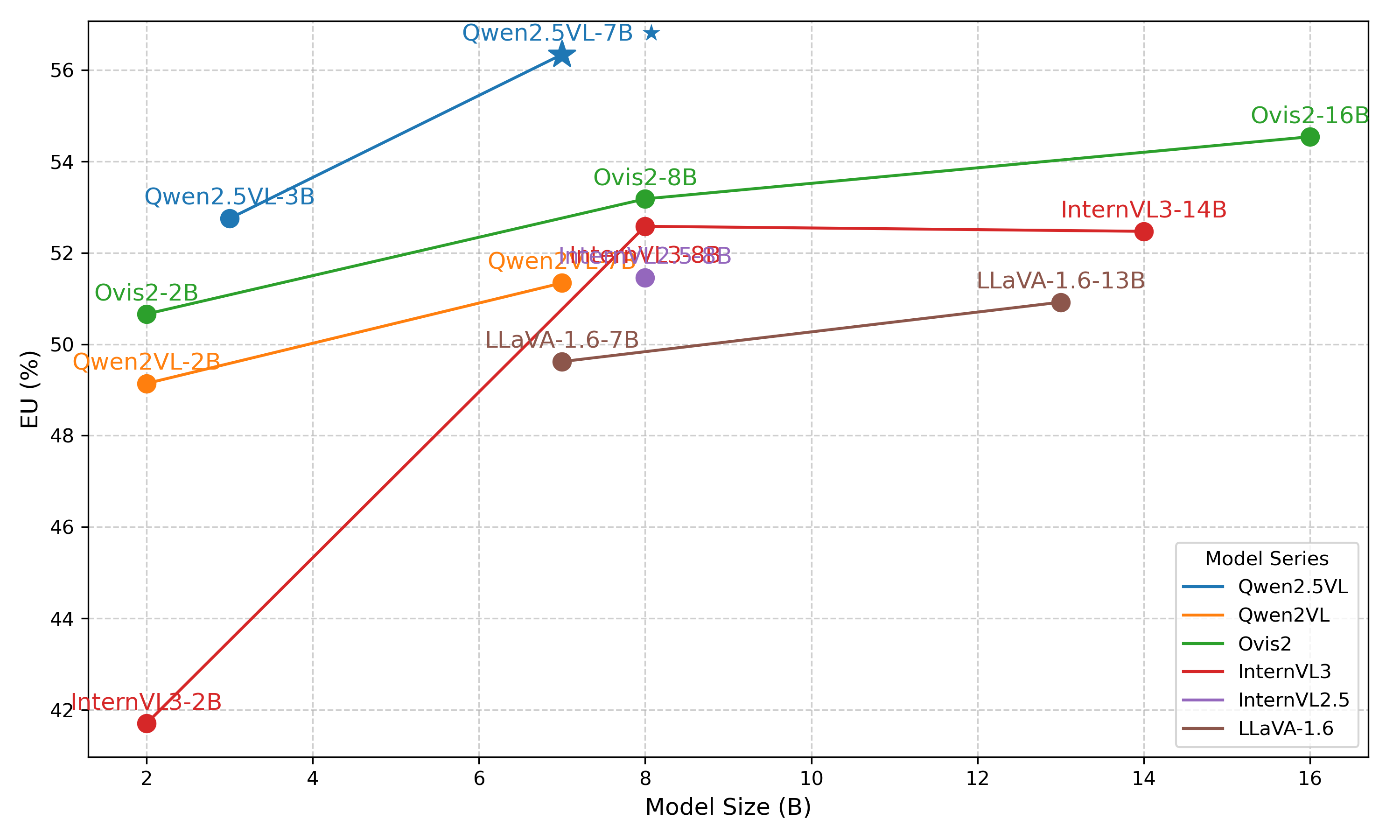}
    }

    \vskip\baselineskip
    
    \subfigure[ER Avg. vs. Model Size]{
        \includegraphics[width=0.40\linewidth]{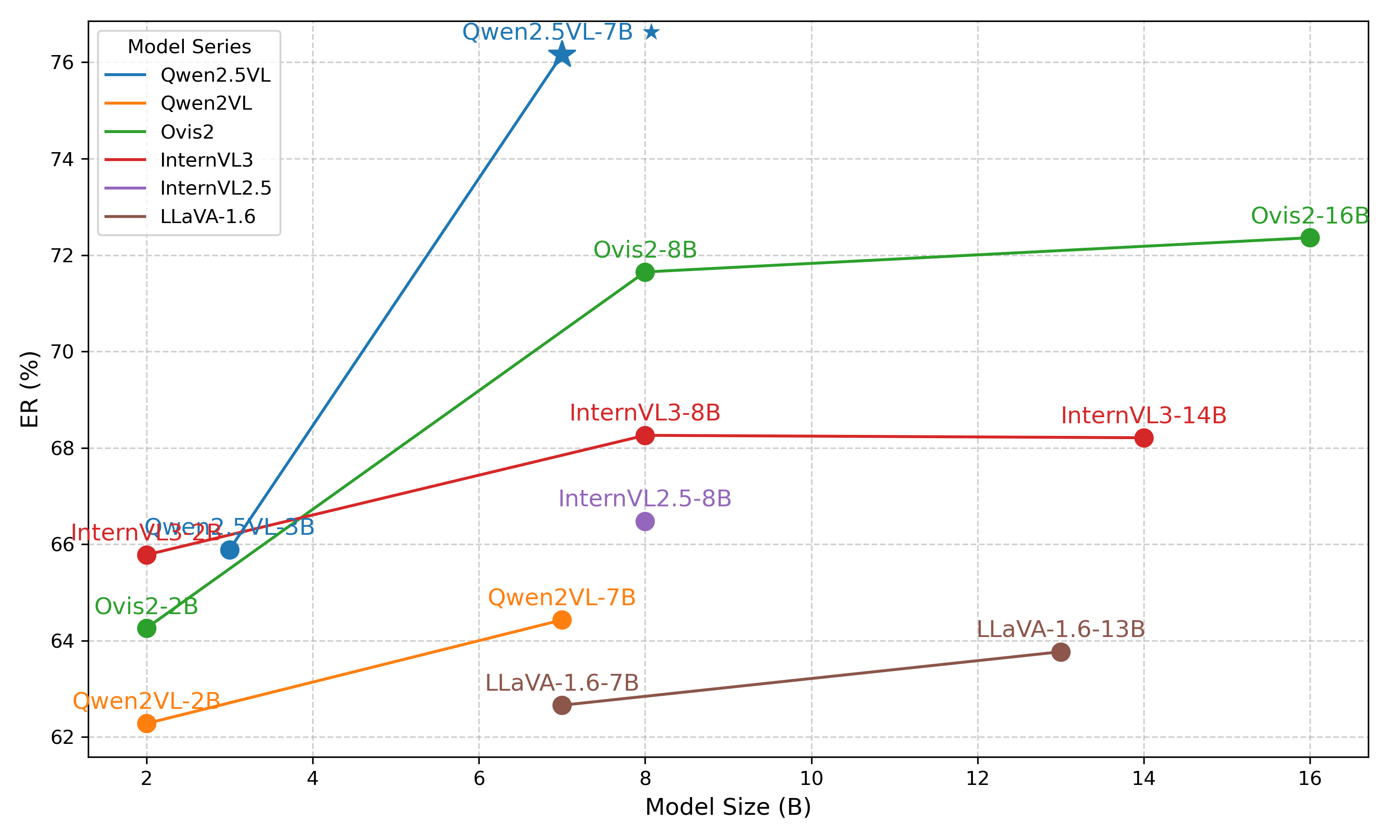}
    }
    \hfill
    \subfigure[EGCG Avg. vs. Model Size]{
        \includegraphics[width=0.40\linewidth]{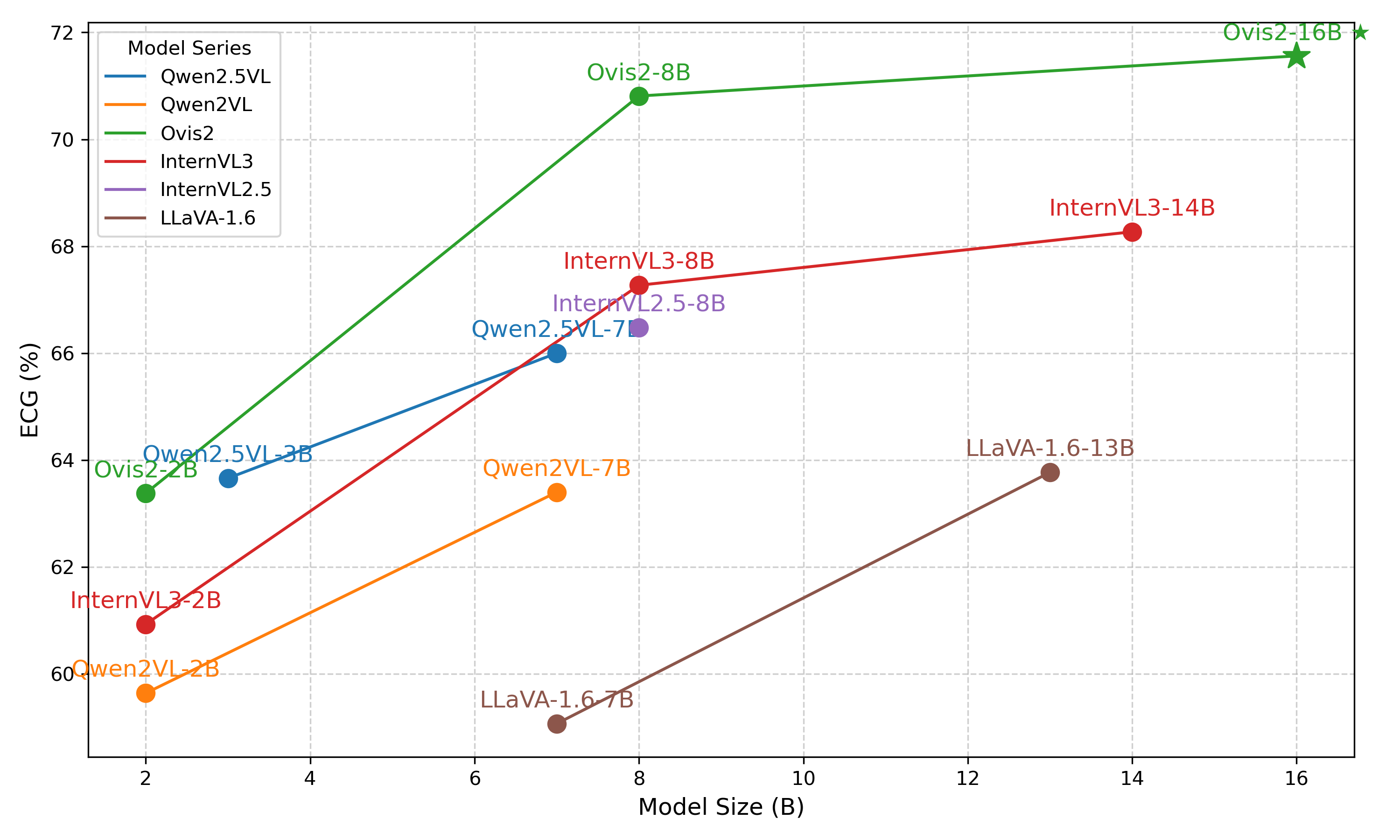}
    }
    \caption{Performance comparison across subtasks (EU, ER, EGCG) and overall average scores for open-source models. Each line represents a model series, with the best-performing model in each subtask marked by a star.}
    \label{fig:sup:subtask-four-panel}
\end{figure*}

\subsection{Analysis of Model Size vs. Performance}
To substantiate our observation that model size is not a reliable predictor of affective performance, we further analyze the scores of open-source models across four evaluation dimensions: \textit{EU Avg.}, \textit{ER Avg.}, \textit{EGCG Avg.}, and \textit{Overall Avg.}, plotted in Figure~\ref{fig:sup:subtask-four-panel}. Each curve represents a model series, and markers indicate performance at different parameter scales.

Several patterns reinforce our conclusion:
\begin{itemize}
    \item \textbf{Qwen2.5VL-7B (65.78\%)} achieves the highest overall average score among all open-source models, outperforming larger models such as \textbf{InternVL3-14B (62.41\%)} and \textbf{LLaVA-1.6-13B (59.96\%)}, underscoring that increased parameter size does not guarantee superior affective capability.
    
    \item \textbf{Qwen2.5VL-3B (61.81\%)} outperforms multiple larger models such as \textbf{InternVL3-14B (62.41\%)} and \textbf{LLaVA-1.6-13B (59.96\%)}, and exhibits particularly strong performance on \textit{ER Avg.} and \textit{EGCG Avg.}.
    \item \textbf{Ovis2-8B (64.29\%)} achieves higher average scores than the 14B and 13B models of other series, including InternVL3 and LLaVA, indicating that even mid-sized models can be highly competitive.
    \item Across series, performance fluctuations do not exhibit monotonic growth with scale: for example, within InternVL3, the smallest model (2B) performs worse overall (56.14\%), but the 8B variant (62.35\%) and the 14B (62.41\%) are nearly tied, suggesting diminishing returns.
    \item In the EGCG dimension in particular, smaller models like \textbf{Qwen2VL-2B (58.64\%)} and \textbf{MiniCPM-2.6 (58.08\%)} are competitive with or even outperform some 7B–13B models.
\end{itemize}

Interestingly, when focusing specifically on the \textit{EGCG Avg.} dimension, a more noticeable trend emerges: models with larger parameter sizes tend to perform better. For instance, \textbf{Ovis2-16B} achieve higher EGCG scores (71.56\%) compared to their smaller counterparts like \textbf{Ovis2-2B (66.04\%)}. This suggests that model size might play a more significant role in generation tasks where maintaining coherence, style, and emotional grounding in generated content requires greater model capacity.

\begin{table*}[t]
\centering
\small
\setlength{\tabcolsep}{4.2pt}
\renewcommand{\arraystretch}{1.08}
\caption{Human evaluation on ER and EGCG (200 images, 10 annotators). Values are mean ratings (no $\pm$).}
\label{tab:human_eval_er_egcg}
\begin{tabular}{l|cccccccccccccc}
\toprule
\textbf{Model} &
\multicolumn{8}{c}{\textbf{ER}} &
\multicolumn{6}{c}{\textbf{EGCG}} \\
\cmidrule(lr){2-9}\cmidrule(lr){10-15}
&
\multicolumn{2}{c}{Align} &
\multicolumn{2}{c}{Desc} &
\multicolumn{2}{c}{Causal} &
\multicolumn{2}{c}{Overall} &
\multicolumn{2}{c}{Align} &
\multicolumn{2}{c}{Desc} &
\multicolumn{2}{c}{Overall} \\
\cmidrule(lr){2-3}\cmidrule(lr){4-5}\cmidrule(lr){6-7}\cmidrule(lr){8-9}
\cmidrule(lr){10-11}\cmidrule(lr){12-13}\cmidrule(lr){14-15}
&
Base & GAT &
Base & GAT &
Base & GAT &
Base & GAT &
Base & GAT &
Base & GAT &
Base & GAT \\
\midrule
GPT-4o        &
4.12 & 4.36 &
3.05 & 3.61 &
3.18 & 3.74 &
3.45 & \textcolor[rgb]{0.0, 0.6, 0.0}{\textbf{3.90}} &
4.20 & 4.48 &
3.22 & 3.79 &
3.71 & \textcolor[rgb]{0.0, 0.6, 0.0}{\textbf{4.02}} \\
Qwen2.5-VL-7B &
3.88 & 4.14 &
2.94 & 3.47 &
3.02 & 3.56 &
3.28 & \textcolor[rgb]{0.0, 0.6, 0.0}{\textbf{3.72}} &
3.95 & 4.21 &
3.10 & 3.62 &
3.53 & \textcolor[rgb]{0.0, 0.6, 0.0}{\textbf{3.86}} \\
\bottomrule
\end{tabular}
\end{table*}

We speculate that this correlation may be due to the inherently more complex nature of text or image generation in affective settings, which benefits from the richer representation capabilities and longer context handling of larger models. Nevertheless, the trend is still not strictly linear, as some smaller models like \textbf{Qwen2.5VL-3B} and \textbf{Qwen2VL-7B} still demonstrate competitive performance, reinforcing the importance of model training and alignment strategies.

\section{GAT Prompting}
\label{app:gat_prompting}

This section outlines the concrete prompts used by the Grounded Affective Tree (GAT) framework across the three tasks in AICA-Bench. Each prompt is designed following the principles described in Section~\ref{subsec:gat_method}, combining region-level visual scaffolding with structured affective reasoning or grounded generation. In addition to the prompts, we also describe the implementation of visual scaffolding, which serves as a key component in guiding the model's understanding and generation of affective content. Furthermore, we provide results from human evaluation to validate the effectiveness of GAT, ensuring that the improvements are not only supported by automatic metrics but also by human judgments.

\label{app:visual_scaffolding}
\begin{figure*}[t]
    \centering
    \captionsetup{skip=2pt}
    \includegraphics[width=\linewidth]{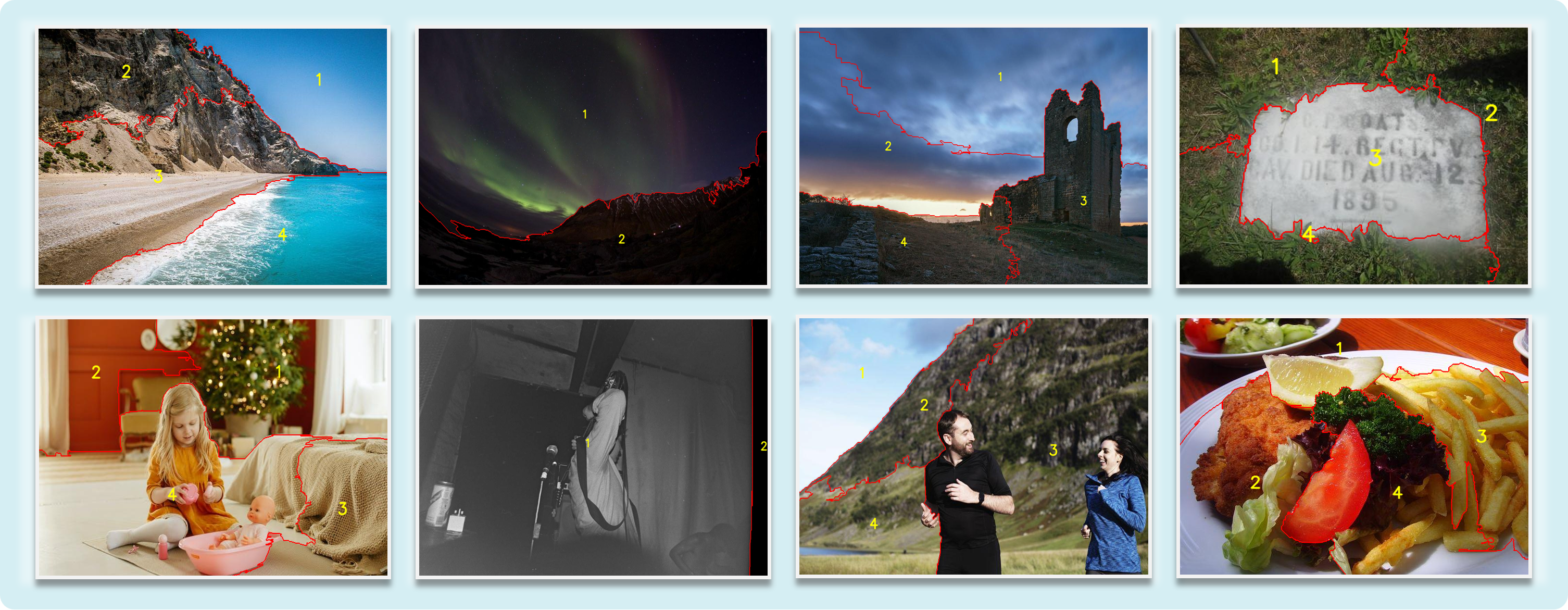}
    \caption{Visual Scaffolding examples used in GAT Prompting. The red contours highlight the segmented regions that serve as visual anchors for guiding the model’s grounded reasoning.}
    \label{fig:visual_scaffolding_showcase}
\end{figure*}

\paragraph{Visual Scaffolding Implementation.}  
For image segmentation, we use the graph-based image segmentation method that divides the image into regions based on pixel similarities. The segmentation is controlled by three parameters: \texttt{scale}, \texttt{sigma}, and \texttt{min\_size\_ratio}. After the initial segmentation, we dynamically merge the smallest regions until the number of regions is reduced to a target range (2–4 regions) based on the adjacency of regions and their relative areas. The process was implemented in Python using the \texttt{skimage} library and is available on GitHub~\cite{felzenszwalb_segmentation_github}.

Figure~\ref{fig:visual_scaffolding_showcase} presents representative examples of the Visual Scaffolding used in GAT Prompting. Each image is decomposed into several coarse regions using graph-based segmentation, where each region is assigned a unique numerical ID (e.g., 1, 2, 3...). Subsequently, these IDs and contours are overlaid on the raw image before feeding it into the VLM. These annotated regions serve as explicit visual anchors for downstream prompting, providing a structured basis for guiding the model to identify objective visual cues before performing affective reasoning or generation.

\paragraph{Human evaluation.}
In addition to the automatic evaluation metrics reported in Table~\ref{tab:gat_improvement_on_tasks}, we further conduct a human evaluation to validate the effectiveness of GAT from a human judgment perspective. Specifically, we randomly sample 200 instances and select two representative models, GPT-4o and Qwen2.5-VL-7B. For each instance, we collect both the Baseline and GAT-generated responses for the ER and EGCG tasks. These paired responses are then evaluated by 10 human annotators, who independently assign scores following the same affective criteria used in AICA-Bench, including emotion alignment and descriptiveness for both tasks, as well as causal soundness for ER. All scores are given on a 1--5 Likert scale. We report the mean ratings aggregated across annotators and samples in Table~\ref{tab:human_eval_er_egcg}. The results show that GAT consistently receives higher human ratings across models and tasks, providing complementary evidence to the automatic evaluation results.

% NOTE: requires \usepackage[most]{tcolorbox} in the preamble.

% ===================== EU =====================
\noindent \textbf{GAT Prompt for EU (Tree-of-Thoughts Strategy).}

% Stage 1
\begin{promptbox}{Stage 1: Indexed Region Observation (Root Node)}
\small
\textit{Input:} An image with numbered segmented regions.\\[2pt]
\textit{Task:} Systematically describe the visual content of each marked region based on its ID.\\[4pt]
\textit{Instruction:}\\
For each numbered region visible in the image, describe its key visual clues (shapes, textures, object parts, or interactions). Do NOT infer emotions yet. Focus on objective description.

\textit{Output Format:}
\begin{itemize}[leftmargin=*]
    \item Region 1: [Description of visual elements in this specific area]
    \item Region 2: [Description of visual elements in this specific area]
    \item ...
    \item Region N: [Description...]
\end{itemize}
\end{promptbox}

\vspace{6pt}

% Stage 2
\begin{promptbox}{Stage 2: Candidate Generation (Branching)}
\small
\textit{Input:} The descriptions of Regions 1 to N and the global polarity estimate.\\[2pt]
\textit{Goal:} Propose three distinct emotion candidates. You must cite specific Region IDs to justify your choice, specifically explaining the intensity.\\[4pt]
\textit{Instruction:}\\
Generate three hypotheses. For each, explain which Region number supports the emotion and its intensity level.

\begin{itemize}[leftmargin=*]
    \item Hypothesis A: [Emotion Label] (Intensity: Low/Medium/High)\\
    Evidence: Region [X] shows [detail] which suggests [level] arousal. Region [Y] shows [detail]...
    \item Hypothesis B: [Different Emotion Label] (Intensity: Different Level)\\
    Evidence: Region [Z] contradicts the previous hypothesis because it shows...
    \item Hypothesis C: ...
\end{itemize}
\end{promptbox}

\vspace{6pt}

% Stage 3
\begin{promptbox}{Stage 3: ID-Based Verification (Evaluation)}
\small
\textit{Role:} You are a grounded verifier.\\[2pt]
\textit{Input:} A proposed Hypothesis (Emotion + Intensity) AND the Region Descriptions.\\[4pt]
\textit{Instruction:}\\
Verify if the contents of the cited Regions actually support the claimed intensity.
\begin{enumerate}[leftmargin=*]
    \item Fact Check: The hypothesis claims Region [X] shows evidence of High Intensity. Look at the description of Region [X]. Is this true? (e.g., If Region X is a "relaxed hand", it contradicts "High Excitement").
    \item Completeness: Do other Regions (e.g., Region [Y]) suggest a different conclusion?
\end{enumerate}
\textit{Verdict:}
\begin{itemize}
    \item Score (0.0 - 1.0): Probability that this hypothesis is correct based strictly on the regions.
    \item Critique: Mention specifically if a Region ID was misinterpreted.
\end{itemize}
\end{promptbox}

\vspace{12pt}

% ===================== ER =====================
\noindent \textbf{GAT Prompt for ER.}

\begin{promptbox}{Emotion Reasoning (ER)}
\small
You are given an image with segmented regions.\\[4pt]
\textit{Step 1 -- Extract visual clues.}\\
Identify the important visual elements from the regions (people, actions, objects, background, lighting, color, and composition).\\[4pt]
\textit{Step 2 -- Explain the emotion.}\\
Using only these grounded clues, explain which emotion the image conveys and why. Refer explicitly to the regions or elements that support your interpretation.\\[4pt]
Provide a concise, evidence-based explanation grounded entirely in the visual content.
\end{promptbox}

\vspace{12pt}

% ===================== EGCG =====================
\noindent \textbf{GAT Prompt for EGCG.}

\begin{promptbox}{Emotion-Grounded Content Generation (EGCG)}
\small
You are given an image with segmented regions and a target emotion label.\\[4pt]
\textit{Step 1 -- Identify visual anchors.}\\
From the segmented regions, list the key elements that should guide the emotion (e.g., characters, objects, environment, lighting, colors, and their relationships).\\[4pt]
\textit{Step 2 -- Generate an emotionally aligned description.}\\
Write a short paragraph that is emotionally consistent with the target emotion and grounded in the listed visual elements. Use specific details rather than generic statements, and avoid adding objects or events that are not present in the image.
\end{promptbox}

\vspace{6pt}

\section{Experiments compute resources}
% The appendix includes information on the computer resources used, such
% as GPU type, memory, and execution time.
All open-source models were evaluated using NVIDIA A100 GPUs on a cloud computing platform. Each evaluation task was executed on a single GPU, depending on model size. For most 7B models, each task (e.g., EU or ER) required approximately 1–2 hours to complete over the full benchmark set. Larger models (13B–16B) typically required 2–4 hours. Inference was run with float16 precision using the Hugging Face Transformers or official inference libraries where available.

Closed-source models were accessed via public APIs, including OpenAI, Google, and Alibaba endpoints. All requests were made under default inference settings without any batch acceleration.

% \subsection{Broader Impacts}

% This work contributes to the broader goal of building emotionally intelligent AI systems by introducing a benchmark that rigorously evaluates vision-language models on image-based affective understanding, reasoning, and emotionally aligned content generation.

% As emotional intelligence becomes increasingly critical in AI applications—ranging from mental health support to education and empathetic assistants—this benchmark offers a systematic foundation to assess and advance models' affective competence. The World Health Organization (WHO) has highlighted a 25\% global surge in depression and anxiety following the COVID-19 pandemic and called for technological systems that can detect and respond to emotional distress\footnote{\url{https://www.who.int/news/item/02-03-2022-covid-19-pandemic-triggers-25-increase-in-prevalence-of-anxiety-and-depression-worldwide} (Accessed: February 10, 2025).}. Similarly, UNESCO advocates for emotionally intelligent digital tools to reduce learning disparities and support student well-being\footnote{\url{https://www.oecd.org/en/about/projects/future-of-education-and-skills-2030.html} (Accessed: February 10, 2025).}.

% By promoting models that can recognize and respond to emotional content with greater nuance and sensitivity, our work supports the development of more human-centered AI systems. This is particularly relevant given recent reports of emotionally tone-deaf chatbot responses in sensitive mental health contexts, which may lead to user discomfort or harm\footnote{\url{https://www.theguardian.com/technology/2024/oct/23/character-ai-chatbot-sewell-setzer-death} (Accessed: February 10, 2025).}.
% 

\section{License and Intended Use}
\label{app:license_intended_use}
We will release the AICA-Bench artifacts under the CC BY 4.0 license for academic use. We follow the original terms of all source datasets used for secondary use and do not redistribute any restricted raw content; users should obtain any required original images from the respective sources under their terms. AICA-Bench is intended for research and benchmarking of affective image content analysis with vision-language models, and is not intended for high-stakes or sensitive applications or for identifying individuals.

\section{Show Case}

\subsection{Emotion Understanding Case}

\begin{figure}[H]
    \centering
    % \captionsetup{skip=2pt}
    \includegraphics[width=0.8\linewidth]{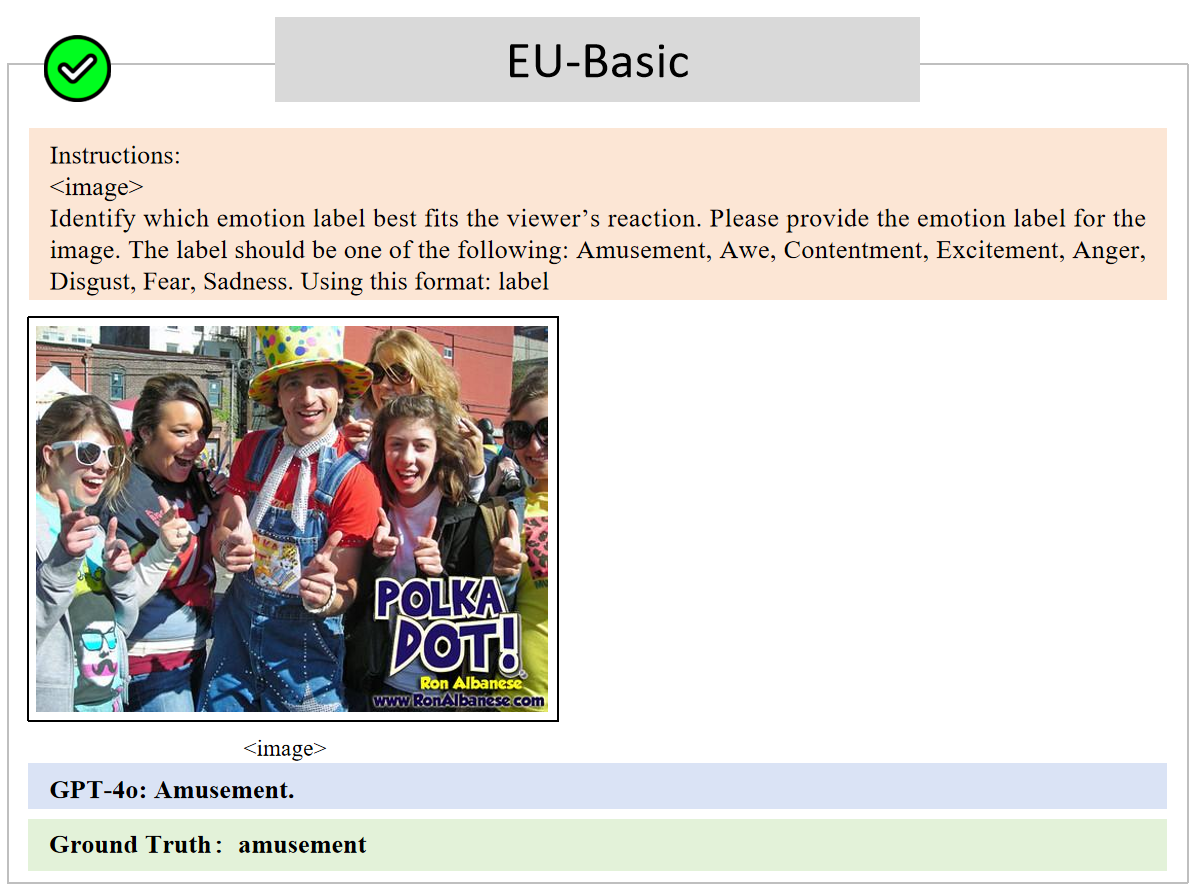}
    \caption{A sample correct case of EU-Basic.}
    \label{sup:fig:eu-basic1}
\end{figure}

\begin{figure}[H]
    \centering
    % \captionsetup{skip=2pt}
    \includegraphics[width=0.8\linewidth]{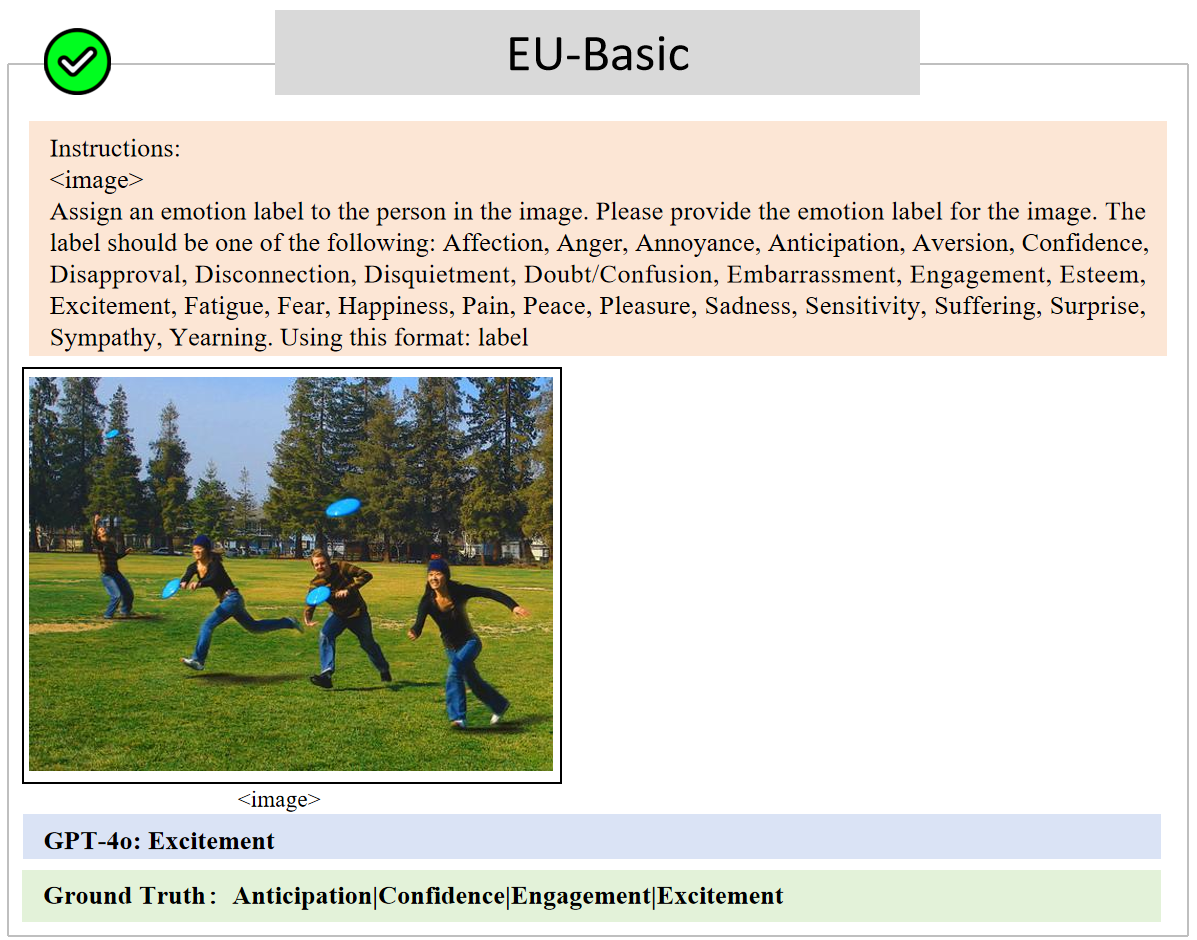}
    \caption{A sample correct case of EU-Basic.}
    \label{sup:fig:eu-basic2}
\end{figure}

\begin{figure}[H]
    \centering
    % \captionsetup{skip=2pt}
    \includegraphics[width=0.8\linewidth]{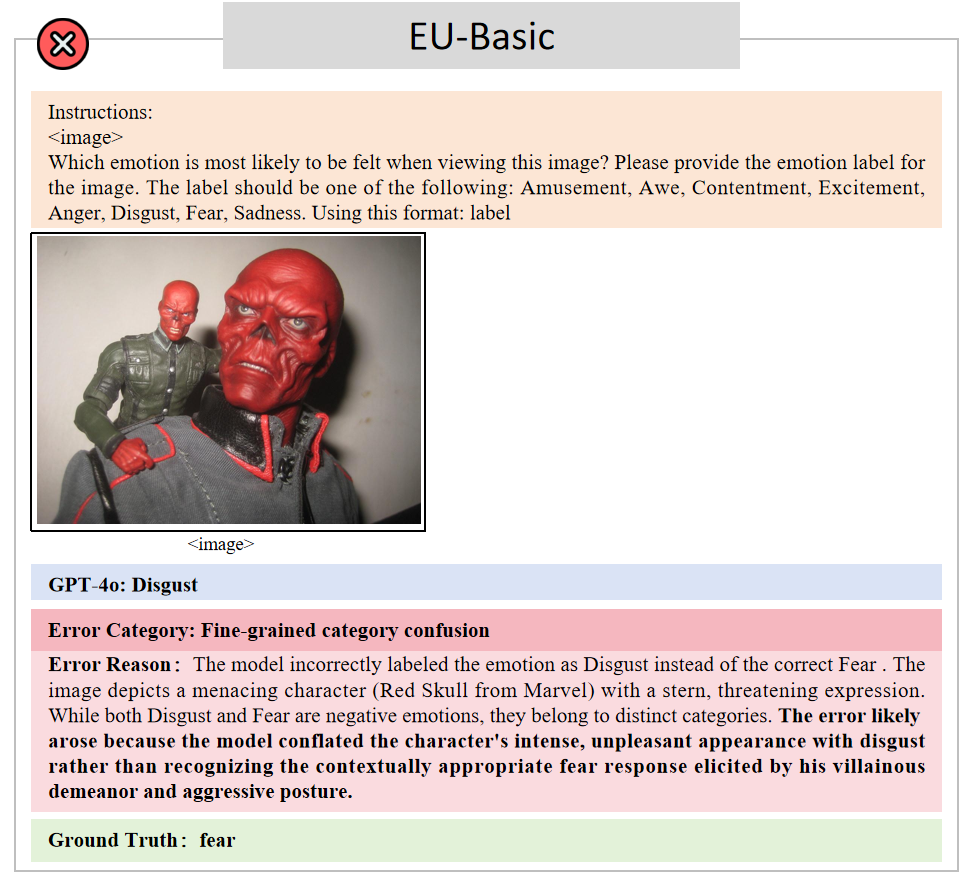}
    \caption{A sample error case of EU-Basic.}
    \label{sup:fig:eu-basic3}
\end{figure}

\begin{figure}[H]
    \centering
    % \captionsetup{skip=2pt}
    \includegraphics[width=0.8\linewidth]{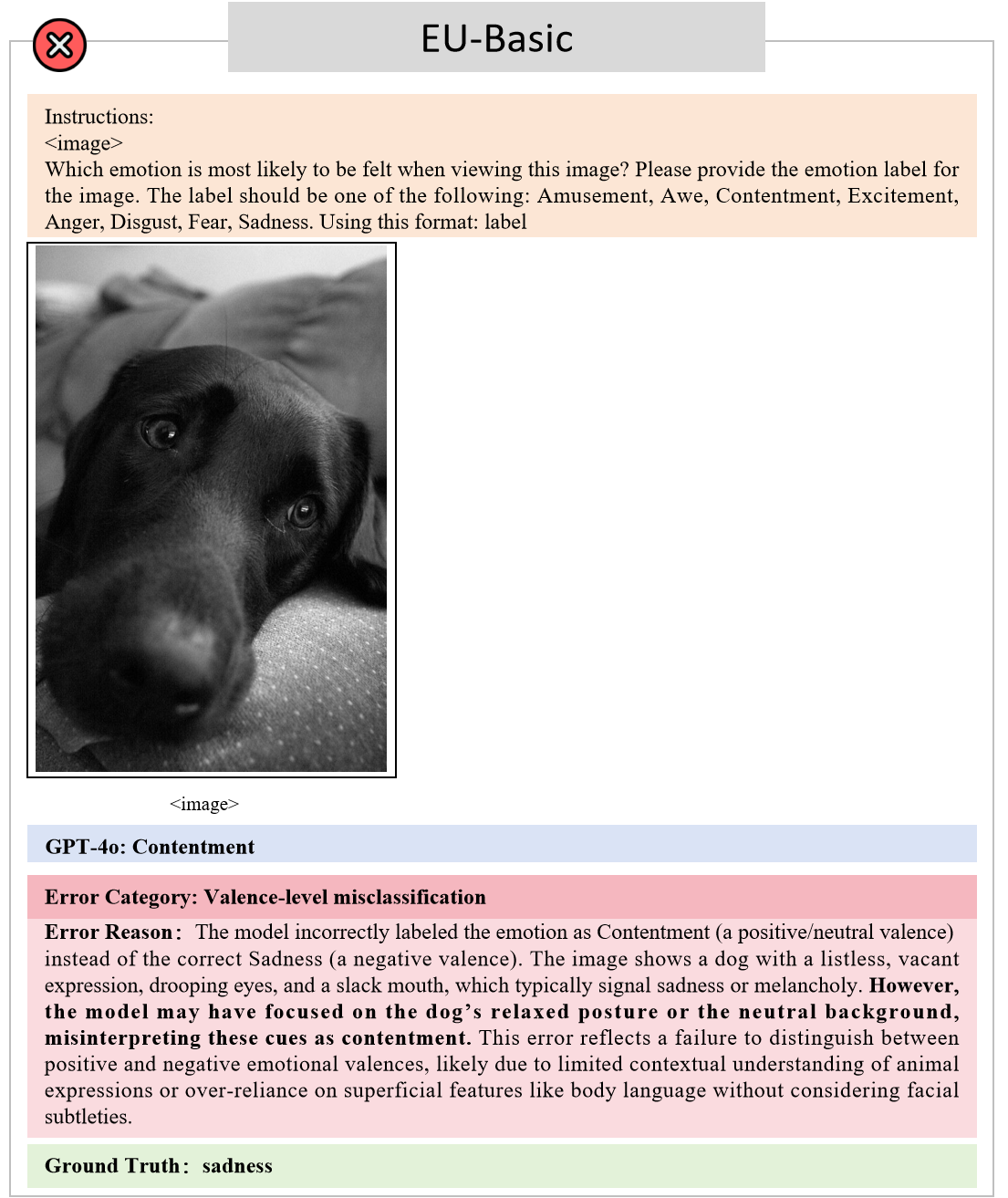}
    \caption{A sample error case of EU-Basic.}
    \label{sup:fig:eu-basic4}
\end{figure}

\begin{figure}[H]
    \centering
    % \captionsetup{skip=2pt}
    \includegraphics[width=0.8\linewidth]{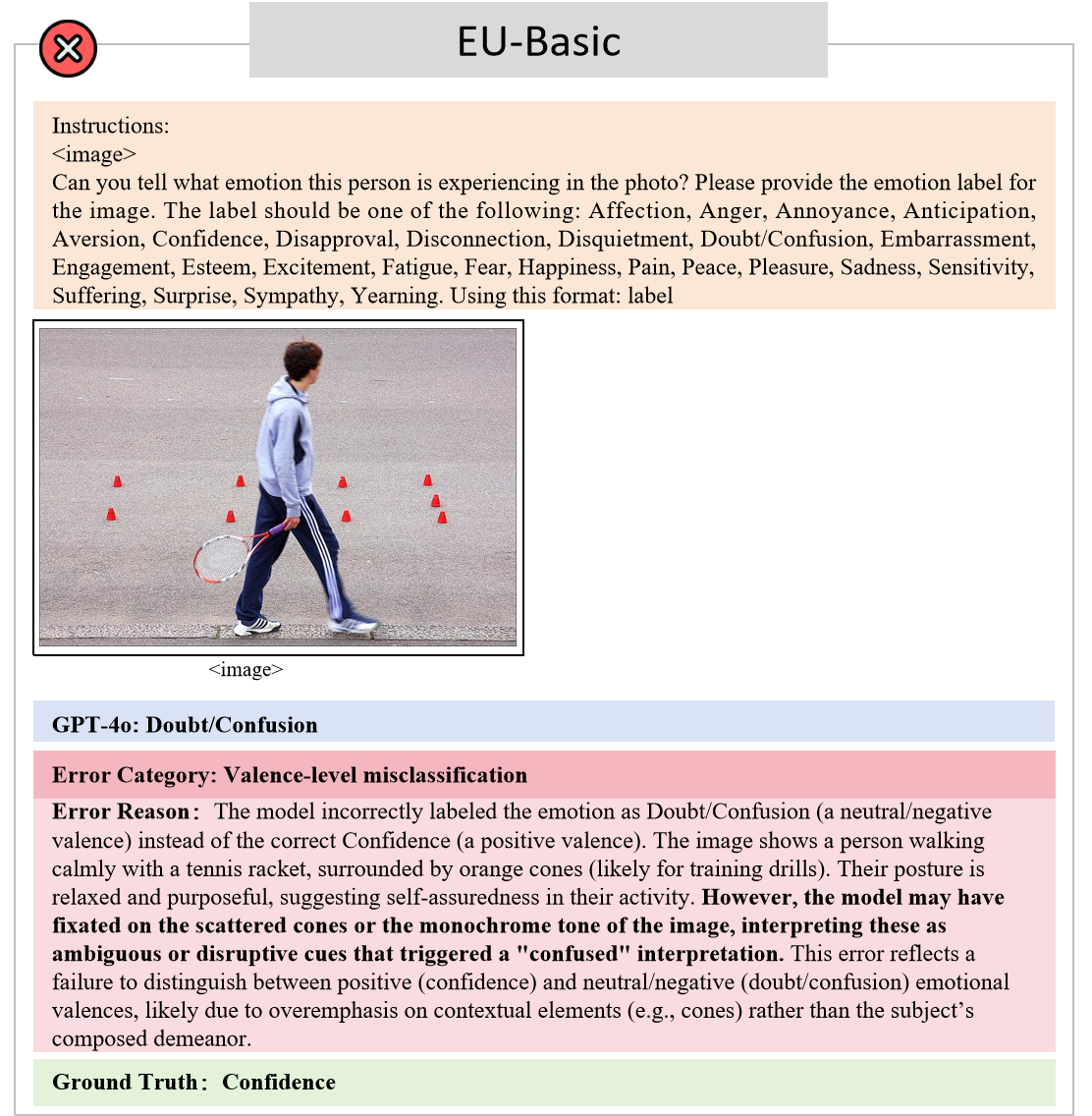}
    \caption{A sample error case of EU-Basic.}
    \label{sup:fig:eu-basic5}
\end{figure}

\begin{figure}[H]
    \centering
    % \captionsetup{skip=2pt}
    \includegraphics[width=0.8\linewidth]{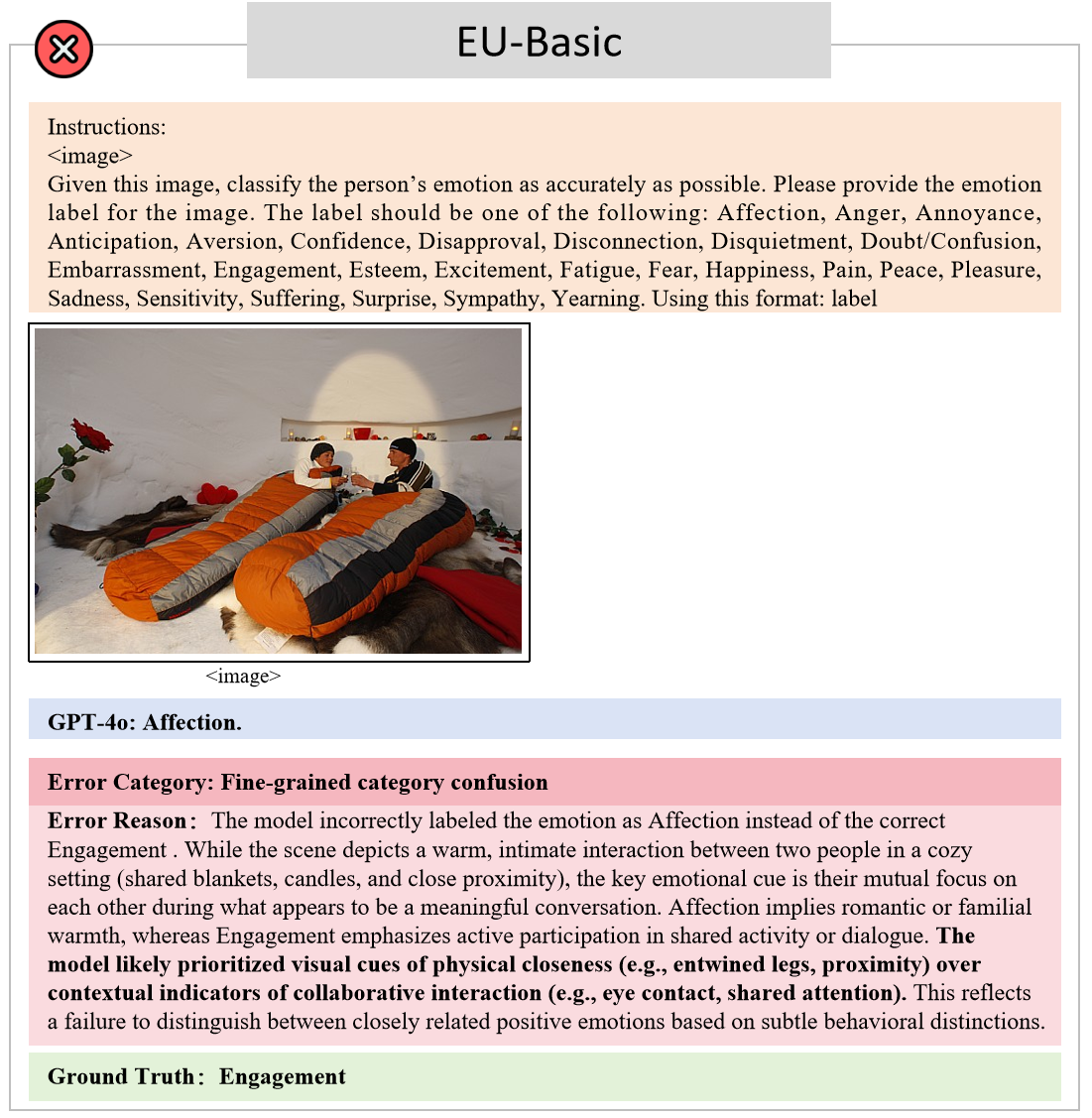}
    \caption{A sample error case of EU-Basic.}
    \label{sup:fig:eu-basic6}
\end{figure}

\begin{figure}[H]
    \centering
    % \captionsetup{skip=2pt}
    \includegraphics[width=0.8\linewidth]{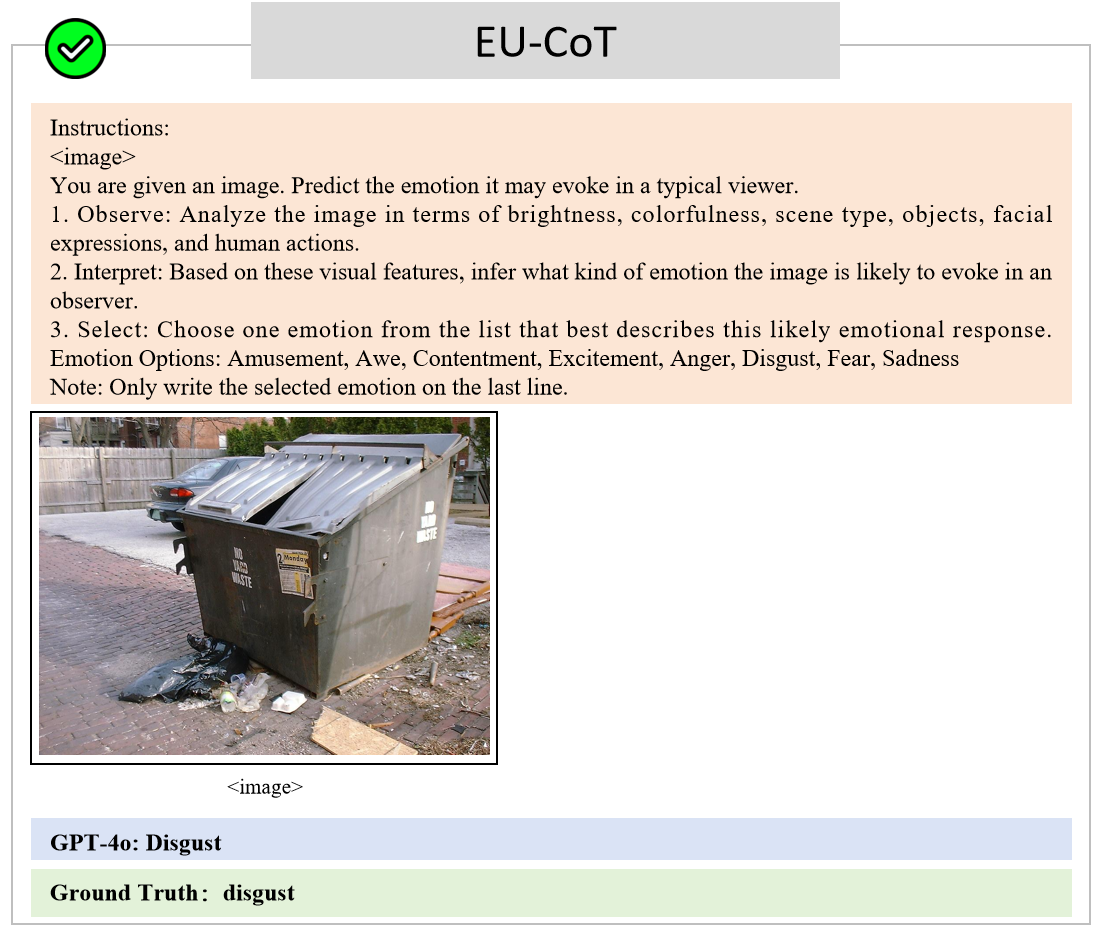}
    \caption{A sample correct case of EU-CoT.}
    \label{sup:fig:eu-cot1}
\end{figure}

\begin{figure}[H]
    \centering
    % \captionsetup{skip=2pt}
    \includegraphics[width=0.8\linewidth]{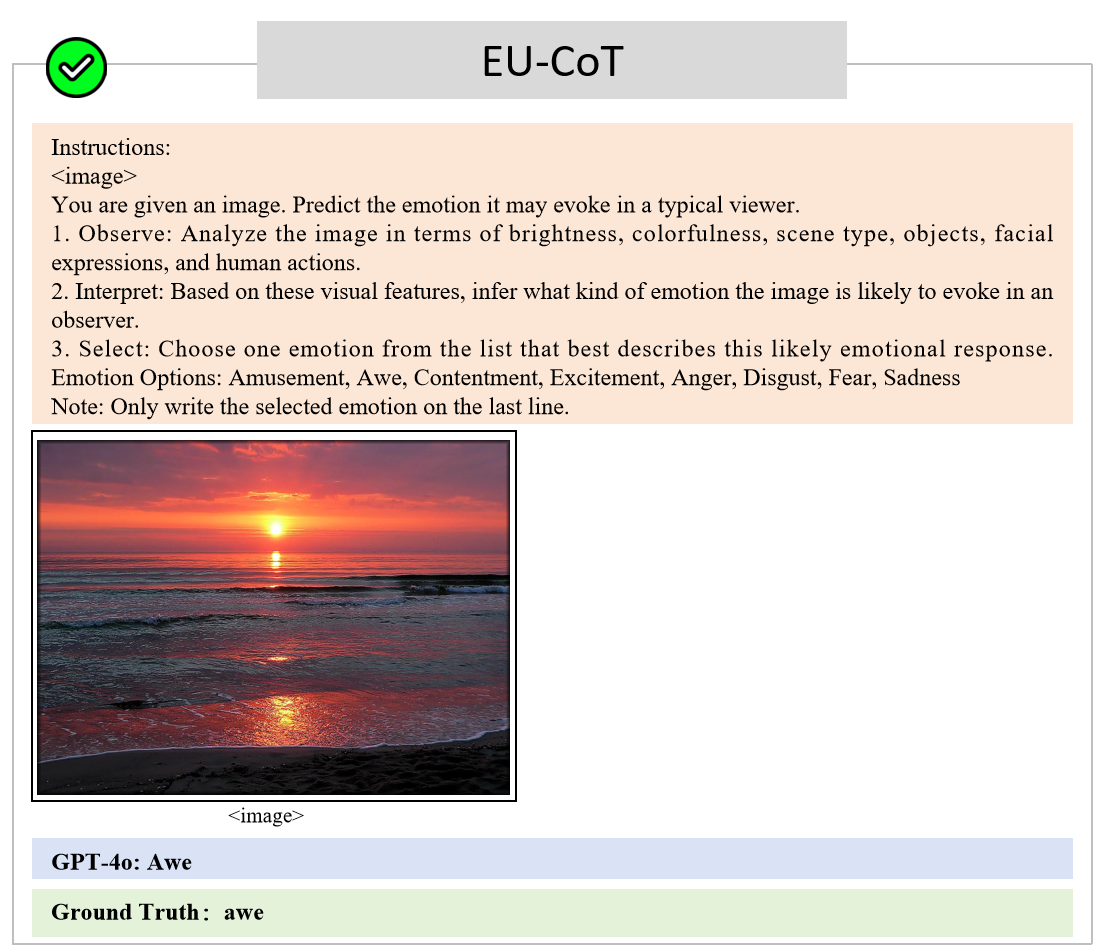}
    \caption{A sample correct case of EU-CoT.}
    \label{sup:fig:eu-cot2}
\end{figure}

\begin{figure}[H]
    \centering
    % \captionsetup{skip=2pt}
    \includegraphics[width=0.8\linewidth]{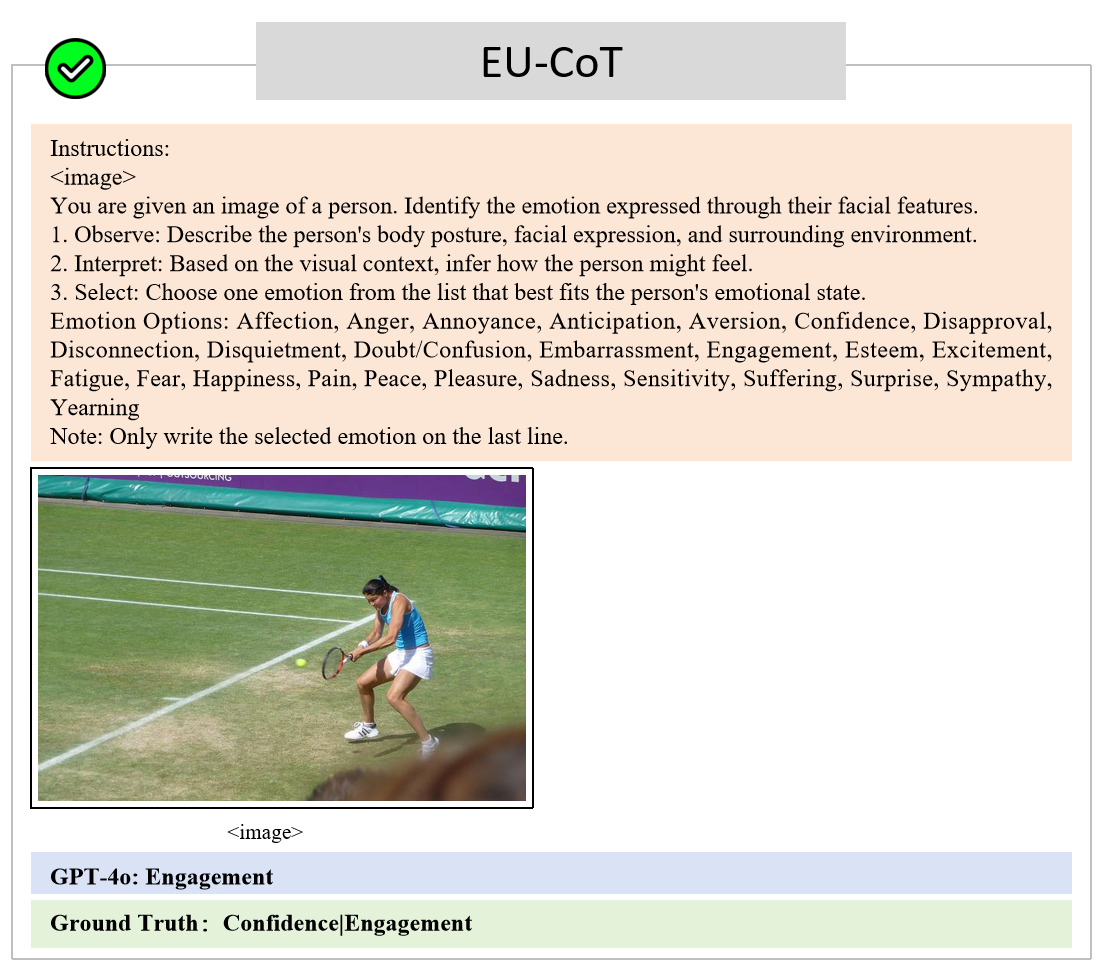}
    \caption{A sample correct case of EU-CoT.}
    \label{sup:fig:eu-cot3}
\end{figure}

\begin{figure}[H]
    \centering
    % \captionsetup{skip=2pt}
    \includegraphics[width=0.8\linewidth]{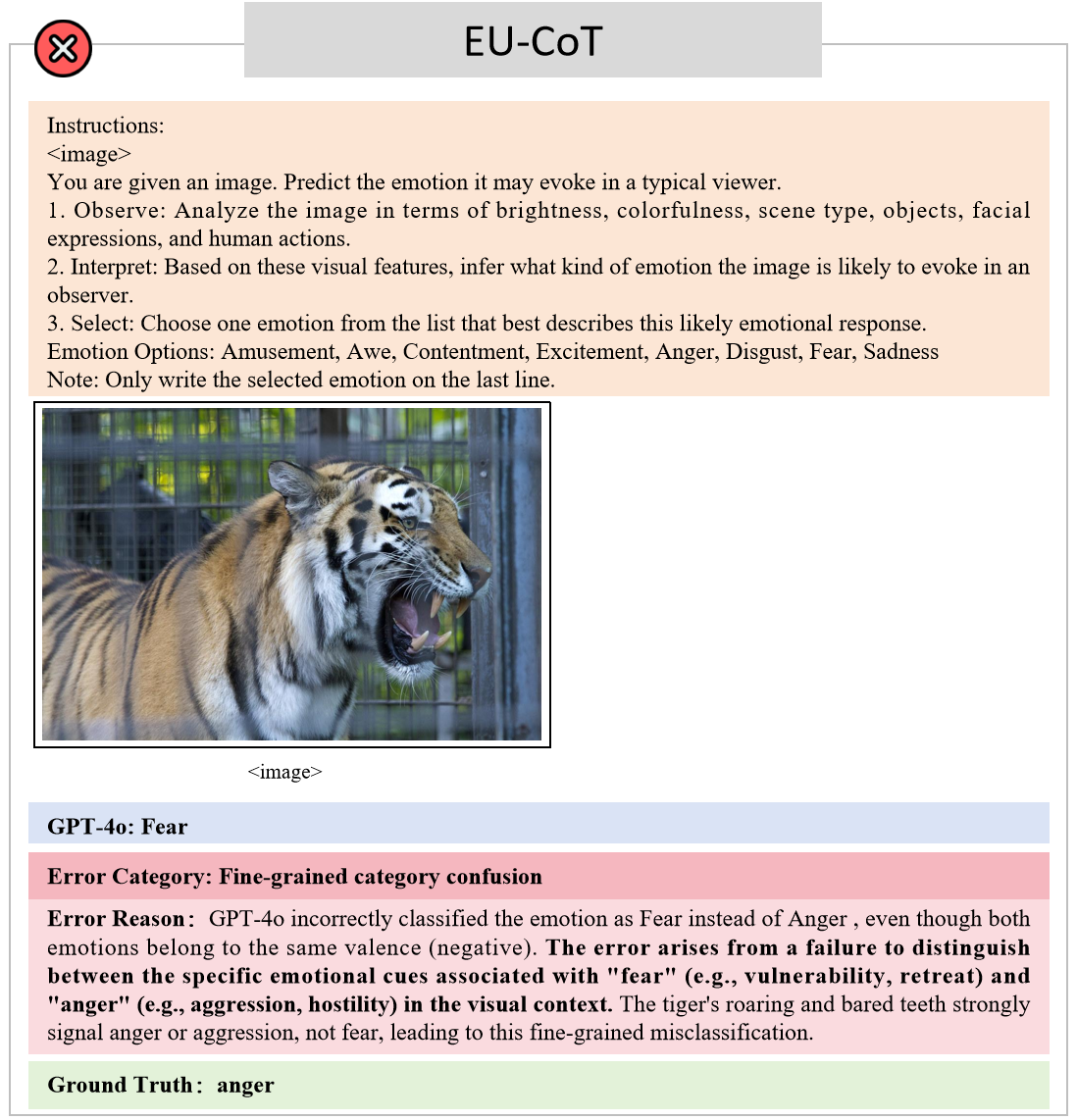}
    \caption{A sample error case of EU-CoT.}
    \label{sup:fig:eu-cot4}
\end{figure}

\begin{figure}[H]
    \centering
    % \captionsetup{skip=2pt}
    \includegraphics[width=0.8\linewidth]{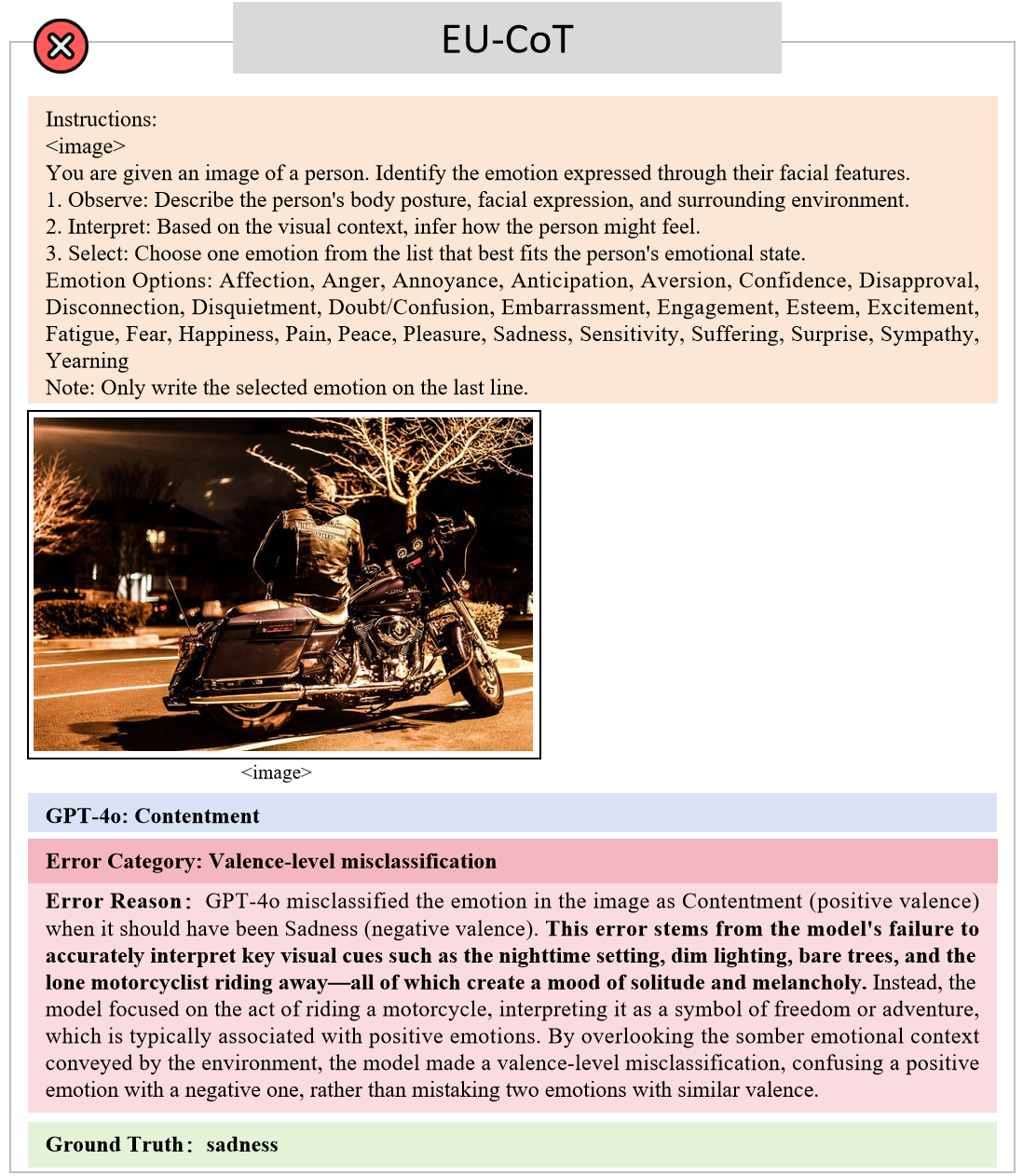}
    \caption{A sample error case of EU-CoT.}
    \label{sup:fig:eu-cot5}
\end{figure}

\begin{figure}[H]
    \centering
    % \captionsetup{skip=2pt}
    \includegraphics[width=\linewidth]{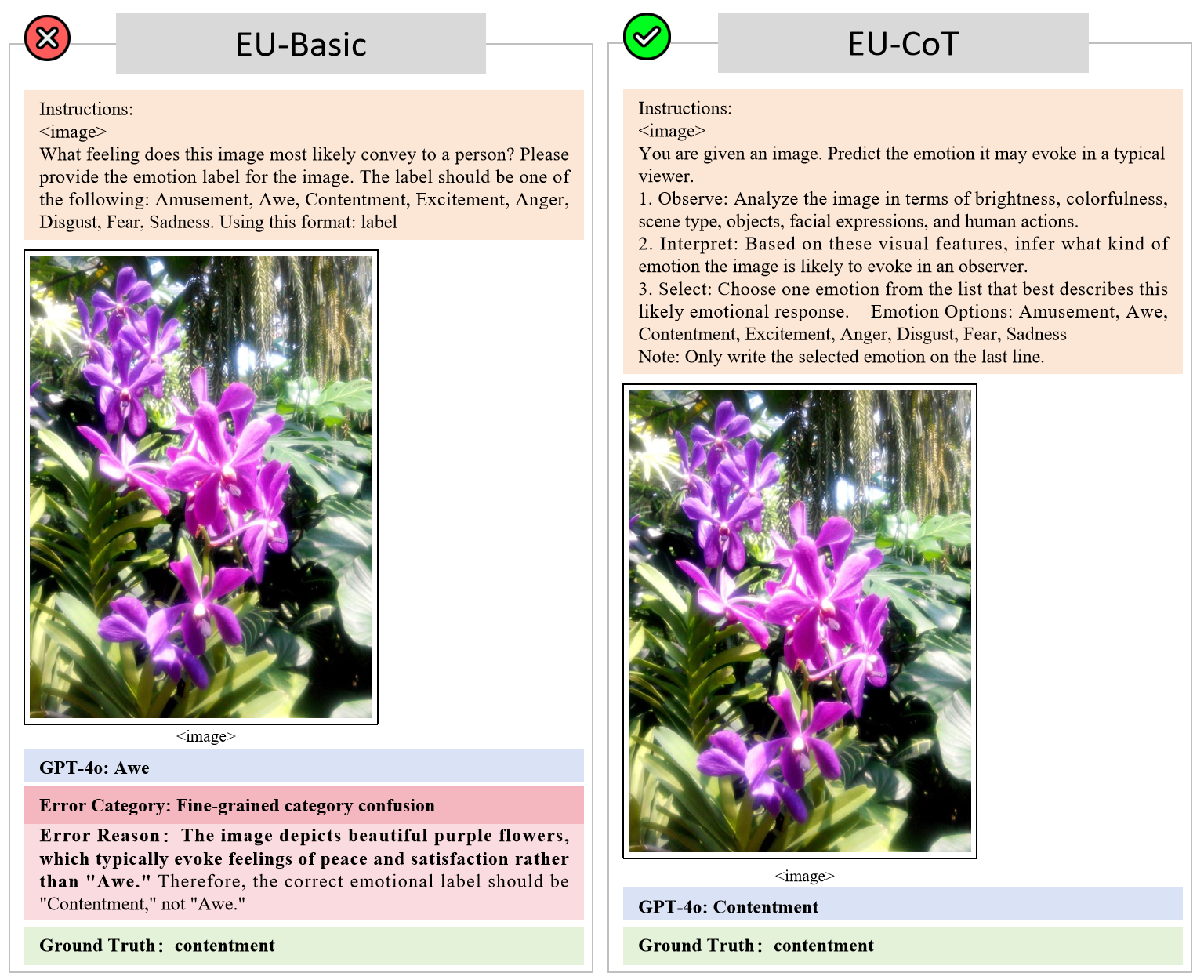}
    \caption{Basic Prompting vs. CoT Prompting Case Study}
    \label{sup:fig:EU_basicvscot_1}
\end{figure}

We present sample analyses of EU cases, including 5 correct and 8 error examples (see Figures~\ref{sup:fig:eu-basic1}, \ref{sup:fig:eu-basic2}, \ref{sup:fig:eu-basic3}, \ref{sup:fig:eu-basic4}, \ref{sup:fig:eu-basic5}, \ref{sup:fig:eu-basic6}, \ref{sup:fig:eu-cot1}, \ref{sup:fig:eu-cot2}, \ref{sup:fig:eu-cot3}, \ref{sup:fig:eu-cot4}, \ref{sup:fig:eu-cot5}, \ref{sup:fig:EU_basicvscot_1})
.

\begin{figure}[H]
    \centering
    % \captionsetup{skip=2pt}
    \includegraphics[width=0.8\linewidth]{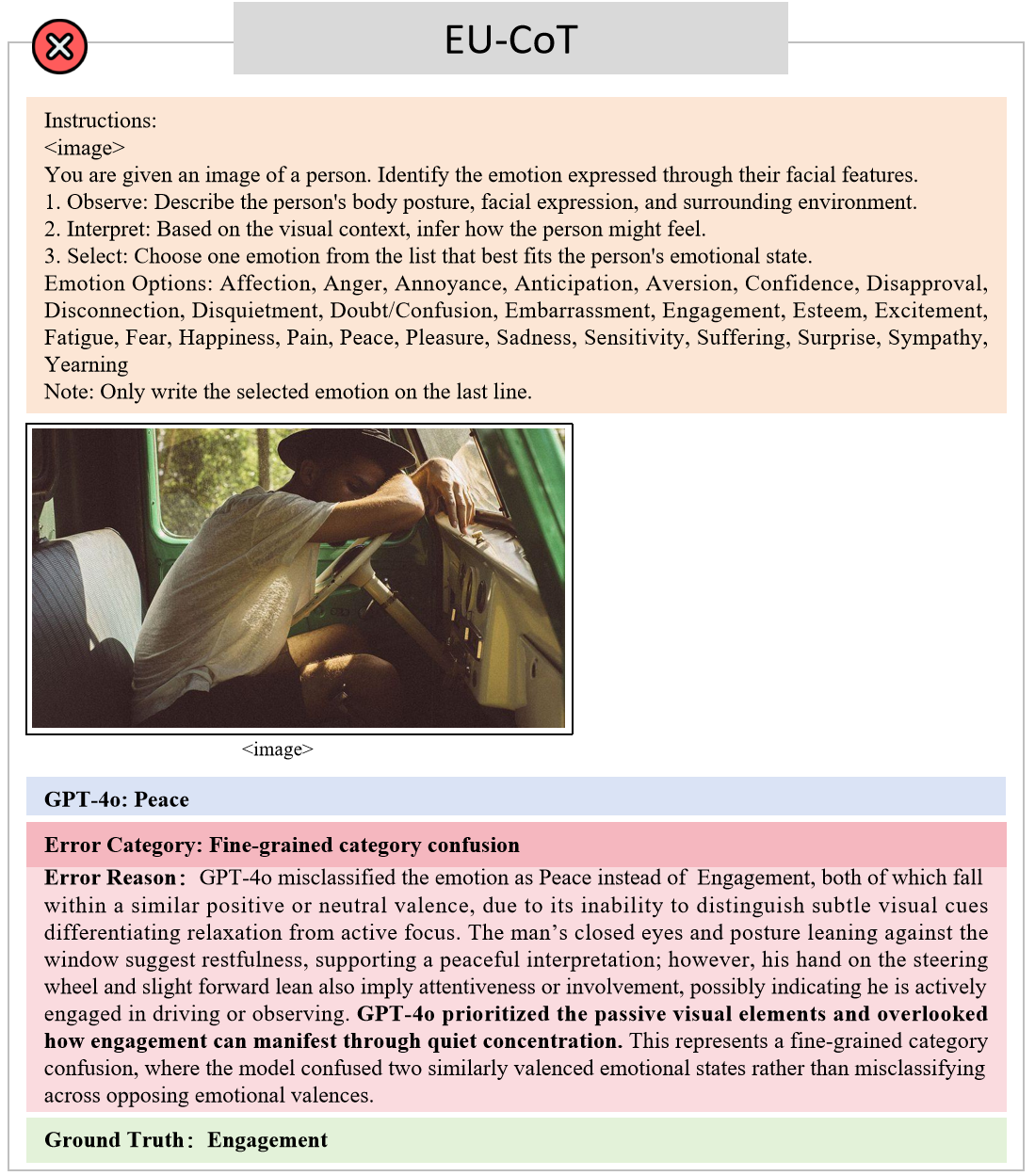}
    \caption{A sample error case of EU-CoT.}
    \label{sup:fig:eu-cot6}
\end{figure}

\begin{figure}[H]
    \centering
    % \captionsetup{skip=2pt}
    \includegraphics[width=0.8\linewidth]{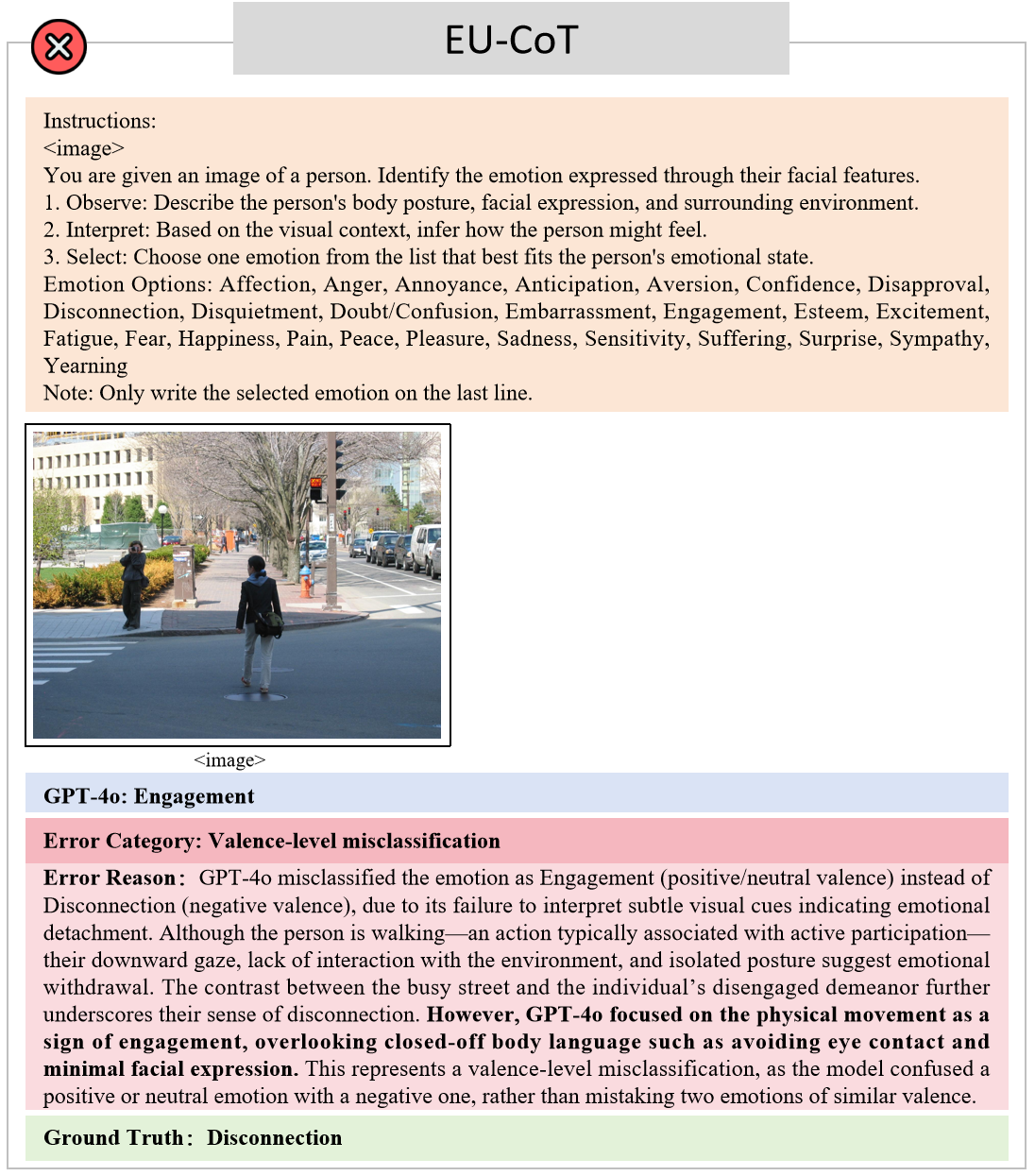}
    \caption{A sample error case of EU-CoT.}
    \label{sup:fig:eu-cot7}
\end{figure}

% \subsubsection{Basic Prompting vs. CoT Prompting Case Study}

\subsection{Emotion Reasoning Case}

\begin{figure}[H]
    \centering
    % \captionsetup{skip=2pt}
    \includegraphics[width=\linewidth]{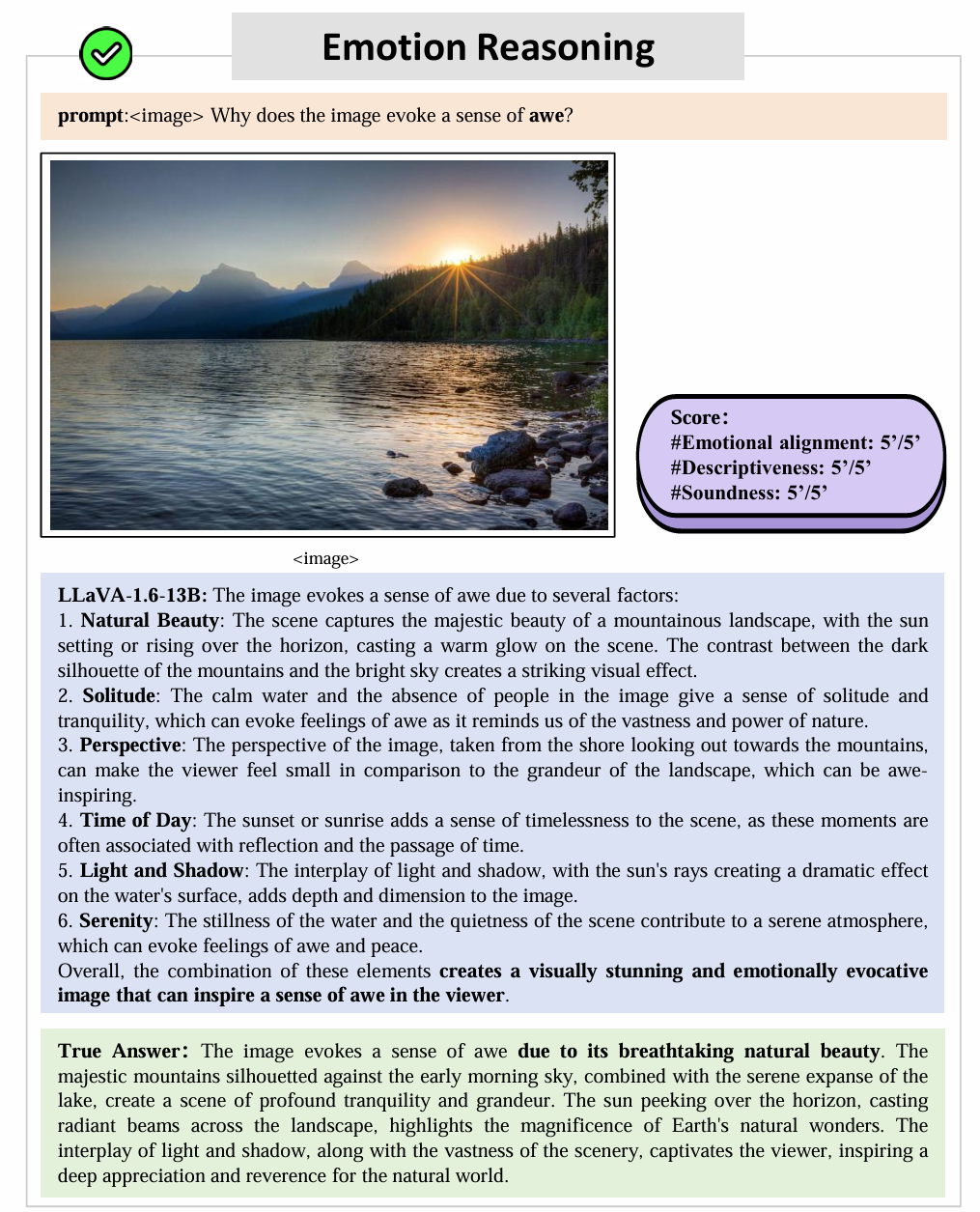}
    \caption {A sample correct case of EGCG}
    \label{sup:fig:er1}
\end{figure}

\begin{figure}[H]
    \centering
    % \captionsetup{skip=2pt}
    \includegraphics[width=\linewidth]{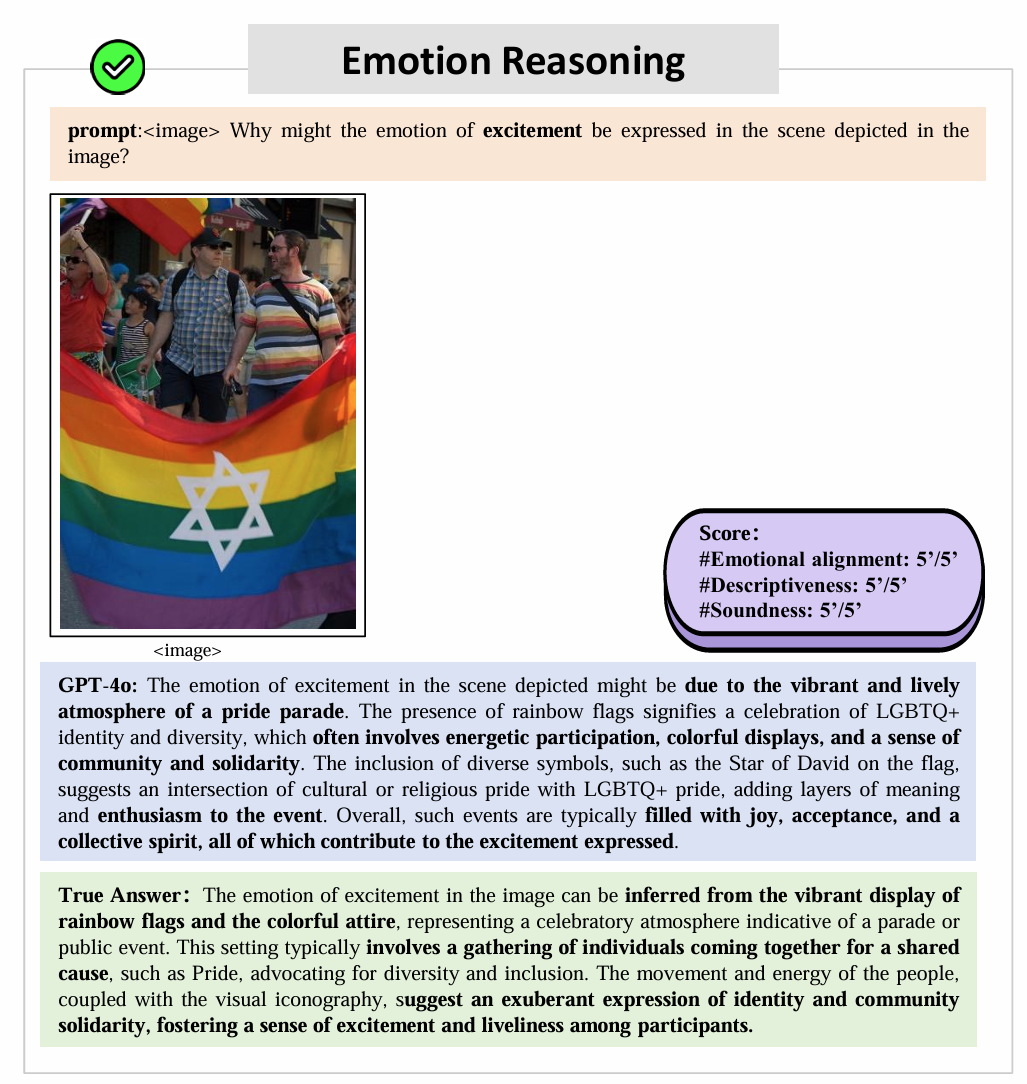}
    \caption {A sample correct case of EGCG}
    \label{sup:fig:er2}
\end{figure}

\begin{figure}[H]
    \centering
    % \captionsetup{skip=2pt}
    \includegraphics[width=\linewidth]{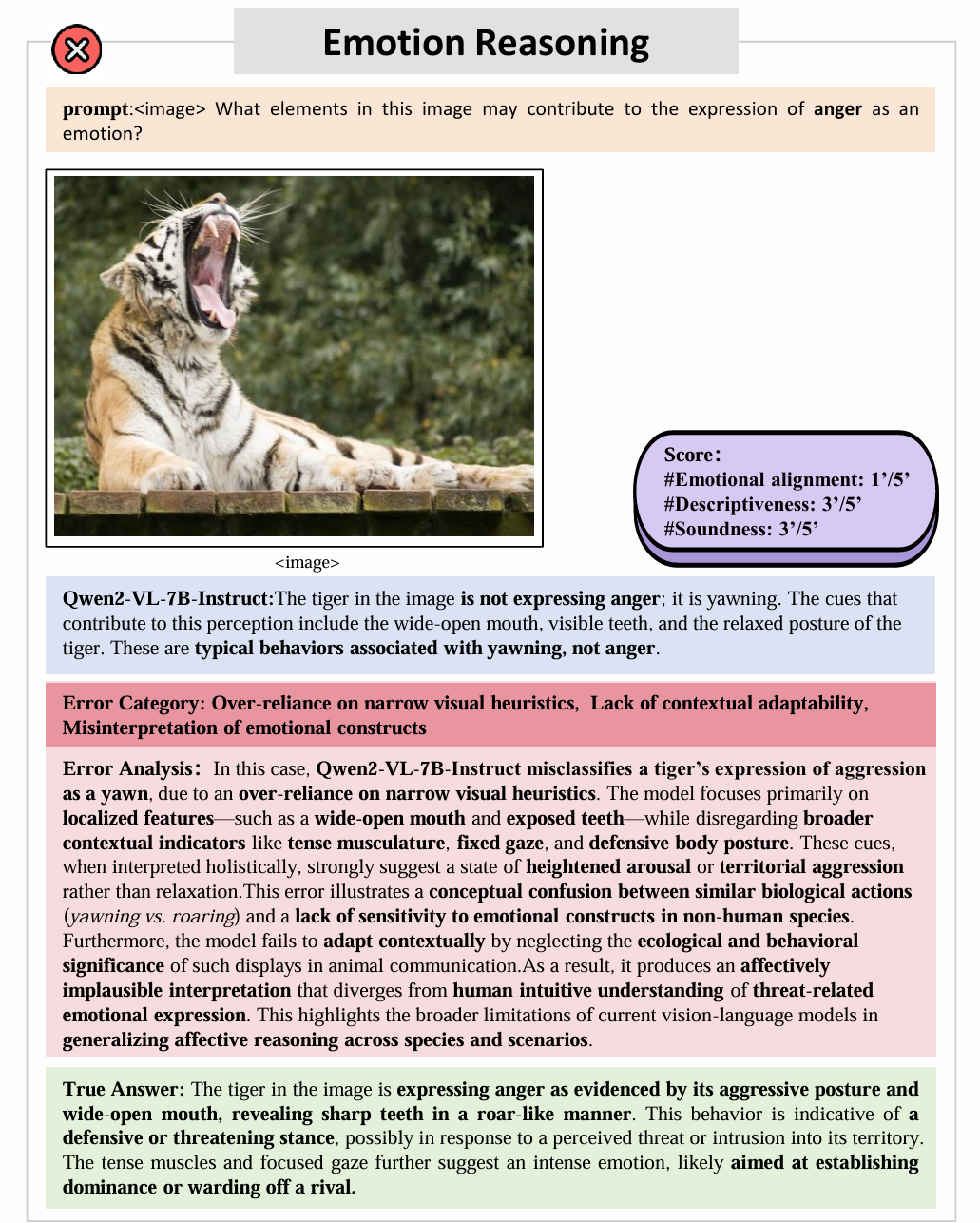}
    \caption {A sample correct case of EGCG}
    \label{sup:fig:er3}
\end{figure}

\begin{figure}[H]
    \centering
    % \captionsetup{skip=2pt}
    \includegraphics[width=\linewidth]{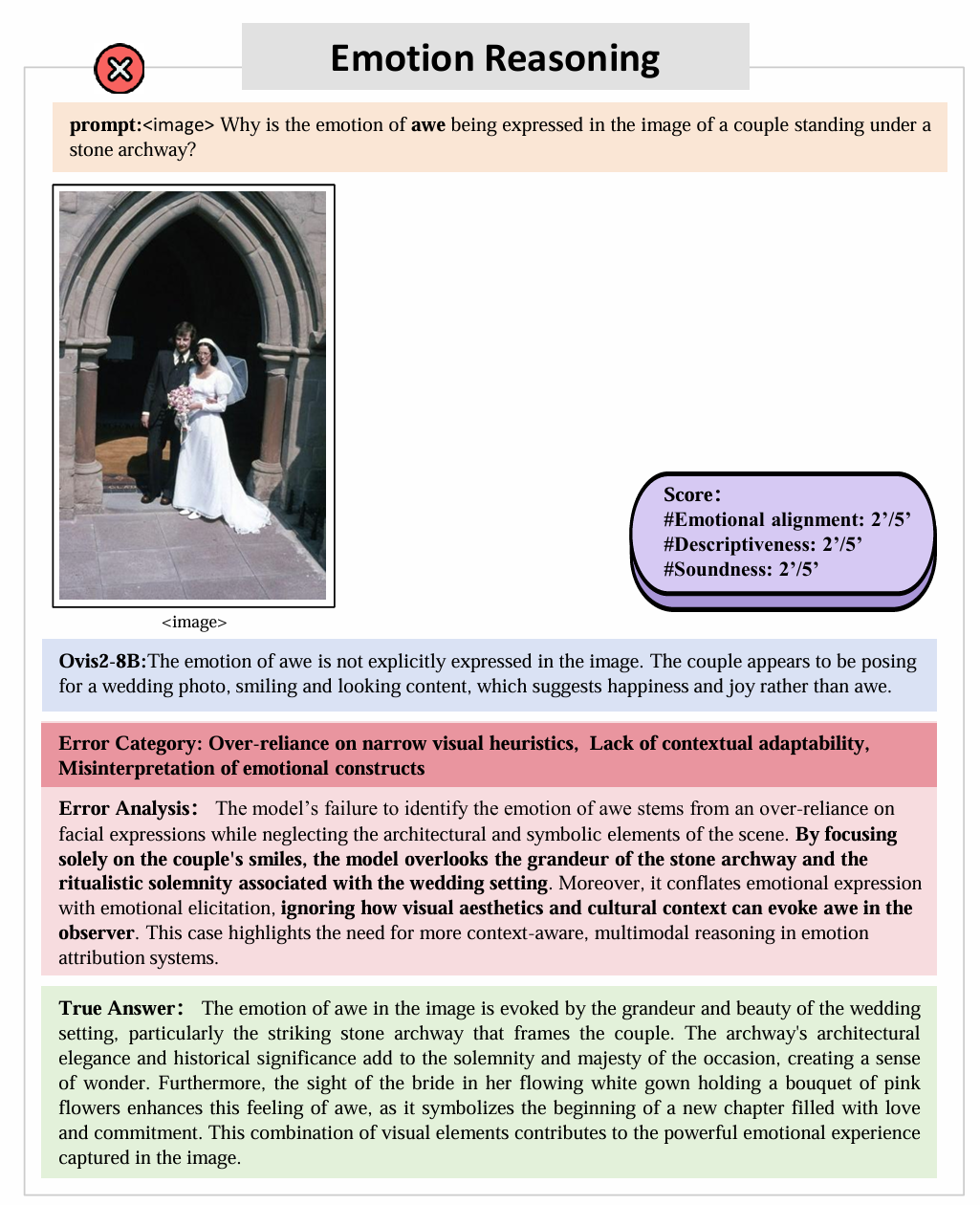}
    \caption {A sample correct case of EGCG}
    \label{sup:fig:er4}
\end{figure}

The following presents our sample analysis of Emotion Reasoning (ER) cases, including representative correct and error examples. Refer to Figures~\ref{sup:fig:er1},\ref{sup:fig:er2},\ref{sup:fig:er3}, and~\ref{sup:fig:er4} for detailed illustrations.

In each figure, we manually annotate the \textit{True Answer}, which represents the ground-truth reasoning outcome based on the emotion context of the image. This information was not provided by the model, but serves as a human-verified reference to facilitate comparison.
For the bad cases, we further conduct manual analysis to identify typical reasoning failures, such as incorrect emotion attribution, misunderstanding of causal context, or mismatches between textual and visual cues.

% In Figure
% ~\ref{sup:fig:er1}, a mountainous landscape at sunrise or sunset evokes a strong sense of awe. The model LLaVA-1.6-13B accurately attributes this emotional response to the scene’s natural beauty, serenity, and expansive perspective. It effectively integrates visual features such as light, shadow, time of day, and environmental stillness to construct a holistic emotional explanation. This response closely mirrors human affective interpretation, recognizing both the aesthetic and symbolic power of vast, tranquil nature scenes.

% In Figure ~\ref{sup:fig:ER2}, Qwen2-VL-7B-Instruct misinterprets a tiger’s display of aggression as a yawn. The model’s reasoning is based solely on localized visual features—such as a wide-open mouth and visible teeth—without accounting for broader contextual indicators like muscular tension, focused gaze, and defensive body posture. These cues, when considered holistically, suggest a state of heightened arousal and territorial aggression. The model fails to recognize this affective significance due to its narrow reliance on visual heuristics and its limited ability to interpret emotion in non-human species. As a result, the output appears biologically plausible but emotionally inaccurate, diverging from human intuitive understanding of threat-related expressions in animals.

\subsection{Emotion-guided Content Generation Case}
\begin{figure}[H]
    \centering
    % \captionsetup{skip=2pt}
    \includegraphics[width=\linewidth]{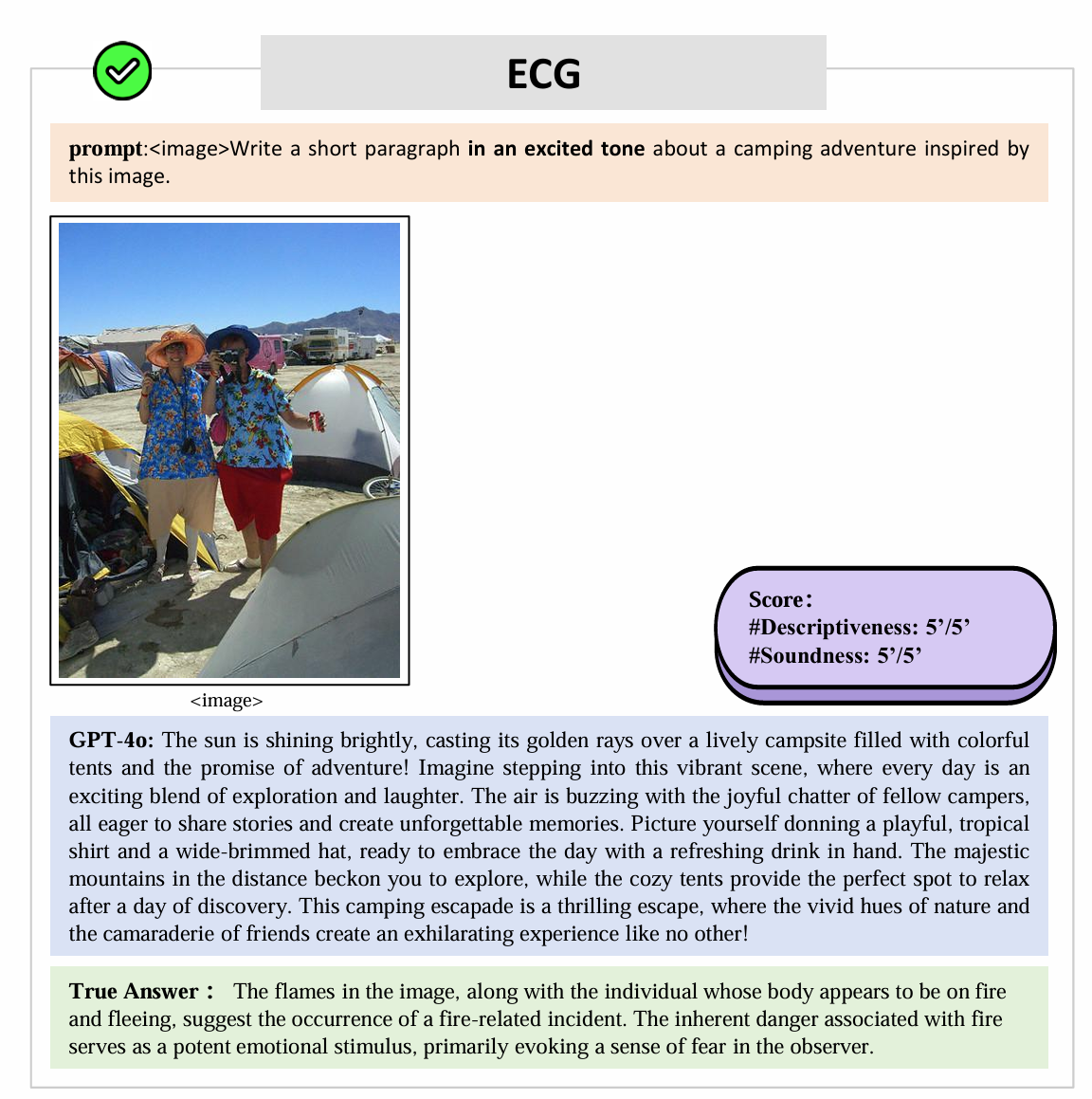}
    \caption {A sample correct case of EGCG}
    \label{sup:fig:ecg1}
\end{figure}

\begin{figure}[H]
    \centering
    % \captionsetup{skip=2pt}
    \includegraphics[width=\linewidth]{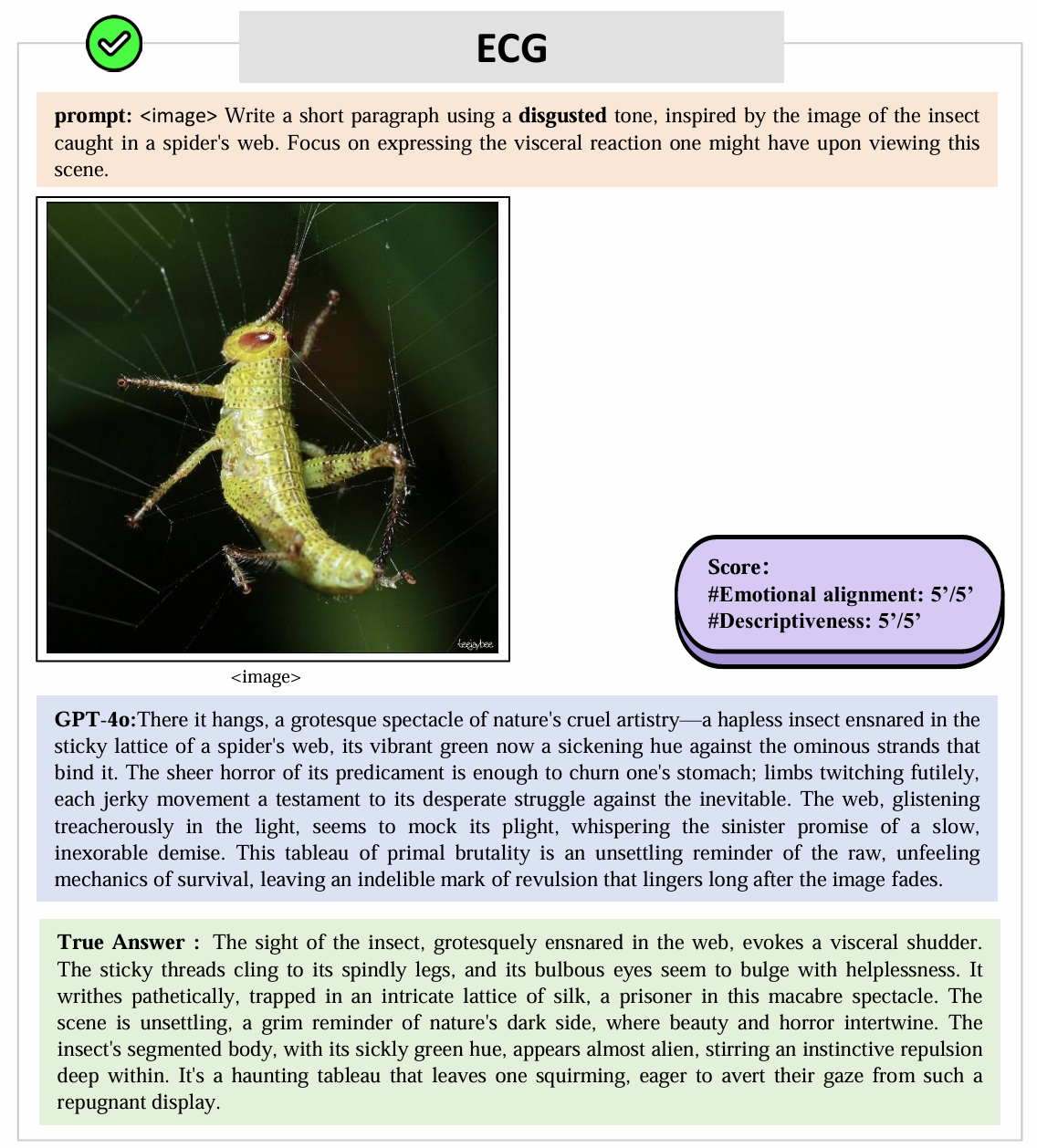}
    \caption {A sample correct case of EGCG}
    \label{sup:fig:ecg2}
\end{figure}

\begin{figure}[H]
    \centering
    % \captionsetup{skip=2pt}
    \includegraphics[width=\linewidth]{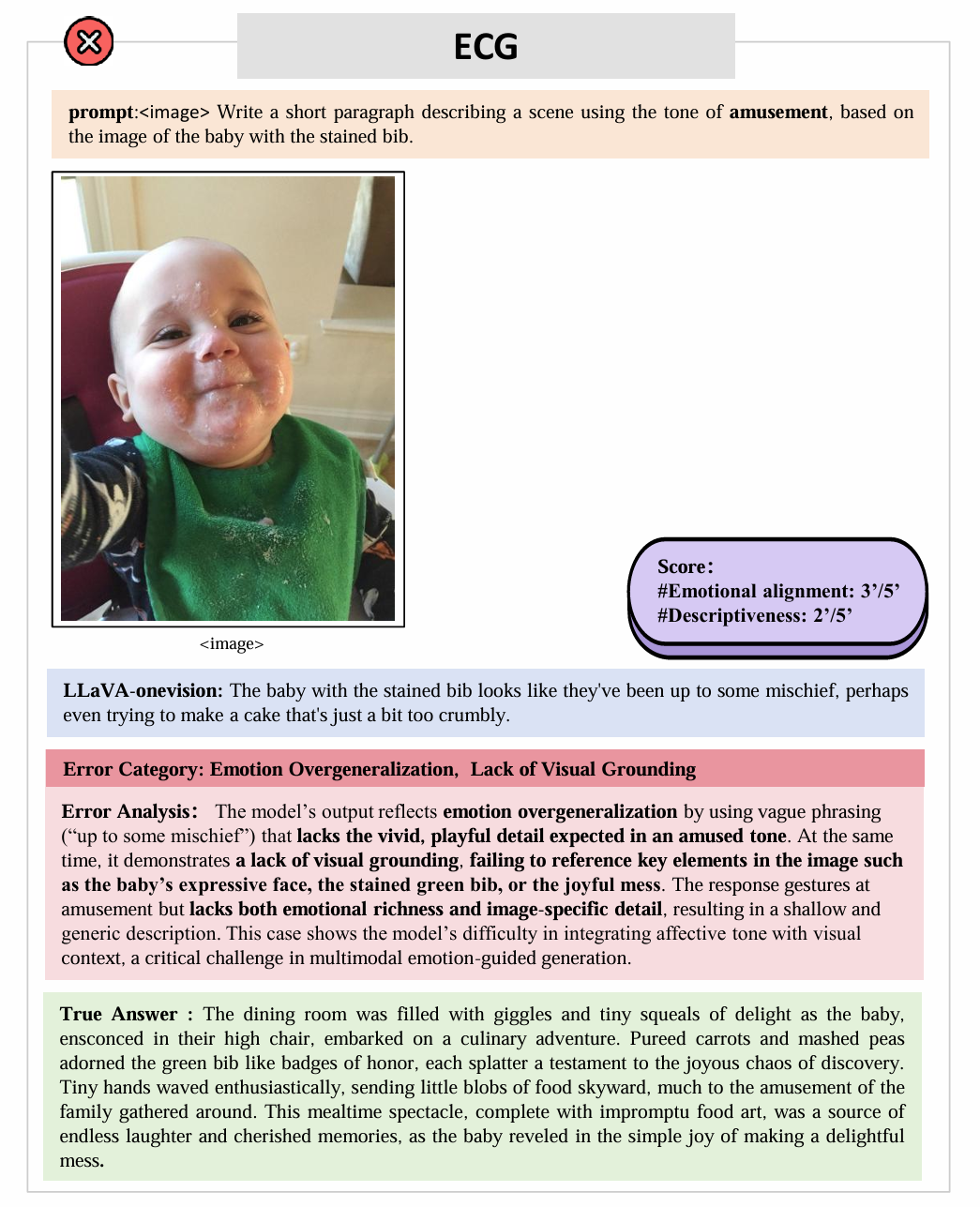}
    \caption {A sample error case of EGCG}
    \label{sup:fig:ecg3}
\end{figure}

\begin{figure}[H]
    \centering
    % \captionsetup{skip=2pt}
    \includegraphics[width=\linewidth]{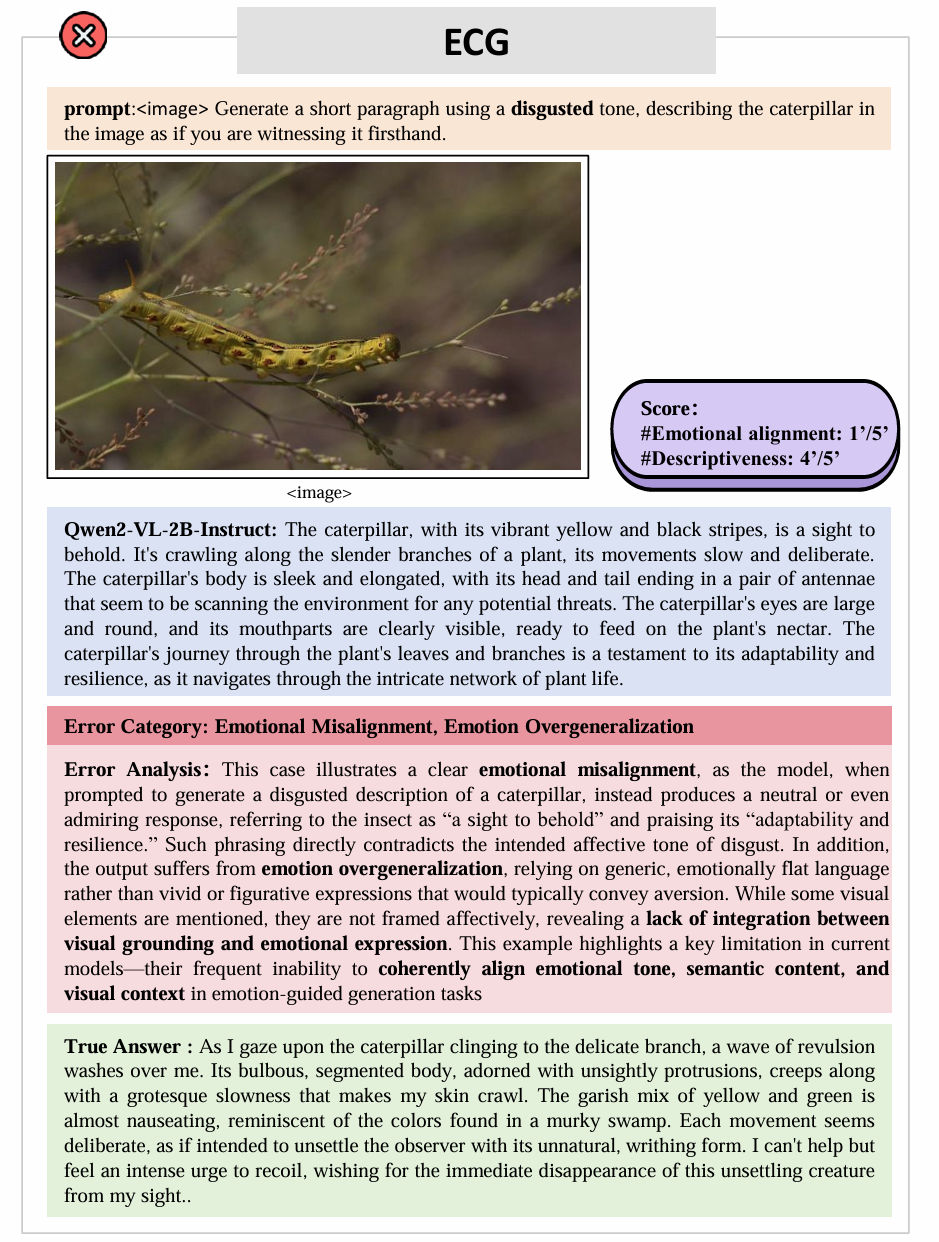}
    \caption {A sample error case of EGCG}
    \label{sup:fig:ecg4}
\end{figure}

We also include representative cases from the EGCG task to illustrate model performance (Figures~\ref{sup:fig:ecg1},~\ref{sup:fig:ecg2}, ~\ref{sup:fig:ecg3}, ~\ref{sup:fig:ecg4}). Each case displays a manually annotated \textit{True Answer}—the target content aligned with the intended emotion. For generation errors, we briefly analyze where the output fails to reflect the emotional intent.
% Current visual-language models demonstrate promising capabilities in generating emotionally coherent and visually relevant descriptions. When tasked with producing content guided by specific emotions, these models generally succeed in reflecting the requested emotional tone through appropriate word choice and narrative style. They often effectively leverage visual details from provided images to enrich emotional expressions, resulting in vivid and contextually fitting narratives. However, their performance can vary depending on the complexity and specificity of the emotional cues and visual context provided.

% Figure ~\ref{sup:fig:EGCG1} features an image of individuals joyfully posing on a camping site. Upon prompting to write an excited description, the model successfully captured the lively atmosphere, emphasizing visual details such as bright sunlight, colorful tents, and the carefree attire of the campers. The generated narrative employs enthusiastic language that aligns convincingly with the depicted scene, illustrating the model's ability to accurately convey excitement by integrating emotional tone with visual context.

% In Figure~\ref{sup:fig:EGCG3}, the model received an image depicting a cheerful baby with a stained bib, prompting a description using an amused tone. However, the generated output described the scene vaguely, merely suggesting the baby was "up to some mischief" without incorporating specific, playful details from the image. This generalization weakened the emotional expressiveness of the response. Moreover, the model failed to reference key visual details such as the baby's joyful expression, the messy bib, or the playful context implied in the scene. Such omissions indicate a clear deficiency in visually grounded narrative generation, limiting both descriptive richness and emotional authenticity.

\end{document}